\def\publicationclean{1}
\documentclass[11pt]{article}
\usepackage[utf8]{inputenc}

\usepackage{amsmath,latexsym,amssymb}
\usepackage{graphicx}
\usepackage{lscape}
\usepackage{eso-pic, tikz} 
\usepackage{dsfont}
\usepackage{appendix}
\usepackage{multirow}
\usepackage{listings}
\usepackage{adjustbox}
\usepackage{rotating}
\usepackage{booktabs}
\usepackage{array}
\usepackage{mdframed}
\usepackage{etoolbox}
\usepackage{placeins}

\newcolumntype{L}[1]{>{\raggedright\arraybackslash}p{#1}}
\newcolumntype{C}[1]{>{\centering\arraybackslash}p{#1}}
\newcolumntype{R}[1]{>{\raggedleft\arraybackslash}p{#1}}
\newcommand{\beautytable}{\footnotesize\renewcommand{\arraystretch}{1.16}\setlength{\tabcolsep}{4pt}}
\newcommand{\beautytablecompact}{\scriptsize\renewcommand{\arraystretch}{1.12}\setlength{\tabcolsep}{4pt}}

\usepackage[hidelinks]{hyperref}
\usepackage{apacite}
\let\cite\shortcite
\let\citeA\shortciteA

\newif\ifcleanpublication
\ifdefined\publicationclean
  \cleanpublicationtrue
\else
  \cleanpublicationfalse
\fi
\ifcleanpublication
  \colorlet{blue}{black}
  \colorlet{red}{black}
\fi

\newcommand{\bl}[1]{\textcolor{blue}{{#1}}}

\newcommand{\Remm}[1]{}

\numberwithin{equation}{section}

\definecolor{MyGray}{rgb}{0.92,0.92,0.92}


\begin{document}

	\author{
Ronald Richman\footnote{insureai and Bayes Business School, Johannesburg, South Africa; ron@insureai.co}
}

\date{Version of May 1, 2026}
\title{Scaling Laws, Tabular Data and Actuarial Ratemaking Models}
\maketitle

\begin{abstract}
	\noindent
	Scaling laws in modern deep learning describe how held-out loss improves as model capacity, training data, and compute increase, often following power-law trends. We investigate whether analogous scaling regularities arise in actuarial ratemaking, where data are tabular, heterogeneous, and noisy, and where classical models such as GLMs remain strong baselines. Using a real-world motor insurance portfolio, we train models from different families across increasing fractions of the training data and multiple random seeds, evaluating out-of-sample \bl{Poisson deviance, a likelihood-based loss for Poisson count predictions in which lower values indicate better held-out fit}. We find that all model families improve with additional data, but scaling exponents differ substantially: TabM exhibits markedly stronger data scaling than purely supervised tabular Transformers and standard MLP baselines. Transformer variants show weak parameter scaling unless augmented with additional inductive biases (TabM-style adaptation or self-supervision). \bl{These results provide quantitative guidance on model selection by data regime and suggest that effective scaling on actuarial tabular tasks depends on architecture and loss function objective design, with simple increases in Transformer size providing limited gains.}
	\\[2mm]
\end{abstract}

\section{Introduction}

A natural actuarial intuition is that model performance improves as the amount of credible data available for model fitting increases. More data allows for better estimation of model parameters, leading to more accurate predictions. This relationship is particularly evident in actuarial ratemaking, where large and diverse datasets enable the construction of more reliable pricing models. Of course, fundamental limitations exist - highly volatile claims in certain lines of business or rapidly changing market trends may constrain the benefits of additional data. Nevertheless, this intuition aligns with practical experience across both reserving and pricing applications. A related intuition concerns the marginal returns to additional data. For traditional actuarial models used in ratemaking, there appears to be a ``saturation" point beyond which additional data yields negligible improvements in model performance. Machine learning and deep learning approaches, by contrast, may continue to benefit from additional data well beyond this saturation threshold. Put differently, the point at which diminishing returns set in may depend fundamentally on the model class being employed. A final intuition is that models often benefit from additional complexity that captures the structure and nuance of a problem, however, beyond a point, additional complexity may yield models that fail to perform well, besides for other disadvantages, such as the difficulty of maintaining highly complex models. 

The objective of this paper is to quantify these intuitions and connect them to the field of \textit{scaling laws} in the machine learning literature. This framework provides a principled approach to understanding and predicting how model performance evolves with key resources. In this paper, we apply this framework to actuarial ratemaking models, leveraging recent advances in tabular data and machine learning to develop a principled approach to understanding and predicting how model performance evolves with key modelling choices and with data availability.

The concept of scaling laws has gained prominence through work on large language models (LLMs). \citeA{Kaplan2020} and \citeA{Hoffmann2022} show that, over wide ranges, held-out loss (model performance) can often be described by power-law improvements in key resources. A common form of scaling laws is an irreducible loss plus additive power-law terms,

\begin{equation}\label{eq:scaling_law}
\bl{L(P, N, C) \approx L_{\infty} + A_P P^{-\alpha_P} + A_N N^{-\alpha_N} + A_C C^{-\alpha_C},}
\end{equation}

\bl{where $P$ is the number of model parameters, $N$ measures training data size, and $C$ measures training compute; $A_P$, $A_N$, and $A_C$ are fitted amplitude constants for the parameter, data, and compute terms; $L_{\infty}$ denotes the empirical performance floor; and $(\alpha_P,\alpha_N,\alpha_C)$ quantify marginal returns to scaling each resource.} Remarkably, these scaling laws have been shown to hold for a wide range of tasks and model classes, in \bl{particular}, the large-scale decoder Transformer models used in Large \bl{Language} Models (LLMs) \cite{Kaplan2020, Hoffmann2022}.

The question naturally arises: do analogous scaling regularities govern actuarial ratemaking models, despite the smaller data regime and the noisy, heterogeneous structure of actuarial \bl{ratemaking} covariates? We will review the scaling-law literature and the potential challenges for tabular actuarial tasks in Section~\ref{sec:literature}.

Despite these challenges, there are compelling reasons to expect that some form of scaling behavior might exist in actuarial modelling. The fundamental insight underlying scaling laws - that model performance is constrained by the interplay between model capacity, data volume, and computational resources - is not specific to language modelling. Indeed, the actuarial intuitions outlined above suggest precisely this. Moreover, classical statistical theory provides foundations for understanding how parameter estimation improves with sample size, and credibility theory in actuarial science has long formalized the relationship between data volume and the reliability of estimates. The question is whether these relationships can be unified and extended to encompass the full spectrum of models available to modern actuaries, from traditional GLMs to sophisticated deep learning architectures.

In this paper, we investigate scaling laws for actuarial ratemaking models empirically, using a large-scale experiment on a real-world motor insurance dataset. \bl{We focus specifically on claim frequency because claim counts are observed for every policy-period record, whereas severity is observed only for the smaller subset of records with claims; the frequency task is therefore better suited to scaling-law experiments across data, parameter, and compute regimes.} In our experiments, ``data size'' is operationalized by the number of policy-period records (rows) used for training (with corresponding total exposure-years as the actuarial analogue of ``example count''). We focus on two broad categories of models as representing the two ends of the model complexity spectrum: (i) statistical models, including generalized linear models (GLMs) and their regularized variants and (ii) deep learning models, including feed-forward neural networks (FFNs) and more advanced architectures such as Transformers\footnote{We note that intermediate approaches such as gradient boosting machines occupy a middle ground that we leave for future investigation.}. For each model class, we examine how out-of-sample predictive performance, measured by the Poisson deviance loss, varies as we systematically increase the amount of training data.

\bl{Poisson deviance is the natural metric for this frequency study because the response is a claim count and each prediction is an exposure-adjusted Poisson mean. In the exponential-family likelihood framework, deviance is twice the difference between a saturated log-likelihood and the fitted model log-likelihood, so reductions in mean Poisson deviance have a direct interpretation as average held-out log-likelihood gains per policy \cite{Wuthrich}. We therefore use Poisson deviance as the statistical loss for scaling-law estimation, while recognizing that translating such gains into premium adequacy, loss ratio, or profit requires additional pricing assumptions beyond the frequency model itself \cite{Israni2026DualEvaluation,Hedges2025LossRatio}.}

Our investigation is guided by several research questions. First, can we identify distinct \textit{data regimes} - small, medium, and large - in which different model classes are optimal? We hypothesize that GLMs may be optimal in small data regimes (e.g., fewer than 100,000 exposure years) and that deep learning models may be optimal in large data regimes where their ability to scale model capacity with data becomes advantageous. Second, at what data sizes do the returns from adding more data begin to diminish for each model class? Understanding these saturation points has important practical implications for determining how to select data for building actuarial models, as well as the selection of model classes \bl{depending} on the amount of data available. Third, can we derive explicit power-law relationships analogous to those found for LLMs, and if so, what are the relevant exponents? Finally, how are these relationships affected by the characteristics of the modelling problem, such as the number of rating factors, the signal-to-noise ratio, and the presence of time trends? As we shall see, addressing these questions also leads us to examine why certain architectures, particularly Transformers, may fail to scale in the actuarial context, and to develop some novel solutions.

Answering these questions requires careful experimental design. A key challenge is that, unlike LLMs where scaling experiments can be conducted by simply sampling more data from vast text corpora, actuarial datasets are finite and their characteristics are fixed. We address this by sampling progressively larger subsets from our available data, ensuring that the samples are representative of the full dataset. We also control for confounding factors such as the number of covariates and the complexity of the rating structure.

The contributions of this paper are fourfold. First, we provide the first systematic empirical investigation of scaling laws for actuarial ratemaking models, establishing whether and how performance scales with data size across different model classes. Second, we develop practical guidance for actuaries on model selection, identifying the data regimes in which each model class is likely to be optimal. Third, we derive explicit functional forms for the scaling relationships we observe, enabling actuaries to make predictions about model performance as a function of available data. Fourth, in the course of our experiments, we diagnose some problems that prevent scaling in Transformer models applied to ratemaking and, in response, develop several new models for \bl{ratemaking}.

Beyond methodological contributions, our findings carry significant implications for actuarial practice and regulatory governance. As insurers increasingly adopt sophisticated predictive models, regulators have intensified scrutiny of these techniques, raising questions about appropriate model complexity, data requirements, and validation standards. Quantifiable scaling relationships between data volume and model performance could provide principled justification for methodology choices, informing both internal model governance frameworks and regulatory discussions. For instance, if we can establish that a given model class requires a \bl{minimum training size $N_{\min}$, measured in policy-period rows or exposure-years,} to achieve reliable out-of-sample performance, this threshold becomes a defensible criterion for model selection and validation. Similarly, understanding the rate at which performance improves with additional data, characterized by a scaling exponent, enables insurers to make informed decisions about investments in data collection versus model development, optimizing resource allocation in predictive modeling efforts.

\subsection{Credibility as an actuarial scaling law}
\label{sec:intro-credibility-scaling}

Credibility theory can be viewed as an early actuarial scaling law. In the B\"uhlmann--Straub framework \cite{BuhlmannStraub1970}, the credibility factor $Z(n)$ increases with exposure volume $n$ as
\begin{equation}
Z(n) = \frac{n}{n + k},
\end{equation}
where $k$ is the within- versus between-risk variance ratio (the ``credibility constant''). Rather than $Z$ itself, consider the \emph{distance to full credibility},
\begin{equation}
1-Z(n)=\frac{k}{n+k}.
\end{equation}
In the large-exposure regime ($n\gg k$), this becomes $1-Z(n)\approx k/n \propto n^{-1}$: credibility implies a power-law decay of residual uncertainty once exposure substantially exceeds $k$.
\bl{The useful distinction between classical credibility and modern neural scaling laws is that modern scaling-law work estimates the rate of improvement empirically. In a relationship such as $L(N)-L_\infty \approx A_NN^{-\alpha}$, the fitted exponent $\alpha$ is an effective, experiment-specific summary of how quickly reducible loss falls as data increase; it can vary by model class, task, metric, and data regime.}

Our empirical scaling analysis uses an analogous ``distance to asymptote'' framing. Let $L(N)$ denote test Poisson deviance for a model trained on $N$ rows (with total exposure-years as the actuarial analogue of example count), and let $L_\infty$ denote an empirically fitted performance floor. Define the \emph{reducible loss}
\begin{equation}
U(N) := L(N) - L_\infty.
\end{equation}
\bl{Equivalently, the empirical fits use $U(N)\approx A_N N^{-\alpha}$, with $A_N$ recording the fitted distance-to-floor scale under the chosen row-count convention.} Under this lens, $\alpha$ acts as a portfolio-level ``credibility speed'' parameter: larger $\alpha$ means performance approaches the fitted floor faster as data grow.

This suggests a credibility-style curve derived directly from scaling fits. Fix a baseline $N_0$ (e.g., the smallest training fraction) and define
\begin{equation}
Z_{\text{scale}}(N) := 1-\frac{U(N)}{U(N_0)} = 1-\frac{L(N)-L_\infty}{L(N_0)-L_\infty}.
\end{equation}
Under the power law, $Z_{\text{scale}}(N)\approx 1-(N/N_0)^{-\alpha}$, which increases from 0 at $N_0$ and approaches 1 as $N\to\infty$. Likewise, a ``minimum credible data'' requirement can be expressed by selecting a tolerance $\delta$ and solving $L(N)\le L_\infty+\delta$, yielding \bl{$N \ge (A_N/\delta)^{1/\alpha}$}, directly analogous to solving $n = k Z/(1-Z)$ in classical credibility.
Finally, a simple interpolation between B\"uhlmann--Straub and power-law scaling is the generalized curve $Z_{\alpha}(n):=n^\alpha/(n^\alpha+k^\alpha)$, which recovers the classical formula at $\alpha=1$ and satisfies $1-Z_{\alpha}(n)\sim (k/n)^\alpha$ for $n\gg k$.

More broadly, actuaries already encounter many scaling relationships - for example, standard errors for Poisson rates and GLM coefficients shrink as exposure or sample size to the $-1/2$, and \bl{Monte Carlo error decays with the number of simulation draws at the same $-1/2$ rate}. Our contribution is to make such scaling explicit for modern tabular model families, estimate the corresponding exponents empirically, and connect them to practical questions about when models become ``credible'' at portfolio scale.

\subsection{Organization of the Manuscript}

With these intuitions and their connection to classical credibility in mind, we now outline how the paper is structured.

One theme that shapes the structure of the paper is the behavior of Transformer models. In contrast to the monotonic parameter scaling often reported for LLM-scale Transformers, we find that straightforward supervised tabular Transformers exhibit weak and sometimes non-monotone returns to additional parameters on our ratemaking task. For this reason we study a \emph{sequence} of Transformer variants rather than a single baseline: we progressively modernize the architecture (pooling, stabilization and value-side capacity mechanisms, and mixture-of-experts routing) and then add a swap-style self-supervised learning (SSL) objective that provides additional training signal. This progression motivates the detailed Transformer exposition in Sections~\ref{sec:model_descriptions} and~\ref{sec:results}, and the Results are organized to mirror this ``from baselines to scaling fixes'' narrative.

The remainder of this paper is organized as follows. \bl{Section~\ref{sec:rationale_explainer} provides a non-technical rationale, metric explanation, and reading guide for pricing actuaries.} Section~\ref{sec:literature} reviews scaling laws in modern deep learning and the relevant literature on tabular insurance modeling. Section~\ref{Data} describes the dataset and descriptive statistics. Section~\ref{sec:methodology} describes our experimental methodology. Section~\ref{sec:model_descriptions} summarizes the model families studied. Section~\ref{sec:results} presents empirical scaling results, and Section~\ref{sec:discussion} concludes.

\bl{Appendix~\ref{app:math_primer} provides a compact notation reference for the symbols used in Section~\ref{sec:rationale_explainer} and later technical sections.} Appendix~\ref{app:model_configs} provides technical detail and the component-level formulae used across model families. \bl{Appendix~\ref{app:chinchilla_explainer} gives a concise technical explainer for the Chinchilla-style allocation result used in the scaling discussion. A separate supplementary material provides an open-source, simpler implementation of the main workflow on public French MTPL (FMTPL) frequency data, intended to make the empirical pipeline easier to inspect and reproduce on non-confidential data. The data are from \texttt{CASdatasets} by Dutang and Charpentier, and we use the corrected version documented by W\"uthrich and Merz \cite{CASdatasetsFreMTPL,WuethrichMerz2023}.\footnote{The accompanying repository is available at \url{https://github.com/RonRichman/frmtpl-scaling-laws}.}}

\section{\texorpdfstring{\bl{Rationale, Metric, and Reading Guide}}{Rationale, Metric, and Reading Guide}}
\label{sec:rationale_explainer}
\begingroup\color{blue}

This section is intended as a reader's guide for pricing actuaries who want the main motivation, notation, and interpretation before the technical details. Readers who are already comfortable with Poisson frequency modelling, neural-network notation, and scaling-law terminology can skip to the literature review. Appendix~\ref{app:math_primer} keeps the notation tables in one place for reference.

\subsection{Why scaling laws matter for actuarial pricing}

Large language models made scaling laws prominent because their held-out loss often improves smoothly as model size, training data, and compute increase. \citeA{Kaplan2020} illustrate this idea by plotting test loss against compute, dataset size, and parameter count on log--log axes: over the fitted regime, the points form smooth downward trends that can be summarized by power laws. The important visual message is that loss improvement is not erratic; it follows a predictable diminishing-returns curve. Figure~\ref{fig:section2-best-data-scaling} below gives a small actuarial analogue of this idea for the best-achieved TabM/TabM-mini data-scaling fit in our experiment.

The actuarial question is whether a similar empirical regularity appears in ratemaking. Actuaries already reason about credibility, data volume, model stability, and diminishing returns. A scaling-law analysis puts numbers on these ideas. Instead of asking only whether a model is better on one dataset size, we ask how fast its held-out loss improves as the portfolio grows and whether extra model capacity still converts into better predictions.

\subsection{The frequency task in actuarial terms}

Each observation in the experiment is a policy-period row. We write $y_i$ for the observed claim count and $E_i$ for earned exposure. The model predicts a claim frequency rate, and exposure converts that rate into an expected claim count. In symbols, the neural network or GLM produces a log-rate
\begin{equation}
\eta_i=f_\theta(x_i),
\end{equation}
where $x_i$ contains the rating factors and $\theta$ denotes the trainable model parameters. The predicted claim rate is $\hat{r}_i=\exp(\eta_i)$ and the predicted mean claim count is
\begin{equation}
\hat{\lambda}_i = E_i\hat{r}_i = E_i\exp(\eta_i).
\end{equation}
Equivalently, $\log\hat{\lambda}_i=\log E_i+\eta_i$, which is the familiar Poisson offset form. This matters because our models are predicting the claim-frequency component, rather than modelling severity conditional on a claim or modelling pure premium directly. Frequency is observed for every policy-period record, while severity is observed only for the smaller subset with claims; this is why frequency is the more natural target for the scaling-law experiment in this paper.

\subsection{Why Poisson deviance is the loss}

Poisson deviance is a likelihood-based loss for count-frequency models. For a policy-period row with observed count $y_i$ and predicted mean count $\hat{\lambda}_i$, the per-row deviance is
\begin{equation}
D_{\mathrm{Pois}}(y_i,\hat{\lambda}_i)
=2\left[
y_i\log\left(\frac{y_i}{\hat{\lambda}_i}\right)-(y_i-\hat{\lambda}_i)
\right],
\end{equation}
with the usual convention that the first term is zero when $y_i=0$. Lower deviance means the model assigns higher Poisson likelihood to the observed claim counts on the held-out data.

\bl{For interpretation, aggregate Poisson deviance is twice the saturated log-likelihood minus twice the fitted model log-likelihood. Therefore, if one model improves mean deviance by $\Delta$, this is an average held-out log-likelihood lift of $\Delta/2$ nats per policy-period row, where a nat is one unit of log-likelihood on the natural-logarithm scale.} This is a statistical fit statement about emerging claim counts. Translating it into premium adequacy, loss ratio, or profit requires a pricing layer with severity or pure-premium modelling, calibration and bias controls, margins and expenses, and competitive or retention effects.

\subsection{The three resources: data, parameters, and compute}

The paper studies the three ``resources'' available to a pricing actuary when building a model. First, $N$ is the amount of labelled training data, i.e., the policy records with observed exposure and claims, measured as policy-period rows, with exposure-years as the actuarial analogue of example volume. Second, $P$ is the number of trainable model parameters. For a GLM, these are coefficients; for a neural network, they include embedding tables, dense weights, attention weights, and prediction heads. Third, $C$ is a training compute proxy. In this paper, $C$ is measured per epoch and, for ensemble results, scaled to match the five trained seeds used to produce the seed-averaged prediction.

We refer to the coefficients of these laws as indexed amplitude constants:
\begin{align}
L(N) &\approx L_\infty + A_N N^{-\alpha} \qquad \text{(data scaling)},\\
L(P) &\approx L_\infty + A_P P^{-\beta} \qquad \text{(parameter scaling)},\\
L(C) &\approx L_\infty + A_C C^{-\gamma} \qquad \text{(compute scaling)}.
\end{align}
The fitted exponents in these laws have a direct reading. The data exponent $\alpha$ says how quickly loss improves as more rows are added. The parameter exponent $\beta$ says how effectively a model family converts extra capacity into lower test deviance. The compute exponent $\gamma$ says how fast the best observed frontier improves as compute increases. A small exponent still means improvement, but with strong diminishing returns.

Here $L_\infty$ is an empirical fitted floor over the range of experiments, rather than a universal Bayes-risk claim. After subtracting the floor, the exponent is the slope on a log--log plot. For example, doubling $N$ multiplies the remaining reducible loss by $2^{-\alpha}$.

\subsection{A short example and how to read the scaling claims}

A typical data-scaling law in this paper has the form
\begin{equation}
L(N)-L_{\infty}\approx A_N N^{-\alpha},
\end{equation}
\bl{where $L(N)$ is held-out Poisson deviance at training size $N$, $L_{\infty}$ is the fitted floor, and $A_N N^{-\alpha}$ is the remaining reducible loss.} As a concrete preview, Figure~\ref{fig:section2-best-data-scaling} shows the best-achieved TabM/TabM-mini data-scaling envelope that is analyzed later in Section~\ref{sec:results-scaling-laws}. Using $N$ measured in policy-period rows, the fitted constants for this preview are
\begin{equation}
L(N)-0.28749 \approx 1.03\,N^{-0.409}.
\end{equation}
Thus, doubling the training rows multiplies the reducible loss by $2^{-0.409}\approx 0.75$. In words, each doubling of data removes about 25\% of the remaining gap to the fitted floor, within the experimental range and model family being studied.

The same logic applies to parameter and compute scaling. If a family has a larger $\alpha$, its held-out deviance approaches the fitted floor faster as labelled data are added. If a family has a near-zero $\beta$, adding parameters mostly adds complexity without producing much held-out deviance gain under the training recipe used here. These exponents summarize empirical curves, not a single train/test comparison. The claims are conditional on the dataset, frequency target, feature set, tuning protocol, and IID train/test split used here. The fitted floor $L_\infty$ is an empirical anchor for the observed range of models and training sizes, rather than a literal irreducible error for all possible ratemaking models or future data.

A related question is how to allocate a fixed training budget between labelled data and model capacity. Later in the paper we use a standard compute-efficient scaling argument for this purpose. The result should be read as a heuristic: we measure training cost approximately as the cost of one pass through the training data, rather than exact total training cost, and labelled insurance rows are not created on demand. Additional rows require written policies, exposure, and claims emergence. In practice, once labelled data are fixed or slow to grow, the more immediate levers are model capacity, training recipe, and representation learning, including self-supervised use of policy or quote records without emerged claims.

The main text gives enough model notation to read the empirical comparison. Appendix~\ref{app:math_primer} collects notation, and Appendix~\ref{app:component_formulae} records the component-level formulae for readers who want implementation-level detail.

\begin{figure}[!htbp]
\centering
\IfFileExists{figures/section2_best_data_scaling.pdf}{%
\includegraphics[width=0.88\textwidth]{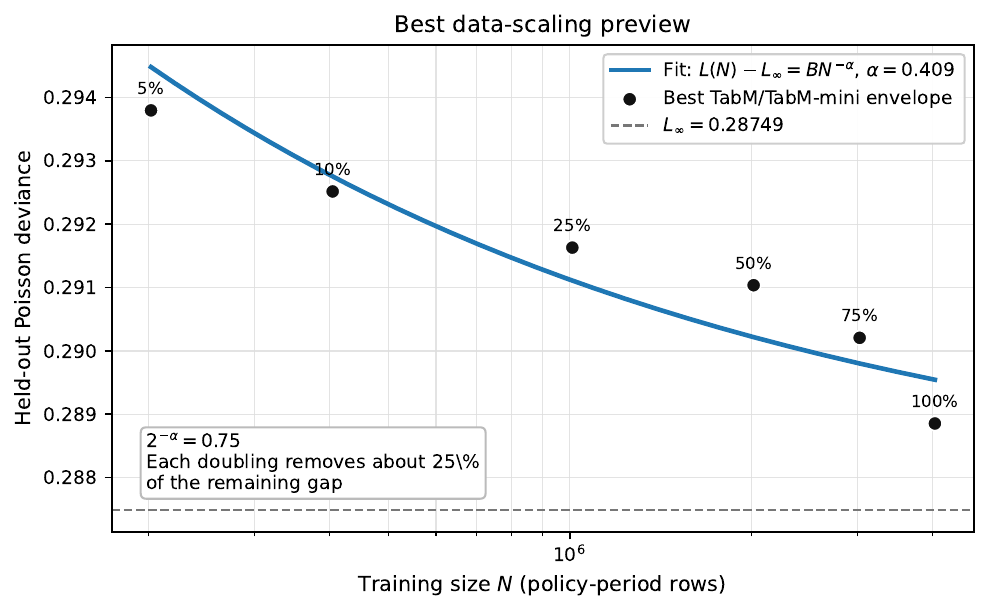}%
}{%
\fbox{\parbox{0.9\textwidth}{\centering Missing preview figure: \texttt{figures/section2\_best\_data\_scaling.pdf}}}%
}%
\caption{\bl{Standalone preview of the best-achieved data-scaling law. Points are the best TabM/TabM-mini seed-averaged ensemble deviance at each nested training fraction, including the full-data-only TabM-mini extension at 100\%. The fitted curve uses the same floor and exponent convention as Section~\ref{sec:results-scaling-laws}.}}
\label{fig:section2-best-data-scaling}
\end{figure}

\FloatBarrier
\subsection{How to read the model families}

\bl{Figure~\ref{fig:section2-model-family-map} gives a compact map of how the model families used in this paper relate to one another.} A useful way to read the model families is to ask how much of the rating structure is specified by the modeller and how much is learned from data. Consider a policy-period row with driver age, vehicle value, territory, and prior-claims band. A GLM is closest to a traditional tariff: it estimates rating-factor effects and combines them on the log-frequency scale, with interactions added when the modeller specifies them. The neural models use the same rating-factor inputs but give the model more freedom to learn nonlinear effects and interactions.

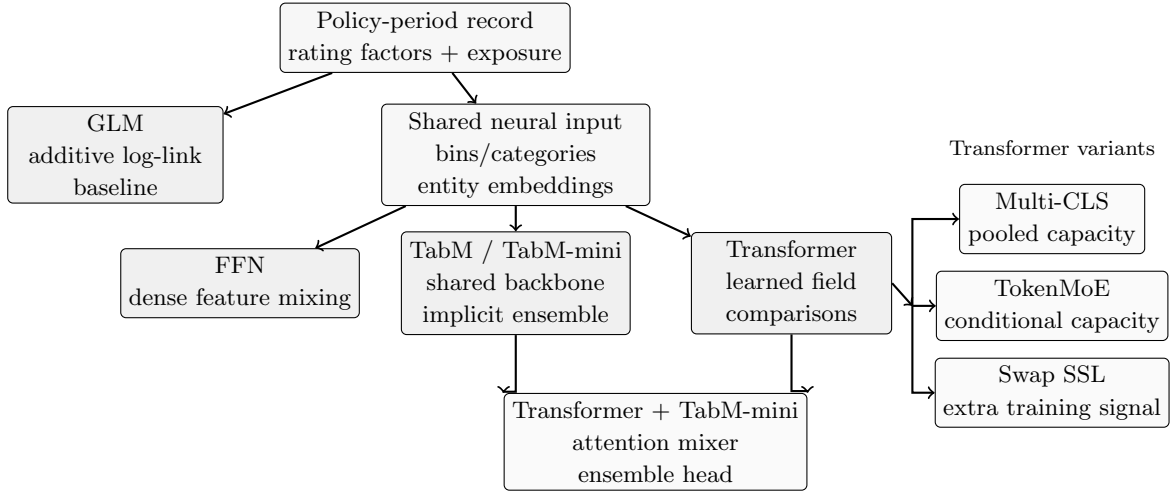
\begin{figure}[!htbp]
\centering
\begingroup\color{blue}
\begin{adjustbox}{max width=\textwidth}
\begin{tikzpicture}[
  font=\footnotesize,
  box/.style={draw=blue, rounded corners=2pt, align=center, inner sep=3pt, minimum height=0.74cm, text=blue, fill=blue!3},
  family/.style={box, fill=blue!6, minimum width=2.65cm},
  variant/.style={box, fill=blue!2, minimum width=2.45cm},
  arrow/.style={->, thick, draw=blue}
]
\node[box, minimum width=3.65cm] (data) at (0,0) {Policy-period record\\rating factors + exposure};
\node[family, minimum width=2.85cm] (glm) at (-4.1,-1.55) {GLM\\additive log-link\\baseline};
\node[box, minimum width=3.55cm] (embed) at (1.2,-1.55) {Shared neural input\\bins/categories\\entity embeddings};
\node[family] (ffn) at (-2.45,-3.25) {FFN\\dense feature mixing};
\node[family] (tabm) at (1.2,-3.25) {TabM / TabM-mini\\shared backbone\\implicit ensemble};
\node[family] (tr) at (4.85,-3.25) {Transformer\\learned field\\comparisons};
\node[variant, minimum width=3.2cm] (hybrid) at (3.05,-5.35) {Transformer + TabM-mini\\attention mixer\\ensemble head};
\node[text=blue, font=\scriptsize] (vlabel) at (8.3,-1.45) {Transformer variants};
\node[variant] (mcls) at (8.3,-2.4) {Multi-CLS\\pooled capacity};
\node[variant] (moe) at (8.3,-3.55) {TokenMoE\\conditional capacity};
\node[variant] (ssl) at (8.3,-4.7) {Swap SSL\\extra training signal};
\coordinate (trfork) at (6.45,-3.55);

\draw[arrow] (data) -- (glm);
\draw[arrow] (data) -- (embed);
\draw[arrow] (embed) -- (ffn);
\draw[arrow] (embed) -- (tabm);
\draw[arrow] (embed) -- (tr);
\draw[arrow] (tabm.south) -- ++(0,-0.75) -| (hybrid.north west);
\draw[arrow] (tr.south) -- ++(0,-0.75) -| (hybrid.north east);
\draw[arrow] (tr.east) -- (trfork);
\draw[arrow] (trfork) |- (mcls.west);
\draw[arrow] (trfork) -- (moe.west);
\draw[arrow] (trfork) |- (ssl.west);
\end{tikzpicture}
\end{adjustbox}
\endgroup
\caption{\bl{Conceptual map of the model families studied in the paper. The GLM provides the additive actuarial baseline. The neural families share the same tabular input representation, then differ in how they combine rating-factor information: dense layers for FFNs, shared ensemble-style adapters for TabM/TabM-mini, learned field-comparison blocks for Transformers, and a hybrid Transformer-plus-TabM head for Transformer+TabM-mini.}}
\label{fig:section2-model-family-map}
\end{figure}

\begin{itemize}
  \item \textbf{GLM}: an additive benchmark. It estimates relativities for rating factors under a log-link Poisson frequency model. It is stable and interpretable, but interactions usually need to be specified manually.
  \item \textbf{FFN}: a feed-forward neural network. It first converts rating factors into numerical representations and then passes them through dense layers. This lets the model learn nonlinear combinations, such as territory mattering more for certain vehicle or prior-claims profiles.
  \item \textbf{Transformer}: a neural network that repeatedly compares fields within the same policy record before making a prediction. In the technical sections, the internal field representations are called \emph{tokens} and the learned comparison mechanism is called \emph{attention}. For the reader's guide, the main point is simpler: the model can learn which rating factors should be read together, rather than relying only on a pre-specified interaction list.
  \item \textbf{TabM and TabM-mini}: compact ensemble-style neural models. They share most weights but allow several lightweight member-specific views of the same row, aiming to capture some of the robustness of an ensemble without training many fully separate networks.
  \item \textbf{Transformer+TabM-mini}: a hybrid model. It uses Transformer-style field comparisons first and then a TabM-mini-style prediction head, so it combines learned interactions with compact ensemble-style stabilization.
\end{itemize}

This paper treats model complexity as an empirical question. Added flexibility is useful in some regimes and wasteful in others. In small data regimes, strong regularization and simple structure can dominate. In larger data regimes, additional capacity can become valuable because the model has enough observations to learn genuine structure rather than mainly fitting noise.

\endgroup

\section{Literature Review}
\label{sec:literature}

This section reviews two research threads that motivate our study. The first concerns \emph{neural scaling laws}, which characterize how generalization performance improves as a function of computational budget, model capacity, and data volume. The second concerns \emph{deep learning for actuarial work}, where predictive modeling is predominantly tabular, the signal-to-noise ratio is often low, and interpretability and governance constraints shape feasible model classes and evaluation criteria.

\subsection{Scaling Laws in Modern Deep Learning}
\label{sec:lit_scaling}

\subsubsection{Empirical Regularities and Power-Law Improvement}

Transformer-based large language models (LLMs) have substantially advanced natural language processing and catalyzed renewed interest in empirical scaling behavior \cite{Vaswani}. A key finding in this domain has been the discovery of \emph{scaling laws} that govern the performance of these models. These laws demonstrate that model performance can be predicted as a function of three principal factors: model size (number of parameters $P$), dataset size $D$ (measured in training tokens, or tabular examples in our setting), and computational resources $C$ (typically measured in FLOPs) \cite{Kaplan2020}.

Early systematic evidence across diverse tasks showed approximate power-law relationships between error and resources, implying that progress can often be forecast from smaller experiments \cite{Hestness}. These findings helped shift model development from architecture-centric trial-and-error to a more engineering-oriented perspective in which scale is treated as a first-class design variable. Specifically, the influential scaling law analysis of \citeA{Kaplan2020} reports that cross-entropy test loss $L$ follows remarkably stable power laws across orders of magnitude in model parameters, training data tokens, and training compute. A typical univariate form is
\begin{equation}
L(P) \approx A \, P^{-\alpha} + L_{\infty},
\end{equation}
with analogous expressions for $D$ and $C$, where $L_{\infty}$ denotes an irreducible loss floor under the chosen data distribution and evaluation metric. Empirically, \citeA{Kaplan2020} report remarkably stable power-law trends for cross-entropy test loss across orders of magnitude in parameters, tokens, and compute: $L \propto P^{-0.076}$, $L \propto D^{-0.095}$, and $L \propto C^{-0.050}$ in their setting. These relationships support the use of scaling laws as \emph{predictive tools} for estimating marginal returns to additional model capacity, data, or compute.

\subsubsection{Compute-Optimal Training and Data--Parameter Balance}

Subsequent work refined the interpretation of scaling laws by emphasizing compute-optimal training. \citeA{Hoffmann2022} demonstrate that many pre-2022 flagship language models were \emph{undertrained} (too few tokens per parameter) relative to their compute budgets, and that substantially better performance can be obtained by allocating compute to both parameters and data more evenly. Their analysis can be summarized by a bivariate scaling law of the form
\begin{equation}
\widehat{L}(P,D) = \frac{A}{P^{\alpha}} + \frac{B}{D^{\beta}} + E,
\end{equation}
where the constants $A, B, E$ are fit empirically and the exponents $(\alpha, \beta)$ quantify the marginal returns to parameters and data \cite{Hoffmann2022}. \bl{This displayed equation follows the notation of \citeA{Hoffmann2022}; in our empirical sections we use indexed amplitudes such as $A_N$, $A_P$, and $A_C$ so that fitted constants are not confused with data size or parameter count.} A central practical implication is that, for a fixed compute budget, there exists an \emph{efficient frontier} for allocating resources between model size and training data volume, which can change recommended training recipes and deployment economics.

\subsubsection{Estimating and Interpreting Scaling Laws}

Because scaling studies require many training runs, a parallel line of work studies how to \emph{estimate} scaling laws efficiently and robustly. \citeA{Choshen} provide methodological guidance for fitting and validating scaling laws, including the use of intermediate checkpoints to improve estimation accuracy and the importance of accounting for variability across random seeds. They find that fitting to intermediate checkpoints during training runs substantially improves accuracy compared to using only fully trained models, and that estimates are generally most accurate when derived from models of similar sizes, though training multiple small models can sometimes provide better estimates than a single large one due to variance across seeds. These practices are directly relevant when scaling-law fits are used as decision tools for choosing architectures, data pipelines, or training budgets. In this work, we \bl{train} all models five times and use the average predictions to evaluate losses and scaling laws.

\subsubsection{Beyond Simple Power Laws: Regime Changes and Emergent Capabilities}

While power laws offer strong first-order regularities, they do not preclude regime-dependent behavior. In supervised learning, ``double descent'' and related phenomena highlight that test error may be non-monotone in model capacity in finite-sample regimes \cite{Belkin2019}. More broadly, \citeA{Caballero2022} formalize \emph{broken} or piecewise scaling behavior in which different scaling regimes arise across ranges of compute or model size, motivating caution when extrapolating beyond observed scales.

A distinct debate concerns the apparent emergence of qualitative capabilities in LLMs. \citeA{Wei} document ``emergent abilities'' on certain benchmarks as model scale increases, identifying capabilities that appear only after certain parameter thresholds are crossed. However, \citeA{Schaeffer2023} argue that many apparent discontinuities are artifacts of metric choice and thresholded evaluation, and that alternative continuous metrics often reveal smoother trends. Together, these studies suggest that scaling relationships can be reliable in aggregate while still being sensitive to evaluation design, a point that becomes especially salient when translating scaling perspectives to tabular actuarial tasks where the signal-to-noise ratio is lower, the data are heterogeneous and many relatively powerful model forms are available.

The discovery of these scaling laws has fundamentally changed how researchers approach model development, enabling principled decisions about resource allocation and performance expectations. Rather than focusing solely on architectural innovations, researchers can now make data-driven decisions about whether to increase model size, gather more training data, or allocate additional computational resources based on quantifiable performance expectations. The strategic significance of scaling has been summarized in broader reflections that emphasize the long-run dominance of general methods that scale with compute, relative to methods that encode domain knowledge explicitly \cite{Sutton2019}. In actuarial contexts, this raises a tension between scale-driven performance gains and institutional requirements for interpretability, stability, and governance.

However, the empirical regularities above were established in high-signal, massive-data regimes with homogeneous tokenized inputs; the extent to which similar scaling behavior holds for noisy, heterogeneous, regulated actuarial tasks remains unclear, which we discuss next.

\subsection{Deep Learning for Actuarial Work and Tabular Insurance Data}
\label{sec:lit_actuarial}

\subsubsection{Tabular Learning and the Persistence of Tree-Based Methods}

Most actuarial prediction problems, including pricing, lapse, fraud detection, and transaction-level reserving, can easily be structured as supervised learning on heterogeneous tabular features - see \citeA{Richman2021a} for some formulations of actuarial problems in this way. While traditional actuarial practice has relied primarily on Generalized Linear Models (GLMs) for pricing and chain-ladder or Bornhuetter--Ferguson methods for reserving, a growing body of research has explored more sophisticated approaches \cite{Harris}.

In the tabular machine learning setting, gradient-boosted decision trees (GBDTs), notably XGBoost, LightGBM, and CatBoost, are strong and widely adopted baselines \cite{Chen2016XGBoost, Ke2017LightGBM, Prokhorenkova2018CatBoost}. The continued competitiveness of tree ensembles has been documented in systematic benchmark studies. For ``typical'' medium-sized tabular datasets, \citeA{Grinsztajn2022} find that tree-based methods often outperform standard neural networks, attributing the gap to differences in inductive biases (e.g., handling of uninformative features and irregular decision boundaries). Complementing this, \citeA{McElfresh2023} provide a large-scale benchmark and conclude that performance differences are frequently modest and dataset-dependent, and that careful tuning and preprocessing can matter as much as algorithm choice.

These results are particularly relevant for insurance data, which often exhibits skewness, heavy tails, sparse categorical structure, and low signal-to-noise ratios \cite{Richman2022, WuthrichZiegel2024}. Such properties can reduce the marginal gains from increasing model capacity, potentially pushing actuarial problems into regimes where classical or tree-based methods remain competitive.

A central practical challenge in actuarial modeling is the prevalence of many categorical covariates (e.g., occupation, industry codes, postcode, vehicle model). A foundational approach in tabular deep learning is the use of \emph{entity embeddings} to learn dense representations of categories jointly with the predictive model \cite{Guo}. Embeddings can improve statistical efficiency relative to one-hot encoding and enable neural models to share information across related categories. The first papers to implement entity embedding in actuarial modelling are \citeA{Richman2021a, Richman2021b} and \citeA{Kuo2019}.

\subsubsection{Tabular Transformers: Adapting Attention to Structured Features}

Several architectures adapt Transformers to tabular data by treating features as tokens and learning feature interactions through self-attention. TabTransformer applies contextual attention to categorical variables to produce context-aware embeddings \cite{Huang2020TabTransformer}. FT-Transformer generalizes this approach by tokenizing both numerical and categorical features and applying Transformer blocks to the resulting sequence \cite{Gorishniy2021}. SAINT introduces additional mechanisms such as inter-sample attention and contrastive pretraining to improve generalization \cite{Somepalli2021SAINT}. Across benchmarks, these models can be competitive with GBDTs, but the literature does not identify a universally dominant deep architecture for tabular data \cite{Grinsztajn2022, McElfresh2023}. In insurance settings, attention can also be challenged by sparse categorical structure, weak spatial or sequential priors, and low signal-to-noise ratios, suggesting that additional inductive bias or auxiliary training signal may be needed for effective scaling (see Section~\ref{sec:results-why}).

\subsubsection{Tabular Foundation Models and In-Context Learning}

A recent direction aims to transfer the ``foundation model'' paradigm to tabular learning. Prior-data fitted networks (PFNs) pretrain Transformers on distributions over synthetic datasets to approximate Bayesian inference at test time \cite{Muller2021}. Building on this idea, TabPFN treats supervised classification on small tabular datasets as an in-context learning problem, processing the training set as context and producing predictions in a single forward pass \cite{Hollmann2023TabPFN}. A subsequent study reports that tabular foundation models can yield highly accurate predictions in small-data regimes, effectively amortizing learning across many synthetic tasks \cite{Hollmann2025}. This line of work is notable for actuarial applications because many portfolios and niche coverages fall into the small-to-medium data regime; although much of the tabular foundation-model literature emphasizes classification and synthetic-task pretraining, the broader idea of amortizing learning across task distributions may be relevant when training data is scarce or fragmented.

\subsubsection{When Tabular ML Meets Millions of Rows}
\label{sec:millions}

A recurring challenge in tabular machine learning research is that many widely used academic benchmarks live in a \emph{small-to-medium} data regime, often on the order of $10^3$--$10^5$ rows.
This matters because, in small-data settings, performance comparisons can become dominated by variance across random seeds, hyperparameter sensitivity, and inductive biases that favor sample-efficient methods.
Consequently, conclusions drawn from benchmark leaderboards do not always transfer to industrial settings where tabular datasets commonly contain millions of rows and where training compute becomes an explicit constraint.

\bl{This reflects benchmark design: many suites are built to make cross-validation and hyperparameter search feasible on modest hardware.} For example, the OpenML-CC18 curated classification suite restricts datasets to between 500 and 100{,}000 observations \cite{bischl2021openml}. Recent efforts continue this emphasis. TabArena, a newly introduced ``living'' benchmark for tabular ML, explicitly initializes its v0.1 leaderboard in a small-to-medium regime and leaves large-data settings to future versions \cite{Erickson2025TabArena}. These resources are valuable for standardization and reproducibility, but they also highlight a gap: the literature contains comparatively fewer systematic studies of \emph{tabular scaling behavior} in the million-row regime, which is common in insurance applications and which is where we will illustrate our results. We summarize these figures in Table~\ref{tab:benchmark_comparison}.

\begin{table}[t]
\centering
\begingroup\color{blue}
\beautytable
\caption{Common tabular benchmarking regimes compared with the present scaling study.}
\label{tab:benchmark_comparison}
\begin{adjustbox}{max width=\textwidth}
\begin{tabular}{@{}L{3.3cm}C{2.8cm}L{4.8cm}L{3.4cm}@{}}
\toprule
\textbf{Benchmark} & \textbf{Typical training size $N$} & \textbf{Prediction task and metric} & \textbf{Primary purpose} \\
\midrule
OpenML-CC18 \cite{bischl2021openml}
& $5\times 10^2$--$10^5$ rows
& Supervised classification; evaluation metric varies by task and study
& Practical comparison across many public datasets \\
\addlinespace[0.25em]
TabArena v0.1 \cite{Erickson2025TabArena}
& $\sim 10^3$--$1.5\times 10^5$ rows
& IID classification and regression; leaderboard aggregation over dataset-specific metrics
& Living leaderboard with reproducible model comparisons \\
\addlinespace[0.25em]
This paper (motor ratemaking)
& $2\times 10^5$--$4\times 10^6$ policy-period rows
& Poisson claim frequency; held-out Poisson deviance
& Scaling-law estimation and compute trade-offs in a motor pricing task \\
\bottomrule
\end{tabular}
\end{adjustbox}
\endgroup
\end{table}

\bl{Here, we train on a single realistic ratemaking task with a full training set of approximately $4.5\times 10^6$ rows, using nested training fractions $f\in\{0.05,0.10,0.25,0.50,0.75,1.00\}$ and the corresponding training sizes $N_f$.} Thus, our smallest \bl{training subset} is larger than many of the benchmark datasets used to date in the tabular machine learning literature. This enables us to answer questions that are harder to investigate on small benchmarks: (i) whether losses improve smoothly with training size $N$ (rows) in the high-$N$ regime, (ii) how model-family rankings change as $N$ grows into the multi-million range, and (iii) whether compute-optimal model capacity increases with $N$ (as in compute-optimal scaling analyses for large Transformers).

To summarize data scaling, we will fit power laws of the form

\begin{equation}
L(N)\approx L_{\infty} + A_N\,N^{-\alpha},
\end{equation}

where $L(N)$ is held-out deviance at training size $N$, $L_{\infty}$ is an extrapolated empirical floor, $\alpha$ captures the slope of improvement with additional data, and \bl{$A_N$ is an amplitude term measuring distance to the asymptote at finite $N$}.\footnote{\bl{The numerical value of $A_N$ depends on how $N$ is measured (e.g., raw rows vs. exposure), so we interpret $A_N$ primarily for relative comparisons within a fixed experimental protocol.}}

\subsubsection{Actuarial Applications: Pricing, Reserving, and Model Governance}

Within actuarial science, the adoption of deep learning has been surveyed extensively by \citeA{Richman2021a, Richman2021b}, who outlines practical heuristics for when deep models are likely to outperform generalized linear models and other classical baselines. A key theme is that deep learning becomes more attractive as data volume increases and as feature interactions and nonlinearities become more material, but this must be weighed against interpretability, calibration, and model risk management constraints \cite{Richman2022}. Such governance constraints are not abstract: in practice, actuaries and regulators often require stability under small data refreshes, monotonicity constraints for individual predictions, well-formed uncertainty quantification, and compliance with fairness and protected-class regulations.

Interpretability-oriented neural designs are therefore prominent in the actuarial literature. LocalGLMnet learns feature-dependent coefficients while preserving a GLM-like decomposition, aiming to bridge predictive performance and explanatory structure \cite{LocalGLMnet}. Post-hoc recalibration and reliability control have also been studied as mechanisms to stabilize predictions in low-signal settings; for example, \citeA{WuthrichZiegel2024} discuss isotonic recalibration as a tool for improving probabilistic predictions.

Deep learning for claims reserving has developed in parallel, including neural methods that model loss development patterns and incorporate richer representations than classical triangles. Practical examples include neural reserving architectures proposed by \citeA{Gabrielli2020} and sequence-based methods such as DeepTriangle \cite{Kuo2019}. These studies highlight that reserving performance can be sensitive to data granularity, temporal nonstationarity, and the availability of claim-level features, all of which complicate direct transfer of scaling insights from language modeling.

To our knowledge, there is limited actuarial work that directly frames model development through scaling-law lenses; one example is research from \citeA{Richman2021}, in the context of mortality modeling and forecasting. In that work, a single large and deep neural network was fit to national and sub-national mortality data for a large number of countries. The relationship between model performance and network width was investigated as the size and depth of the networks were varied. The results present intriguing evidence that increasing the number of parameters (through increasing the width of the network layers) leads to increased out-of-sample performance, but only up to a limit. Comparing this situation to \citeA{Kaplan2020}, who increase the amount of data concurrently with the number of parameters, the mortality study held data constant, possibly explaining why the widest network decreased in performance. This preliminary evidence suggests that scaling laws for large-scale actuarial models can be discovered, while also highlighting important domain-specific constraints.

Importantly, we must also mention that not all model classes can be used in every jurisdiction for practical pricing and \bl{ratemaking}; in particular, in the United States, many state insurance regulators require models used for rating to be filed and large-scale deep learning (and other machine learning) models may not be acceptable to these regulators. Nonetheless, studying scaling laws remains valuable for several reasons: it informs model selection and capacity planning in jurisdictions where advanced models are permitted; it provides quantitative evidence that can shape evolving regulatory attitudes toward algorithmic pricing; and, even in constrained regulatory environments, understanding how performance scales with data volume can guide decisions about data collection, feature engineering, and the choice among permissible model classes. Finally, even if an insurer's rates are determined using simpler models, it can be useful to maintain internally models which can provide insight into the insurer's \bl{portfolio}, even if these are not actually used for pricing.

\subsubsection{Implications for Actuarial Scaling-Law Studies}

The literature collectively suggests a fundamental mismatch between the regimes in which LLM scaling laws were established and typical actuarial settings. Several important distinctions warrant careful investigation:

First, actuarial problems typically feature lower signal-to-noise ratios than language modeling tasks, as predicting infrequent insurance events like accidents involves substantial stochastic elements not present in next-token prediction for language. Second, actuarial data exhibits significant time trends and seasonality, while LLMs are usually trained on static corpora. Third, insurance datasets are dramatically smaller than those used for LLM training, both because each insurer has limited policyholders and because older data becomes less relevant for training actuarial models than for LLMs. Fourth, actuarial models must process heterogeneous input types requiring explicit variable selection, unlike the homogeneous token sequences processed by LLMs. Finally, traditional actuarial models employ relatively constrained parameter counts compared to the billions or trillions of parameters in modern LLMs.

These observations motivate an actuarial-specific study of scaling behavior: whether performance improves smoothly with model capacity $P$ and data volume $D$, whether it exhibits regime changes or early saturation, and which architectural inductive biases (trees, feed-forward nets, tabular Transformers, or hybrid GLM--NN models) best translate additional scale into actuarially meaningful gains. 

\section{Data}
\label{Data}

Our scaling experiments use an anonymized motor insurance dataset contributed by a multi-national insurer. The extract contains approximately 4.5 million policy-period records across multiple underwriting years and on the order of 60--70 raw rating-factor columns (plus split/sampling fields used for the scaling protocol). To respect commercial sensitivity, we report only aggregated summaries and anonymized marginal patterns. \bl{The data are entirely from the pre-COVID period, so the experiment does not mix pandemic-era frequency or severity effects into the scaling analysis; we cannot disclose more granular policy-year or evaluation-date information for confidentiality reasons.}

For the frequency modeling we will perform here for each record $i$, the response is the claim count $y_i$ (column \texttt{no\_clm}) and the exposure offset is the earned exposure for each policy record, $E_i$. We model $y_i$ using a Poisson likelihood with mean $\lambda_i=E_i r_i$, where $r_i$ is the predicted claim rate per exposure-year. 

The rating factors cover standard actuarial rating dimensions: policyholder and driver demographics (ages, gender, marital status), vehicle characteristics (age, body type, fuel type, power, insured amount), usage and geographic indicators, prior-insurance and claims history. We exclude a small number of extremely high-cardinality categorical fields (more than 100 distinct levels) so that embedding parameterization is controlled and does not dominate architecture comparisons; after this filter, most predictors are categorical with low to moderate cardinality (binary indicators up to roughly 75 levels), complemented by continuous variables such as ages, durations, and score-like fields. In line with common actuarial practice and to keep tokenization consistent across all model families, we discretize continuous covariates using quantile binning and treat the resulting bins as categorical indices. This avoids the additional modeling choices required for numerical embeddings and keeps our architecture comparisons focused on interaction modeling and scaling behavior. Of course, this choice is debatable and recent literature has \bl{focused} on the importance of embedding continuous covariates appropriately, nonetheless, binning was selected as appropriate for the reasons given above.

We briefly summarize the key portfolio figures. Total exposure is approximately 1 million exposure-years, with about 240,000 claims, yielding an overall claim frequency of 0.240 claims per exposure-year. On a record basis, approximately 5\% of policy-periods have a claim and 95\% have none, highlighting the zero-inflated nature of frequency data in motor insurance. Exposure per record is monthly reflecting policy start and end dates within the observation window. The data include a 90/10 train--test split indicator and a pre-generated uniform variate used for nested subsampling in the scaling experiments.

The portfolio mix is dominated by comprehensive cover with smaller shares in other cover types. The extract spans multiple underwriting years, reducing the risk that a single-period idiosyncrasy drives the results. Mean driver age is in the middle ages and mean vehicle age is around 5 years, with both distributions right-skewed toward older ages. Vehicle power and sum-insured measures show long right tails, consistent with heterogeneous fleet composition.

\begin{figure}[p]
\centering
\includegraphics[width=\textwidth,height=0.78\textheight,keepaspectratio]{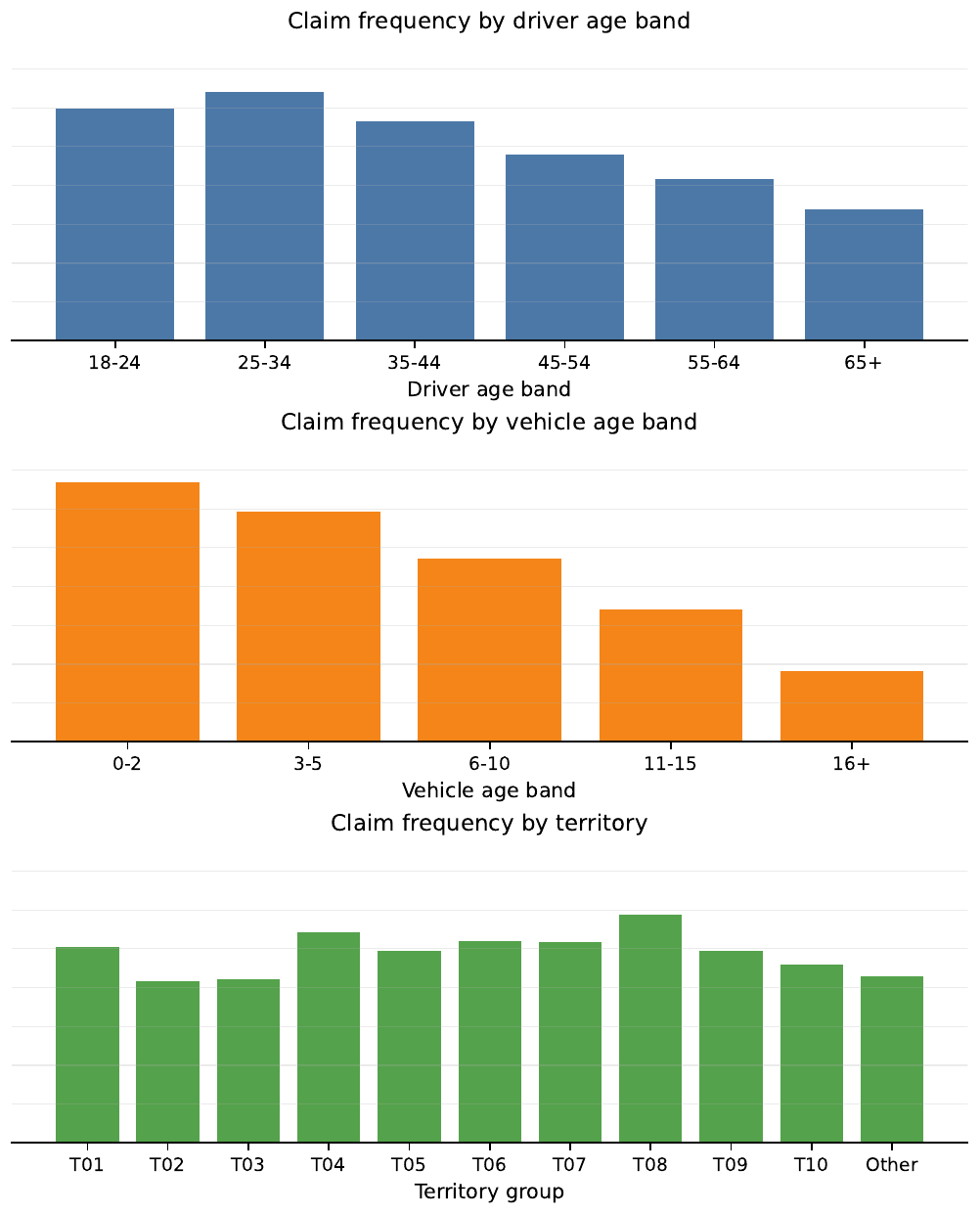}
\caption{One-way claim frequency by selected rating factors (claims per exposure-year). Driver and vehicle age are grouped into actuarial bands; territorial categories are anonymized and ordered by exposure size.}
\label{fig:motor_oneway_frequency}
\end{figure}

Figure \ref{fig:motor_oneway_frequency} provides anonymized one-way summaries of claim frequency (claims per exposure-year) across selected rating factors. Frequency declines monotonically with driver age band after age groups 25-34, with the highest rates among drivers under 35. A similar pattern appears for vehicle age, with newer vehicles showing higher frequency than older ones. Territorial effects are present but more moderate; the figure reports the ten largest territories. These marginal patterns are consistent with actuarial expectations and illustrate the level of risk heterogeneity that models must capture while remaining sufficiently aggregated to avoid revealing proprietary details.

\subsection{Data Preparation}
\label{sec:data_preparation}
We apply minimal preprocessing to align the dataset with the Poisson frequency objective and to avoid leakage:
\begin{itemize}
	    \item Removal of records with non-positive exposure (\texttt{exp\_odpl} $\le 0$), as these are likely prior period correction records.
	    \item Use of \bl{the} \texttt{set} indicator to define a 90/10 IID train--test split.
	    \item Use of the computed uniform variate \texttt{sample\_unif} to define nested training subsets at fractions $f\in\{0.05,0.10,0.25,0.50,0.75,1.00\}$, while keeping the full test set fixed.
	    \item Exclusion of categorical predictors with more than 100 distinct levels (extremely high-cardinality covariates), which would otherwise induce very large and sparse embedding tables and confound controlled capacity comparisons.
	    \item Exclusion of split/sampling columns (\texttt{set}, \texttt{sample\_unif}) from the covariate list.
	    \item Treatment of unknown codes (e.g., negative ``unknown'' values in score-like fields) as distinct categories rather than imputing continuous values.
\end{itemize}

For model inputs, we discretize each continuous predictor into quantile bins computed on the training data and represent all predictors (categorical and binned continuous) using learned entity embeddings (Section~\ref{sec:feature_engineering}). This aligns with traditional actuarial banding practice and standardizes the feature-tokenization layer across model families.

The extract does not contain a sufficiently granular record-level timestamp for temporal validation. We therefore use the provided IID split. Temporal evaluation is treated as a limitation discussed in Section~\ref{sec:limitations}.

\section{Experimental Methodology}
\label{sec:methodology}
\label{sec:methods}

To systematically investigate scaling laws in actuarial ratemaking models, we conducted a series of experiments designed to assess model performance across varying data volumes and model complexities. This section details our experimental design, sampling protocol, training procedure, and evaluation framework.

\subsection{Data Processing and Sampling Strategy}
\label{sec:data_processing}

We structured our experiments to evaluate model performance across different data regimes using \emph{nested} training subsets. Let $u_i\in(0,1)$ denote the pre-generated uniform variate (column \texttt{sample\_unif}) for observation $i$. For each training fraction $f\in\{0.05,0.10,0.25,0.50,0.75,1.00\}$, we train on the subset
\begin{equation}
\mathcal{T}_f \;=\; \{\, i : \texttt{set}_i=\texttt{train},\; u_i \le f \,\},
\end{equation}
and evaluate on the fixed held-out set $\mathcal{S}=\{\,i:\texttt{set}_i=\texttt{test}\,\}$. This design ensures that smaller training sets are strict subsets of larger ones, so changes in performance with $f$ primarily reflect the effect of additional data rather than resampling noise.
\bl{Throughout the paper, $f$ denotes the nested training fraction used to construct the subset, while $N_f=|\mathcal{T}_f|$ denotes the corresponding training size in policy-period rows. Scaling-law equations use $N$ (or $N_f$) as the row-count variable; tables and captions may display $f$ as the compact subset label.}

We operationalize ``data size'' as the number of policy-period rows $N$ in $\mathcal{T}_f$. Because the nested subsets are defined by thresholding a pre-generated uniform variate, they are representative subsamples of the training set; in expectation, total exposure-years in $\mathcal{T}_f$ scales linearly with $N$, so scaling by exposure-years would primarily rescale the horizontal axis rather than change the qualitative regime comparisons.

\bl{For each training fraction,} we first conduct a preliminary analysis to ensure the samples preserve the essential statistical properties of the full dataset:

\begin{equation}
\hat{r}_f = \frac{\sum_{i\in\mathcal{T}_f} y_i}{\sum_{i\in\mathcal{T}_f} E_i}
\end{equation}

where $y_i$ is the claim count and $E_i$ the earned exposure for observation $i$. In practice we use $\hat{r}_f$ as a sanity check that the nested subsets preserve the portfolio's aggregate frequency level, since the Poisson offset initialization and deviance comparisons are sensitive to large shifts in the base rate.

Model architecture details (including the shared feature-tokenization layer and the baseline GLM/FFN definitions) are presented in Section~\ref{sec:model_descriptions}. In what follows we therefore focus on the training recipe, evaluation metric, and how we control for variance and tuning compute when fitting scaling relationships.

\subsection{Training Methodology}
\label{sec:training}

We employed a comprehensive training strategy designed to maximize predictive performance while ensuring robustness. Models were trained using the Poisson deviance loss function, which is the canonical loss for count data:

\begin{equation}
\mathcal{L}_{\text{Poisson}}(\mathbf{y}, \hat{\mathbf{y}}) = \frac{2}{n}\sum_{i=1}^n \left[ \hat{y}_i - y_i - y_i \log\left(\frac{\hat{y}_i}{y_i}\right) \right]
\end{equation}

where $y_i$ represents the observed claim count and $\hat{y}_i$ the predicted count for observation $i$. For $y_i=0$, we use the standard convention that the log term is zero, i.e. $y_i\log(\hat{y}_i/y_i)=0$ (equivalently $y_i\log(y_i/\hat{y}_i)=0$), so the contribution reduces to $\hat{y}_i$.

The optimization procedure utilized AdamW (Adam with decoupled weight decay) \cite{LoshchilovHutter2019,Kingma} with the following hyperparameters:
\begin{align}
\text{Learning Rate} &= 1 \times 10^{-3} \text{ (unless stated otherwise)}\\
\text{Batch Size} &= \text{chosen per architecture to fit hardware constraints} \\
\text{Epochs} &= 50 \text{ (with early stopping)}
\end{align}

To prevent overfitting, we implemented several regularization techniques:
\begin{enumerate}
	    \item Dropout with rates specified in Table \ref{tab:model_configs}
	    \item Early stopping with a patience parameter of 10 epochs
	    \item Learning rate reduction on plateau with factor 0.9 and patience 5
	    \item Additional stabilizers where appropriate (e.g., normalization layers in FFN/Transformer variants)
\end{enumerate}

For each configuration of model size and \bl{training fraction}, we performed five independent training runs with different random initializations. This replication strategy provides robust estimates of model performance and quantifies the variance attributable to initialization conditions. Models were primarily trained on the Colab Research platform from Google, using A100 Graphical Processing Units (GPUs); the largest models are trained on Lambda Labs cloud \bl{platform} using both GH200 and B200 GPUs. We were able to train all models on a single GPU; \bl{there was no need to split training across GPUs}.

To avoid confounding scaling claims with ``tuning budget,'' we distinguish \emph{training compute} from \emph{hyperparameter-search compute}. In this paper, $C$ denotes a \emph{per-epoch} training compute proxy (FLOPs/epoch) for a single trained model (one seed), estimated by profiling the training graph. Total training compute for a run is approximately $C\times$ (epochs trained), and the seed-averaged ensemble multiplies training compute by $R=5$. \bl{For clarity, we write $C_1$ for the one-seed FLOPs/epoch proxy and $C_{\mathrm{ens}}=5C_1$ for the effective per-epoch compute of the five-seed ensemble whose deviance is reported in the main results.} Because we use a shared training schedule with early stopping, we treat FLOPs/epoch as the primary compute scale for controlled cross-family comparisons; estimating total training FLOPs would require recording per-run epoch counts. Compute spent on exploratory hyperparameter search, notebook iteration, and model development is not included in $C$. Operationally, we reduce tuning confounds by fixing the hyperparameter configuration dictionaries upfront (these are shown \bl{in Appendix}~\ref{app:model_configs}) and reusing them across training fractions and random seeds rather than retuning each setting. \bl{We describe this setup as fixed expert-selected recipe scaling:} architecture-specific hyperparameters (including those introduced by our Transformer enhancements, such as the column-weight sparsity penalty and MoE routing settings) were set heuristically\footnote{\bl{By ``heuristically'' we mean that architecture-specific hyperparameters were selected from prior modelling experience and exploratory runs on small slices of the training data. Once selected, these recipes were fixed across training fractions and random seeds.}} rather than optimized via a systematic hyperparameter search. \bl{The hyperparameter choices should therefore not be read as software defaults or as the outcome of a full tuning campaign; a broader search could improve absolute losses and shift some relative rankings.} Thus, differences in scaling slopes primarily reflect differences in model family and capacity, not differences in per-run tuning effort.

\subsection{Evaluation Framework}
\label{sec:evaluation}

Our evaluation uses a single primary metric aligned with standard actuarial frequency modeling: \textbf{Poisson deviance} on the held-out test set (lower is better). For observed claim counts $y_i$ and predicted means $\hat{\lambda}_i$, the deviance is
\begin{equation}
D_{\text{Poisson}}(\mathbf{y}, \hat{\boldsymbol{\lambda}}) = 2\sum_{i=1}^n \left[ y_i \log\left(\frac{y_i}{\hat{\lambda}_i}\right) - (y_i - \hat{\lambda}_i) \right],
\end{equation}
with the standard convention that $y_i \log\left(\frac{y_i}{\hat{\lambda}_i}\right)=0$ when $y_i=0$.
\bl{The tables and figures report the mean deviance $\bar D_{\text{Poisson}}=D_{\text{Poisson}}/n$. For a fixed held-out test set and model $M$ with fitted log-likelihood $\ell_M$,}
\begin{equation}
\bl{\bar D_{\text{Poisson}}(M)=\frac{2}{n}\left(\ell_{\mathrm{sat}}-\ell_M\right),}
\end{equation}
\bl{where $\ell_{\mathrm{sat}}$ is the saturated-model log-likelihood. Hence, for two models evaluated on the same test policies,}
\begin{equation}
\bl{\bar D_{\text{Poisson}}(A)-\bar D_{\text{Poisson}}(B)=\frac{2}{n}\left(\ell_B-\ell_A\right).}
\end{equation}
\bl{A decrease of $\Delta$ in mean Poisson deviance is therefore an average held-out log-likelihood lift of $\Delta/2$ nats per policy, i.e., $\Delta/2$ units on the natural-logarithm scale. This likelihood identity is standard for exponential-family deviance losses and motivates using Poisson deviance both for model fitting and forecast evaluation in count-frequency problems \cite{Wuthrich}.}
We focus on frequency in this paper; extending the scaling study to severity and pure premium objectives is left to future work\footnote{\bl{Since claim frequency is low, materially fewer observations are available for severity modeling, making it harder to demonstrate scaling laws across comparable ranges of parameters and compute.}}.

To assess scaling properties, we computed this metric across each combination of:
\begin{itemize}
    \item \bl{Data amount (nested subset indexed by training fraction $f$, with scaling variable $N_f$)}
    \item Model configuration (architecture family and size)
    \item Train/test split
    \item Random seed (replication)
\end{itemize}

This comprehensive evaluation enables us to quantify the relationship between model performance, data volume, and model complexity.

\subsection{Ensemble Methodology}
\label{sec:ensemble}

To reduce variance and obtain stable scaling comparisons, we train each configuration (architecture family and size) with five independent random seeds at each training fraction $f$. At inference time, we average predictions across seeds. \bl{This is the actuarial seed-averaging procedure introduced by \citeA{Nagging} under the name \emph{nagging}. Throughout the manuscript we use the descriptive term \emph{seed-averaged ensemble} for the same procedure.}

The ensemble prediction for observation $i$ is computed as:

\begin{equation}
\hat{\lambda}^{\text{ens}}_{f}(x_i) = \frac{1}{R}\sum_{r=1}^R \hat{\lambda}^{(r)}_{f}(x_i), \qquad R=5,
\end{equation}

where $\hat{\lambda}^{(r)}_{f}(x_i)$ is the predicted mean claim count from the $r$-th seed for the configuration trained at fraction $f$.

Ensemble predictions typically outperform individual models due to their ability to average out random initialization effects and reduce overfitting. This approach provides a more robust basis for evaluating scaling relationships.

Two distinct notions of ``ensemble'' appear in this paper.
The seed-averaged ensemble above is a classical deep-ensemble-style technique: it trains independent replicas (different initializations and optimization noise) and averages their predictions.
Separately, TabM and TabM-mini are \emph{within-model} ensembles of $K$ implicit submodels trained jointly with extensive weight sharing and lightweight adapters \cite{TabM}.
Because TabM is not equivalent to ``train $K$ independent MLPs and average'', its gains need not be dominated by variance reduction alone; the shared-weights parameterization can also change the bias/representation regime and the optimization dynamics.
In the Results, we therefore complement scaling exponents with stability diagnostics and report both ensemble and mean single-seed scores (Table~\ref{tab:full-data-stability}) to help distinguish improvements due to averaging from improvements due to architecture.

\subsection{Scaling Law Analysis}
\label{sec:scaling_analysis}

To summarize scaling behavior in a way that is comparable across architectures, we fit power laws in the standard ``irreducible loss + power-law'' form. Let $L$ denote test Poisson deviance. We first fit a compute-efficient frontier over all runs,
\begin{equation}
L(C) \approx L_{\infty} + A_C\,C^{-\gamma},
\end{equation}
where $C$ is a training compute proxy \bl{and $A_C$ is the compute-scaling amplitude}. In practice we estimate $C$ as training FLOPs per epoch by profiling the TensorFlow training graph (forward-pass float-ops at batch size 1, scaled by batch size and by a $3\times$ multiplier to approximate forward + backward + parameter update, then multiplied by steps per epoch). \bl{The one-seed proxy is $C_1$; because the plotted loss $L$ is the five-seed ensemble deviance, the compute frontier uses $C_{\mathrm{ens}}=5C_1$.} We fit $L_{\infty}$ and $\gamma$ by restricting to the Pareto set of runs that are not dominated in $(C,L)$.
Because we use early stopping, total training FLOPs differ from FLOPs/epoch by a factor equal to the number of epochs trained; throughout we therefore interpret $C$ as a per-epoch compute proxy.
\bl{Using $C_1$ instead of $C_{\mathrm{ens}}$ would multiply every compute value by the same constant factor, so it would shift the fitted intercept but leave the compute-scaling exponent $\gamma$ unchanged.}

Given a shared fitted floor $L_{\infty}$, we then fit data scaling by model family using best-of-size envelopes over training fractions (i.e., for each training fraction we take the minimum test deviance achieved by any configuration in that family, forming a piecewise ``lower envelope'' curve),
\begin{equation}
L(N) - L_{\infty} \approx A_N\,N^{-\alpha},
\end{equation}
where $N$ is the number of training rows used in the nested subsample (Section~\ref{sec:data_processing}). Finally, to quantify parameter scaling at full data we fit
\begin{equation}
L(P) - L_{\infty} \approx A_P\,P^{-\beta},
\end{equation}
on the full-data Pareto set, where $P$ is the number of trainable parameters \bl{and $A_P$ is the parameter-scaling amplitude}. Using the same parameter profiler, we also report a decomposition $P = P_{\text{emb}} + P_{\text{non-emb}}$ into trainable embedding parameters (embedding tables, positional/CLS embeddings) and trainable non-embedding parameters (all remaining weights). All exponents are estimated by linear regression in log space after subtracting $L_{\infty}$. The resulting fitted $(\alpha,\beta,\gamma)$ values are reported in Section~\ref{sec:results-scaling-laws}.

\section{Model Descriptions}
\label{sec:model_descriptions}

This section describes the neural network model families evaluated. Besides describing standard baselines, we also discuss several adaptations of the Transformer architecture for tabular ratemaking. We follow a consistent notation and highlight only the architectural structure; full configuration tables appear in Appendix~\ref{app:model_configs}. \bl{To reduce notation load, the body gives only the symbols needed for the modelling narrative; Appendix~\ref{app:math_quickref} collects the notation, and Appendix~\ref{app:component_formulae} records the component-level formulae.} While we try to provide a stand-alone account of these models here, we also refer the reader to \citeA{Cred_Trans} for a more detailed exposition of applying the Transformer architecture to tabular actuarial data.

A reading guide: the remainder of the Methodology is split into (i) the experimental protocol (this section) and (ii) the architectural definitions (Section~\ref{sec:model_descriptions} and Appendix~\ref{app:model_configs}). Readers who are mainly interested in the empirical scaling curves can skip the mathematical model definitions and proceed directly to the Results (Section~\ref{sec:results}); Table~\ref{tab:model_family_differences} summarizes the distinguishing components of each family.

Shared setup: all neural models operate on the same tokenized tabular representation - each covariate is treated as a feature token via label encoding (categoricals) or quantile binning (continuous covariates), then mapped to a learned entity embedding - and are trained under an exposure-scaled Poisson objective so that the network predicts a log-rate which is multiplied by exposure at inference time.

We compare the following families and variants:
\begin{itemize}
  \item \textbf{GLM}: an exposure-scaled log-link baseline with an additive (linear) predictor over embedded feature levels; it provides an interpretable reference point and captures no learned interactions beyond the feature representation.
  \item \textbf{FFN}: a feed-forward MLP on concatenated feature embeddings; it learns nonlinearities and cross-feature interactions through dense feature mixing, scaling primarily via embedding dimension and MLP depth/width.
  \item \textbf{TabM-mini}: a parameter-efficient within-model ensemble (BatchEnsemble-style) that broadcasts each example to $K$ implicit members via lightweight rank-1 adapters, then applies a shared MLP backbone; predictions are averaged across members for variance reduction.
  \item \textbf{TabM}: a richer TabM-style ensemble that extends TabM-mini with per-layer adapters, allowing the $K$ implicit members to diverge across depth while still sharing the expensive dense kernels.
  \item \textbf{Transformer (vanilla)}: a supervised tabular Transformer that treats covariates as tokens and uses self-attention to learn interactions explicitly across columns, with a single learned \texttt{[CLS]} token providing a pooled summary for prediction.
  \item \textbf{Transformer (MultiCLS)}: a vanilla Transformer augmented with $n_{\text{cls}}>1$ learned \texttt{[CLS]} tokens; each CLS token acts as a pooling query over the covariate tokens, increasing pooled capacity and exposing $n_{\text{cls}}$ as an explicit scaling knob.
  \item \textbf{Transformer (enhanced / head-fix)}: MultiCLS variants augmented with stabilizers (e.g., LayerScale, drop-path, attention temperature/bias and value gating) and value-side capacity mechanisms (head-fix-style value mixing and layer-wise $v$ embeddings) that improve trainability and let larger models use additional parameters effectively.
  \item \textbf{Transformer (TokenMoE)}: token-wise mixture-of-experts variants that route tokens through expert-specific capacity (in the value pathway), increasing conditional capacity with experts $E$ as a scaling knob and load-balancing regularization to prevent collapse.
  \item \textbf{Transformer (TokenMoE+SSL)}: TokenMoE augmented with a swap-style self-supervised auxiliary objective during training; the auxiliary head is disabled at inference but can improve stability and parameter scaling by providing an additional representation-learning signal.
  \item \textbf{Transformer+TabM-mini}: a hybrid that uses a MultiCLS Transformer front-end to learn interactions and produce a pooled representation, then applies a TabM-mini ensemble head over that pooled vector to combine attention-based interaction learning with within-model ensembling.
\end{itemize}

Because vanilla supervised tabular Transformers show weak parameter scaling in our setting, we treat the Transformer family as a progression from a basic model to ``scaling fixes'' (MultiCLS pooling; stabilization and value-side capacity mechanisms such as head-fix and layer-wise $v$ embeddings; TokenMoE routing; and finally swap-style self-supervised learning). Section~\ref{sec:results-roadmap} summarizes how the Results mirror this progression.

\subsection{Shared Notation and Exposure Scaling}
Let $x_i = (x_{i,1}, \ldots, x_{i,T})$ denote the feature vector for policy $i$, with $T_1$ categorical covariates and $T_2$ continuous covariates, $T=T_1+T_2$. All models are trained for Poisson frequency with exposure scaling:
\begin{equation}
\hat{\lambda}(x_i) = E_i \exp(f_\theta(x_i)),
\end{equation}
so that $f_\theta$ predicts a log-rate and $E_i$ is the earned exposure.

\begin{table}[!htbp]
\centering
\begingroup\color{blue}
\caption{Model families and distinguishing components.}
\label{tab:model_family_differences}
\beautytable
\begin{adjustbox}{max width=\textwidth}
\begin{tabular}{@{}L{0.19\textwidth}L{0.20\textwidth}L{0.28\textwidth}L{0.27\textwidth}@{}}
\toprule
\textbf{Family} & \textbf{Experiment identifiers} & \textbf{Core computation} & \textbf{Distinguishing mechanism / scaling knob} \\
\midrule
GLM
& \texttt{glm}
& Linear log-link model
& Interpretable additive baseline; no learned interactions \\
\addlinespace[0.2em]
FFN
& \texttt{ffn\_\allowbreak *}
& MLP on flattened embeddings
& Dense nonlinear feature mixing; scale via embedding dimension and MLP width/depth \\
\addlinespace[0.2em]
TabM-mini
& \texttt{tabm\_\allowbreak mini\_\allowbreak *}
& Single adapter broadcasts each row to $K$ implicit members, followed by a shared MLP
& Parameter-efficient within-model ensemble; scale via $K$ and embedding dimension \\
\addlinespace[0.2em]
TabM
& \texttt{tabm\_\allowbreak *}
& BatchEnsemble-style adapters throughout dense blocks
& Per-layer member-specific adapters with shared dense kernels; scale via $K$ and width \\
\addlinespace[0.2em]
Transformer (vanilla)
& \texttt{transformer\_\allowbreak *}
& Self-attention over covariate tokens with CLS pooling
& Learned cross-feature interactions; scale via $d$, heads, layers, and FFN dimension \\
\addlinespace[0.2em]
Transformer (MultiCLS)
& \texttt{transformer\_\allowbreak multicls\_\allowbreak *}
& Multiple CLS pooling queries over covariate tokens
& Larger pooled representation; scale via $n_{\text{cls}}$ plus depth/width \\
\addlinespace[0.2em]
Transformer (enhanced/head-fix)
& \texttt{transformer\_\allowbreak enhanced\_\allowbreak *}
& MultiCLS with stabilizers and value-capacity options
& Column weighting, value mixing, sink tokens, LayerScale, drop-path, and attention bias/temperature \\
\addlinespace[0.2em]
Transformer (TokenMoE)
& \texttt{transformer\_\allowbreak tokenmoe\_\allowbreak *}
& Token-wise expert routing in the value pathway
& Conditional capacity; scale via experts $E$ with load balancing \\
\addlinespace[0.2em]
Transformer (TokenMoE+SSL)
& \texttt{transformer\_\allowbreak tokenmoe\_\allowbreak ssl\_\allowbreak *}
& TokenMoE plus swap-style auxiliary task
& Adds row-swap perturbations and swap prediction; scale via experts and auxiliary tokens \\
\addlinespace[0.2em]
Transformer+TabM-mini
& \texttt{transformer\_\allowbreak multicls\_\allowbreak tabm\_\allowbreak mini\_\allowbreak *}
& Transformer front-end followed by a TabM-mini ensemble head
& Combines attention-based interactions with within-model averaging; scale via Transformer size and $K$ \\
\bottomrule
\end{tabular}
\end{adjustbox}
\endgroup
\end{table}

\subsection{Feature Engineering and Input Representation}
\label{sec:feature_engineering}

All neural models in this paper operate on a unified categorical representation of the covariates. For each column $t\in\{1,\ldots,T\}$, we map the raw value $x_{i,t}$ to a non-negative integer index
\begin{equation}
\text{idx}_{t}(x_{i,t}) \in \{0,1,\ldots,n_t-1\},
\end{equation}
where index $0$ is reserved for missing, ``unknown'', or sentinel-coded values (when present).
For categorical variables, $\text{idx}_{t}$ is a label encoding of levels.
For continuous variables, we discretize by quantile binning: letting $c_{t,1}<\cdots<c_{t,n_t-2}$ denote empirical training quantile cutpoints, we map
\begin{equation}
\text{idx}_{t}(x_{i,t}) = 1 + \sum_{m=1}^{n_t-2}\mathbf{1}\{x_{i,t} > c_{t,m}\},
\end{equation}
so that the binned continuous variable is treated as categorical with $n_t-1$ non-missing levels plus the missing/sentinel level $0$.
This design aligns with common actuarial banding practice and avoids additional design choices required for numerical embeddings or separate continuous subnetworks.

Each column $t$ has a learned embedding matrix $\mathbf{E}_t\in\mathbb{R}^{n_t\times b}$, where $b$ is the embedding dimension. The embedded token is
\begin{equation}
\mathbf{e}_t(x_{i,t}) = \mathbf{E}_t[\text{idx}_{t}(x_{i,t})]\in\mathbb{R}^{b}.
\end{equation}
The embedding dimension $b$ is a model hyperparameter: for the GLM baseline we use $b=1$, while for the FFN baseline sizes we scale $b$ as in Table~\ref{tab:model_configs} (and analogously for the Transformer and TabM families; Appendix~\ref{app:model_configs}).

Collecting embeddings across columns yields a token matrix for row $i$,
\begin{equation}
\mathbf{X}_i^\circ = [\mathbf{e}_1(x_{i,1})^\top;\ldots;\mathbf{e}_T(x_{i,T})^\top]\in\mathbb{R}^{T\times b}.
\end{equation}
Feed-forward models flatten $\mathbf{X}_i^\circ$ into a vector, while Transformer-family models treat $\mathbf{X}_i^\circ$ as a short sequence of feature tokens and learn interactions via attention.

\subsection{GLM and FFN Baselines}
The GLM baseline uses the exposure-scaled log-link with a \emph{linear} predictor in the embedded covariates. Let $\mathbf{v}_i=\text{vec}(\mathbf{X}_i^\circ)\in\mathbb{R}^{Tb}$ denote the flattened embedding vector (Section~\ref{sec:feature_engineering}). The GLM predicts
\begin{equation}
\eta_i = \beta_0 + \boldsymbol{\beta}^\top \mathbf{v}_i, \qquad \hat{\lambda}_i = E_i \exp(\eta_i),
\end{equation}
which reduces to a standard GLM when $b=1$ and embeddings act as one-dimensional per-level coefficients.

The FFN generalizes this linear predictor by applying a stack of dense layers to $\mathbf{v}_i$:
\begin{align}
\mathbf{h}_i^{(0)} &= \mathbf{v}_i,\\
\mathbf{h}_i^{(\ell)} &= \text{ReLU}\!\left(\text{BN}(\mathbf{W}_\ell \mathbf{h}_i^{(\ell-1)} + \mathbf{b}_\ell)\right), \quad \ell = 1,\ldots,L,\\
\eta_i &= \mathbf{w}^\top \mathbf{h}_i^{(L)} + b_{\text{out}},\qquad \hat{\lambda}_i = E_i \exp(\eta_i),
\end{align}
with batch normalization (BN) \cite{Ioffe2015BatchNorm} and dropout \cite{Srivastava2014Dropout} between layers. We use the ReLU activation, $\text{ReLU}(x)=\max\{0,x\}$, in our FFN baselines. Batch normalization standardizes layer inputs using minibatch statistics, accelerating training and providing implicit regularization. Dropout randomly zeroes a fraction of activations during training, reducing co-adaptation of neurons and improving generalization. We initialize the output bias to the log of the portfolio base rate,
\begin{equation}
b_{\text{out}}=\log(\bar{r}),\qquad \bar{r}=\frac{\sum_{i\in\mathcal{T}} y_i}{\sum_{i\in\mathcal{T}} E_i},
\end{equation}
so that early in training the model behaves like an intercept-only frequency model and learns deviations from this baseline.

\begin{table}[!htbp]
\centering
\begingroup\color{blue}
\caption{FFN baseline configuration parameters.}
\label{tab:model_configs}
\beautytable
\begin{adjustbox}{max width=\textwidth}
\begin{tabular}{@{}L{0.25\textwidth}C{0.15\textwidth}L{0.25\textwidth}C{0.12\textwidth}C{0.16\textwidth}@{}}
\toprule
\textbf{Config} & \textbf{Embedding dim $b$} & \textbf{Units in hidden layers} & \textbf{Dropout} & \textbf{Params (order)} \\
\midrule
\texttt{ffn\_very\_small} & 5 & \texttt{[16]} & 0.01 & $\mathcal{O}(10^3)$ \\
\texttt{ffn\_small} & 10 & \texttt{[32,\ 16]} & 0.02 & $\mathcal{O}(10^4)$ \\
\texttt{ffn\_medium} & 15 & \texttt{[64,\ 32]} & 0.03 & $\mathcal{O}(10^4)$ \\
\texttt{ffn\_large} & 20 & \texttt{[128,\ 64,\ 32]} & 0.04 & $\mathcal{O}(10^5)$ \\
\texttt{ffn\_very\_large} & 40 & \texttt{[256,\ 128,\ 64]} & 0.05 & $\mathcal{O}(10^6)$ \\
\bottomrule
\end{tabular}
\end{adjustbox}
\endgroup
\end{table}

\subsection{Transformer Architecture for Tabular Data}
We now describe the Transformer model used in our scaling experiments. The reader should have the classical Transformer architecture in mind \cite{Vaswani}, but note that we adapt the encoder Transformer presented there to tabular inputs via a feature tokenizer, learned positional embeddings, and CLS tokens that summarize the entire covariate set; for a full exposition see \citeA{Cred_Trans}. The central motivation for using a Transformer in ratemaking is that attention layers allow covariate components to \emph{communicate} with each other. One may think of each covariate token producing a query and a key; the attention mechanism then finds matches between queries and keys and uses these matches to pass signals forward. In motor pricing, for example, a driver age token may provide a key indicating higher risk for young drivers, while a vehicle power token may provide a query that ``searches'' for this age group, producing an interaction effect that increases expected frequency. Such interactions are fundamental in practice, but are costly to model explicitly in a GLM and can require substantial feature engineering.

Transformer models have been applied successfully to tabular data by embedding tabular columns into low-dimensional Euclidean vectors and treating the resulting set of embeddings as a short ``sequence'' \cite{Huang2020TabTransformer, Gorishniy2021}. In our setting $T$ is modest (dozens of covariates), so the quadratic attention cost in $T$ is negligible; the primary question becomes whether the architecture and training strategy can reliably extract signal in a low signal-to-noise actuarial regime.

\subsubsection{Feature Tokenizer}
Transformer-family models use the same feature engineering and tokenization described in Section~\ref{sec:feature_engineering}: each covariate column is encoded to an integer index and mapped to a learned entity embedding \cite{Guo}. The key difference from feed-forward baselines is that we keep the per-column embeddings as a short token sequence rather than flattening them immediately, enabling interaction learning via attention.

\bl{The exact embedding-table notation and parameter count are given once in Section~\ref{sec:feature_engineering} and summarized again in Appendix~\ref{app:math_quickref}. In the Transformer descriptions below, $\mathbf{X}_i^\circ\in\mathbb{R}^{T\times b}$ denotes the resulting raw token matrix for policy $i$.}
We emphasize that this ``tokenizer'' step is the key bridge between tabular learning and sequence models: after tokenization, the model sees a short set of $T$ vectors in $\mathbb{R}^b$, one per covariate, and can therefore apply attention to learn interactions.

\subsubsection{Positional Encoding}
Attention layers do not have a natural notion of ``position'', so we complement the raw tensor by a learned positional encoding \cite{Vaswani}. For tabular inputs, the covariate ordering is not temporal; nevertheless, we assign each covariate a stable index $t\in\{1,\ldots,T\}$ and learn a $p$-dimensional embedding $\mathbf{e}^{\rm pos}(t)\in\mathbb{R}^p$ for that index. We use two equivalent fusion strategies across our model variants:
	\begin{enumerate}
	    \item \textbf{Concatenated positional embeddings (baseline/multi-CLS).} We concatenate positional and feature embeddings,
	    \begin{equation}
	    \mathbf{X}_{1:T} = \left[\mathbf{X}^\circ_{1:T},\; [\mathbf{e}^{\rm pos}(1), \ldots, \mathbf{e}^{\rm pos}(T)]^\top\right] \in \mathbb{R}^{T \times d}, \qquad d=b+p,
	    \end{equation}
	    which provides each token with an explicit ``identity'' subvector for its column.
		    \item \textbf{Additive positional embeddings (head-fix/TokenMoE).} We project positional embeddings to $\mathbb{R}^b$ and add them to feature embeddings,
	    \begin{equation}
	    \mathbf{X}_{1:T} = \mathbf{X}^\circ_{1:T} + [\mathbf{P}\mathbf{e}^{\rm pos}(1), \ldots, \mathbf{P}\mathbf{e}^{\rm pos}(T)]^\top \in \mathbb{R}^{T \times b},
	    \end{equation}
	    where $\mathbf{P}\in\mathbb{R}^{b\times p}$ is a learned projection (often the identity when $p=b$).
	\end{enumerate}
	We denote by $d$ the resulting token dimension ($d=b+p$ in the concatenation case and $d=b$ in the additive case). The positional embedding contributes $\varrho^{\rm position}=Tp$ trainable parameters. Both designs break permutation symmetry in the token set and allow the model to learn column-specific roles even if two covariates have similar marginal distributions.

\subsubsection{CLS Tokens}
We extend the input tensor by additional special tokens (CLS tokens; CLS stands for ``classification''), inspired by the CLS mechanism introduced in BERT \cite{Devlin2018BERT}. The main idea of a CLS token is that it learns to summarize all the information contained in $\mathbf{X}_i^\circ$ in a lower-dimensional space that is calibrated to produce good regression estimates. In language models, CLS tokens are used to summarize information in a sequence for downstream tasks. In our tabular regression setting, CLS tokens play an analogous role: they provide dedicated slots into which the attention mechanism can write a compact summary of the entire covariate set.

We append $n_{\text{cls}}$ learned CLS tokens $\mathbf{c}_1,\ldots,\mathbf{c}_{n_{\text{cls}}} \in \mathbb{R}^d$ to the input sequence,

\begin{equation}
\mathbf{X}^+_{1:T+n_{\text{cls}}} =
\left[\mathbf{X}_{1:T}^\top, \mathbf{c}_1^\top, \ldots, \mathbf{c}_{n_{\text{cls}}}^\top \right]^\top \in \mathbb{R}^{(T+n_{\text{cls}})\times d}.
\end{equation}
These tokens do not correspond to any observed covariate. Rather, they act as trainable global parameters that are updated by attention to reflect the current policy's covariate information. \bl{Their parameter contribution is $n_{\text{cls}}d$ and is listed with the component formulae in Appendix~\ref{app:component_formulae}.} In all Transformer variants, only the final CLS token(s) are forwarded to the prediction head; the feature tokens are intermediate latent variables whose role is to communicate information to the CLS summaries through attention. When $n_{\text{cls}}>1$ we obtain a multi-CLS Transformer; the pooling and readout of multiple CLS tokens is described in detail below.

\subsubsection{Transformer Layer}
We now describe a Transformer layer. For simplicity we start with one attention head. The query, key, and value projections are computed (after layer normalization) as
\begin{equation}
\mathbf{Q} = \phi(\text{LN}(\mathbf{X}^+) \mathbf{W}_Q), \qquad \mathbf{K} = \phi(\text{LN}(\mathbf{X}^+) \mathbf{W}_K), \qquad \mathbf{V} = \phi(\text{LN}(\mathbf{X}^+) \mathbf{W}_V),
\end{equation}
where $\phi$ is a smooth activation (GELU in our implementation). Attention weights are obtained by the softmax normalization
\begin{equation}
\mathbf{A} = \text{softmax}\left(\frac{\mathbf{Q}\mathbf{K}^\top + \mathbf{B}}{\tau}\right), \qquad \mathbf{H} = \mathbf{A}\mathbf{V},
\end{equation}
where $\mathbf{B}$ is a learnable additive attention bias (a ``soft mask'') and $\tau$ is a learnable temperature (a per-head generalization of the classical $\sqrt{d_h}$ scaling). The output $\mathbf{H}$ has the same shape as $\mathbf{X}^+$ and can be viewed as an attention-weighted re-encoding of the original tokens. Importantly, the rows corresponding to CLS tokens now contain summaries of the full covariate set, because the CLS queries attend to all covariate keys.

For multi-head attention (MHA), we compute $M$ independent heads and concatenate them, followed by an output projection:
\begin{equation}
\text{MHA}(\mathbf{X}^+) = \text{Concat}(\mathbf{H}_1,\ldots,\mathbf{H}_M)\mathbf{W}_O.
\end{equation}
This allows the model to represent different interaction patterns in different subspaces.

We then apply residual connections and a feed-forward block (with layer normalization),
using the standard two-sublayer structure (attention residual, then FFN residual). Concretely,
\begin{align}
\mathbf{X}' &= \mathbf{X}^\ell + \text{LN}\!\left(\text{Dropout}(\text{MHA}(\mathbf{X}^\ell))\right),\\
\mathbf{X}^{\ell+1} &= \mathbf{X}' + \text{LN}\!\left(\text{Dropout}(\text{FFN}(\text{LN}(\mathbf{X}')))\right).
\end{align}
The FFN uses a gated activation (SwiGLU) and dropout. Multi-head attention concatenates $M$ heads and projects back to the token dimension. Appendix~\ref{app:component_formulae} records the component-level formulae used across our variants.

\subsubsection{Prediction Head}
We extract the final CLS representation(s) and map them to a positive rate. For the single-CLS case ($n_{\text{cls}}=1$) we use a small head network $g$ and set
\begin{equation}
\hat{\lambda}_i = E_i \exp\!\left(g(\mathbf{h}_{\text{cls},i})\right).
\end{equation}
For $n_{\text{cls}}>1$ we use multi-CLS pooling and readout; see the next subsection.
To improve optimization stability, the output bias is initialized to the logarithm of the portfolio base rate (the empirical mean frequency), so that early in training the model behaves like a mean model and learns deviations from this baseline.

\subsubsection{Multi-CLS Pooling and Readout}
Multi-CLS models generalize the single-CLS architecture, which is standard in BERT-style Transformer readout \cite{Devlin2018BERT} and in many tabular Transformer models \cite{Gorishniy2021}, by appending $n_{\text{cls}}>1$ learned CLS tokens. These tokens act as multiple \emph{pooling queries}: after the Transformer stack, each CLS token has attended to the feature tokens and thus contains a distinct summary of the policy's covariates and their learned interactions. This use of multiple learnable pooling tokens is conceptually related to pooling-by-attention in set models \cite{Lee2019SetTransformer} and to latent-array summaries in Perceiver-style architectures \cite{Jaegle2021Perceiver}, but is less common in tabular deep learning applications. We first considered using Multi-CLS models when it was observed that single CLS Transformers were not scaling; we hypothesized that this might be due to a single CLS token acting as a ``bottleneck''.

We denote by
\begin{equation}
\mathbf{H}_{{\rm cls},i} = [\mathbf{h}_{{\rm cls},1,i}^\top;\ldots;\mathbf{h}_{{\rm cls},n_{\text{cls}},i}^\top]\in\mathbb{R}^{n_{\text{cls}}\times d}
\end{equation}
the matrix of final CLS states for observation $i$. In our experiments we use two closely related readout schemes.

First, we use a \emph{concatenated} CLS readout. This is used in the baseline multi-CLS models (\texttt{transformer\_\allowbreak multicls\_\allowbreak *}) and in the Transformer+TabM-mini hybrids. We optionally layer-normalize $\mathbf{H}_{{\rm cls},i}$ and vectorize it to obtain a pooled vector
\begin{equation}
\mathbf{h}^{\rm flat}_i=\text{vec}(\text{LN}(\mathbf{H}_{{\rm cls},i}))\in\mathbb{R}^{n_{\text{cls}}d},
\end{equation}
which is then mapped to a log-rate by a small head network $g$,
\begin{equation}
\hat{\lambda}_i = E_i \exp\!\left(g(\mathbf{h}^{\rm flat}_i)\right).
\end{equation}
This design increases representation capacity linearly in $n_{\text{cls}}$ without changing the number of feature tokens $T$.

Second, in the enhanced/head-fix and TokenMoE variants, we use an \emph{ensemble} CLS readout: each CLS token is mapped to a rate by its own small head $g_c$, and the resulting rates are averaged,
\begin{equation}
\hat{\lambda}_i = E_i \cdot \frac{1}{n_{\text{cls}}}\sum_{c=1}^{n_{\text{cls}}} \exp\!\left(g_c(\mathbf{h}_{{\rm cls},c,i})\right).
\end{equation}
This can be interpreted as an internal ensemble with a shared interaction backbone and multiple specialized pooled summaries. Conceptually, multi-CLS differs from multi-head attention: multi-head attention creates multiple interaction channels \emph{within} a layer, whereas multi-CLS creates multiple pooled summaries \emph{at the output} of the interaction stack.

\subsubsection{Rationale and Interpretation}
The key inductive bias of the Transformer in this setting is that interactions between covariates are modeled directly by attention rather than being imposed by manual feature construction. The CLS tokens provide a bottleneck representation: they compress the full set of covariate tokens into a fixed-dimensional vector (or a small set of vectors), which is then mapped to the scalar log-rate. This architecture supports both predictive accuracy and interpretability: the attention weights from CLS tokens to covariate tokens can be inspected to understand which covariates the model relied on for different policies, and multi-head attention provides multiple ``views'' on these interactions.

\subsubsection{Relation to Prior Work and Novel Components}
Transformers for tabular data have been studied extensively in recent years, typically by embedding each column into a shared vector space and treating columns as tokens. TabTransformer contextualizes categorical embeddings through self-attention \cite{Huang2020TabTransformer}, and FT-Transformer generalizes this idea by tokenizing both numerical and categorical features and applying standard Transformer blocks to the resulting token sequence \cite{Gorishniy2021}. Our baseline Transformer follows the FT-Transformer philosophy, but uses a simple quantile-binning tokenizer for continuous covariates (so that all columns are embedded uniformly) and is trained under an exposure-scaled Poisson objective, which is the natural likelihood for claim frequency in ratemaking.

The most novel architectural ingredient here, relative to common tabular Transformer baselines, is our systematic use of \emph{multi-CLS} pooling. Standard Transformer readouts rely on a single CLS token \cite{Devlin2018BERT}, and most tabular Transformer implementations similarly produce one pooled vector. Our multi-CLS design introduces multiple pooling queries and exposes $n_{\text{cls}}$ as an explicit scaling knob, with two readout options (concatenation vs ensemble averaging). This is structurally related to the use of multiple learnable pooling seeds for set inputs \cite{Lee2019SetTransformer} and to latent arrays in Perceiver models \cite{Jaegle2021Perceiver}, but here it is applied as a simple and computationally cheap way to increase capacity and reduce variance in actuarial tabular prediction.

\subsection{Transformer Enhancements}
Building on the base Transformer, we introduce several stability and capacity enhancements.
We motivate these modifications by a practical observation: tabular actuarial datasets often have limited effective sample size relative to model capacity (particularly after accounting for heterogeneity across rating factors), and optimization can be brittle. In this regime, architectural choices that are minor in NLP can materially affect convergence and generalization.

Our enhanced Transformer variants combine several stabilization and capacity mechanisms that have precedents in the broader Transformer literature but are not standard in tabular Transformers. Column weighting provides a differentiable feature-selection gate reminiscent of attentive masks in TabNet \cite{Arik2019TabNet}; we also cite \citeA{RealMLP} with a similar mechanism. The stabilizers that we adapt are deep-Transformer training techniques including RMS normalization \cite{Zhang2019RMSNorm}, stochastic depth (drop-path) \cite{Huang2016StochasticDepth}, and LayerScale \cite{Touvron2021CaiT}, as well as learned per-head temperature/bias and value gating. While the individual ingredients are known, our contribution is to integrate them into a coherent family of tabular ratemaking models and evaluate them across a wide range of model sizes and data regimes.

To reach substantially larger parameter counts, we also study TokenMoE models that introduce conditional computation via mixture-of-experts routing. MoE layers are a standard scaling mechanism in large language models \cite{Shazeer2017MoE, Fedus2021Switch}; our novelty is to adapt token-wise expert routing to tabular covariate tokens and to use the number of experts as a scaling dimension under fixed training budgets.

We include TabM/TabM-mini ensembles and a Transformer+TabM-mini hybrid \cite{TabM}. These models implement parameter-efficient ensembling through shared weights with lightweight rank-1 adapters, akin to BatchEnsemble \cite{Wen2020BatchEnsemble}. Swap-style self-supervision is related in spirit to tabular SSL approaches such as SAINT \cite{Somepalli2021SAINT}, but uses a simple row-swap corruption-and-detection objective to provide an auxiliary training signal and improve stability when scaling up model capacity.

In what follows, we describe these enhancements.

\subsubsection{Depth and Multi-head Attention}
We scale the base Transformer by increasing depth $L$, token dimension $d$, number of heads $M$, and FFN width. Multi-head self-attention allows the model to represent multiple interaction patterns in parallel, and deeper stacks iteratively refine these interaction summaries. Empirically, shallow single-head designs tend to either underfit (insufficient interaction capacity) or train unstably once regularization is relaxed; the modifications below improve the capacity--stability trade-off.

\subsubsection{Column Weighting}
We introduce a differentiable soft feature selection mechanism by learning a nonnegative weight per covariate token and scaling the corresponding token vector. Concretely, with trainable scalars $a_t\in\mathbb{R}$ we define
\begin{equation}
w_t = \min\{1,\max\{0,a_t\}\}, \qquad \tilde{\mathbf{x}}_t = w_t\,\mathbf{x}_t,
\end{equation}
and optionally add a sparsity penalty proportional to $\sum_{t=1}^T w_t$. This reduces the tendency of attention to spread mass over many weakly informative covariates and is particularly useful in the presence of correlated or low-signal predictors.

\subsubsection{Layer-wise Value Mixing}
Enhanced variants optionally allocate extra capacity to the \emph{value} pathway without widening the full attention width.\footnote{\bl{The original public source for this engineering idea is the Twitter/X discussion cited in \citeA{Tweet}; the corresponding public implementation is the \texttt{modded-nanogpt} GitHub repository \cite{ModdedNanoGPT}.}}

Recall that attention can be viewed as a two-stage operation: (i) queries and keys determine \emph{where} information should be gathered from (an interaction pattern across covariates), while (ii) values determine \emph{what} content is aggregated once that interaction pattern is chosen.

In tabular settings, where $T$ is small but the signal-to-noise ratio can be low, we found it beneficial to increase the amount of value-side content available to each layer while keeping the query/key dimension fixed.

\bl{Concretely, enhanced/head-fix variants learn an extended embedding for each covariate token, split into a base stream and $L$ layer-wise value streams. At layer $\ell$, the value tensor used by attention is a learned softmax mixture of current-layer values, cached previous-layer values, and projected layer-$\ell$ value embeddings.}
Intuitively, the v embeddings provide each layer with \emph{fresh, layer-specific content} tied to the original covariate identities, while the cached term provides a short residual memory in the value stream across depth.
In our implementations, v embeddings are set to zero for CLS tokens, so that any additional value-side capacity is anchored in the observed covariates rather than being injected directly into the pooled summary.

This value-pathway augmentation is conceptually related to recent work on \emph{value residual} connections in Transformers \cite{Zhou2024ValueResidualLearning}, which argues that carrying value information across depth can improve information flow in deep networks.
Our setting and design differ in three important ways.
First, we operate in tabular ratemaking with short token sequences, so key--value cache efficiency (a major motivation in language modelling) is not a bottleneck.
Second, rather than sharing the first-layer value embedding across layers, we explicitly learn layer-indexed v embeddings at the tokenizer level and inject them only through the value pathway via learned mixing weights.
Third, in TokenMoE variants we make the v stream itself conditionally parameterized by expert routing, turning value-side capacity into an explicit scaling knob aligned with our scaling-law experiments.

\subsubsection{Stability Enhancements}
Our enhanced Transformer variants combine several stabilizers that make deeper and wider models feasible: RMS normalization of $Q/K$ \cite{Zhang2019RMSNorm}, learned per-head temperatures and additive attention biases, per-head gating of attended values, optional sink tokens that absorb global context, LayerScale on residual branches \cite{Touvron2021CaiT}, an FFN residual gate, and stochastic depth (drop-path) \cite{Huang2016StochasticDepth}. The component-level formulae are recorded in Appendix~\ref{app:component_formulae}.

\subsection{Token-wise Mixture-of-Experts (TokenMoE)}
TokenMoE is a token-level conditional computation mechanism that increases capacity without uniformly increasing compute, inspired by sparsely-gated mixture-of-experts layers and Switch Transformers \cite{Shazeer2017MoE, Fedus2021Switch}. \bl{Let $E$ denote the number of experts. A lightweight router assigns each covariate token to expert-specific embedding increments, and these increments are mixed into the dense base stream and the layer-wise value stream. This allows scaling to much larger parameter counts while keeping per-token computation close to dense baselines; Appendix~\ref{app:component_formulae} gives the routing equations and load-balancing loss.}

\bl{To avoid expert collapse, we add a load-balancing penalty that encourages more uniform expert utilization. We also expose explicit scaling coefficients (e.g., \texttt{moe\_alpha\_base} and \texttt{moe\_alpha\_v}) that control the strength of the MoE residual contribution in different embedding streams, which improves stability.}

The key point for our setting is that TokenMoE acts \emph{before} the attention stack and operates directly on the token embedding streams.
\bl{Each feature token carries two conceptually distinct pieces of information: a base embedding that governs attention interactions (via $Q/K$ and the baseline values), and the layer-wise v embeddings used in the value-mixing mechanism. TokenMoE routes and augments both components. The Transformer stack then runs on the augmented base stream, while each layer receives the matching augmented v slice as its value-embedding input. This ``v-MoE'' design increases conditional capacity in the value pathway while leaving the attention pattern itself relatively constrained, which is helpful when scaling to large parameter counts in noisy actuarial regimes.}

\subsubsection{Swap-style Self-supervision}
Purely supervised training on insurance frequency can provide a relatively weak optimization signal for high-capacity Transformers: claims are sparse, and the supervised loss depends only on a scalar log-rate for each observation. As model size increases, this can lead to brittle optimization and to weak returns from parameter scaling unless the architecture receives additional inductive bias or auxiliary signal (Section~\ref{sec:results-roadmap}).

To provide a stronger label-free training signal that is tailored to our tokenized tabular setting, we introduce a simple corruption-and-detection objective, related in spirit to tabular self-supervised learning methods such as SAINT \cite{Somepalli2021SAINT} and to corruption-based objectives in representation learning more broadly \cite{Devlin2018BERT}.
\bl{Operationally, during training we independently swap selected covariate-token embeddings with embeddings from donor rows in the same minibatch. This preserves each column's marginal embedding distribution while breaking within-row coherence across columns. We append one learned swap CLS token per feature position and train a small auxiliary head to predict which positions were swapped, adding a weighted binary cross-entropy term to the supervised Poisson loss. Appendix~\ref{app:component_formulae} records the exact perturbation, head definition, and loss.}

Thus, we try to predict if, during training, one of the token rows was swapped for another. Unlike contrastive tabular SSL approaches that rely on paired augmentations and representation matching \cite{Somepalli2021SAINT}, our objective is a direct \emph{corruption detection} task trained jointly with the supervised Poisson objective.
Because swaps draw donors from the same minibatch and preserve feature position, the marginal distribution of each column is (approximately) unchanged, and the swap head must therefore leverage cross-feature consistency rather than trivial per-column cues.
\bl{In our implementation the swap probability and loss weight correspond to the configuration keys \texttt{swap\_alpha} and \texttt{swap\_loss\_weight}. At inference time the corruption is disabled and the swap head is ignored; predictions depend only on the supervised Poisson head, but benefit from representations shaped by the auxiliary task.}

\subsection{TabM and TabM-Mini Ensembles}
TabM \cite{TabM} is a parameter-efficient ensembling strategy for tabular MLPs, closely related to BatchEnsemble \cite{Wen2020BatchEnsemble}. The key idea is to pack an ensemble of $K$ implicit MLP \emph{submodels} into a single network by sharing the main dense kernels while equipping each submodel with lightweight rank-1 \emph{adapters}.
For a dense layer with shared kernel $\mathbf{W}\in\mathbb{R}^{d_{\rm in}\times d_{\rm out}}$, member $k\in\{1,\ldots,K\}$ uses per-member adapters $(\mathbf{r}_k,\mathbf{s}_k,\mathbf{b}_k)$ and computes an ``ensemble dense'' transform
\begin{equation}
\mathbf{h}^{\rm out}_{i,k}
= \left((\mathbf{h}^{\rm in}_{i,k}\odot \mathbf{r}_k)\mathbf{W}\right)\odot \mathbf{s}_k + \mathbf{b}_k,
\end{equation}
which is equivalent to using an implicitly modulated weight matrix $\mathbf{W}_k=\mathbf{W}\odot(\mathbf{r}_k\mathbf{s}_k^\top)$. Adding ensemble members therefore incurs only $\mathcal{O}(d_{\rm in}+d_{\rm out})$ parameters per layer (one row in each adapter matrix) rather than duplicating full kernels, while still enabling variance reduction by averaging predictions across submodels.

TabM-mini \cite{TabM} is the minimal version of this idea: it retains only the first multiplicative adapter $\mathbf{r}_k$ and removes the remaining $3N-1$ adapters (where $N$ is the number of linear blocks). Intuitively, this first adapter maps a shared input representation into $K$ slightly different representation spaces \emph{before} the features are mixed by the first dense kernel, after which a shared MLP backbone is applied memberwise.
TabM restores per-layer adapters throughout the backbone but uses a ``better initialization'': all multiplicative adapters beyond the first are initialized to 1, so the model starts by behaving like TabM-mini and can gradually grow additional expressivity during training \cite{TabM}.

In our implementation, we broadcast each example across $K$ submodels (shared training batches) and apply ReLU non-linearities in the MLP backbone. We train with a memberwise Poisson negative log-likelihood on $\mu_{i,k} = E_i \exp(g(\mathbf{h}_{i,k}))$, and at inference average the submodel rates before multiplying by exposure.
Finally, Transformer-plus-TabM-mini hybrids use a multi-CLS Transformer front-end to compute a pooled representation and then replicate it across $K$ members via the TabM-mini adapter, combining attention-based interaction learning with within-model ensemble averaging.

\section{Results}
\label{sec:results}

\subsection{Experimental protocol and metric}
\label{sec:results-protocol}

We evaluate models on a large-scale insurance frequency prediction task with a Poisson objective.
All models predict a non-negative claim count rate which is multiplied by exposure at inference time.
We report \emph{Poisson deviance} on a held-out test set (lower is better).
To study data scaling, we train on fractions $f \in \{0.05, 0.10, 0.25, 0.50, 0.75, 1.00\}$ of the available training rows,
corresponding to approximately
$N \in \{2.02\times 10^5, 4.04\times 10^5, 1.01\times 10^6, 2.02\times 10^6, 3.03\times 10^6, 4.03\times 10^6\}$ examples.
\bl{Unless stated otherwise, each configuration is trained with 5 random seeds and we report the seed-averaged ensemble score, defined as averaging predictions across seeds at inference time (Section~\ref{sec:results-variance}).}

\subsection{Results roadmap: from baselines to scaling fixes}
\label{sec:results-roadmap}
The results are organized to mirror the modeling development narrative.
We begin with classical and MLP baselines (GLM and FFN), then evaluate supervised tabular Transformers and diagnose their weak scaling in this regime, and then progressively add Transformer modifications (MultiCLS pooling; stabilization and value-side capacity mechanisms such as head-fix, layer-wise v embeddings, and v-MoE; and TokenMoE routing).
We then introduce the swap-style self-supervised objective (SSL), which provides a qualitatively different training signal and improves Transformer scaling.
Finally, we evaluate TabM and TabM-mini, and a Transformer+TabM hybrid that transfers TabM-style variance reduction and adaptation mechanisms into the Transformer pipeline.

\begin{figure}[p]
\centering
\IfFileExists{figures/size_scaling_figure.pdf}{%
\includegraphics[width=\textwidth,height=0.82\textheight,keepaspectratio]{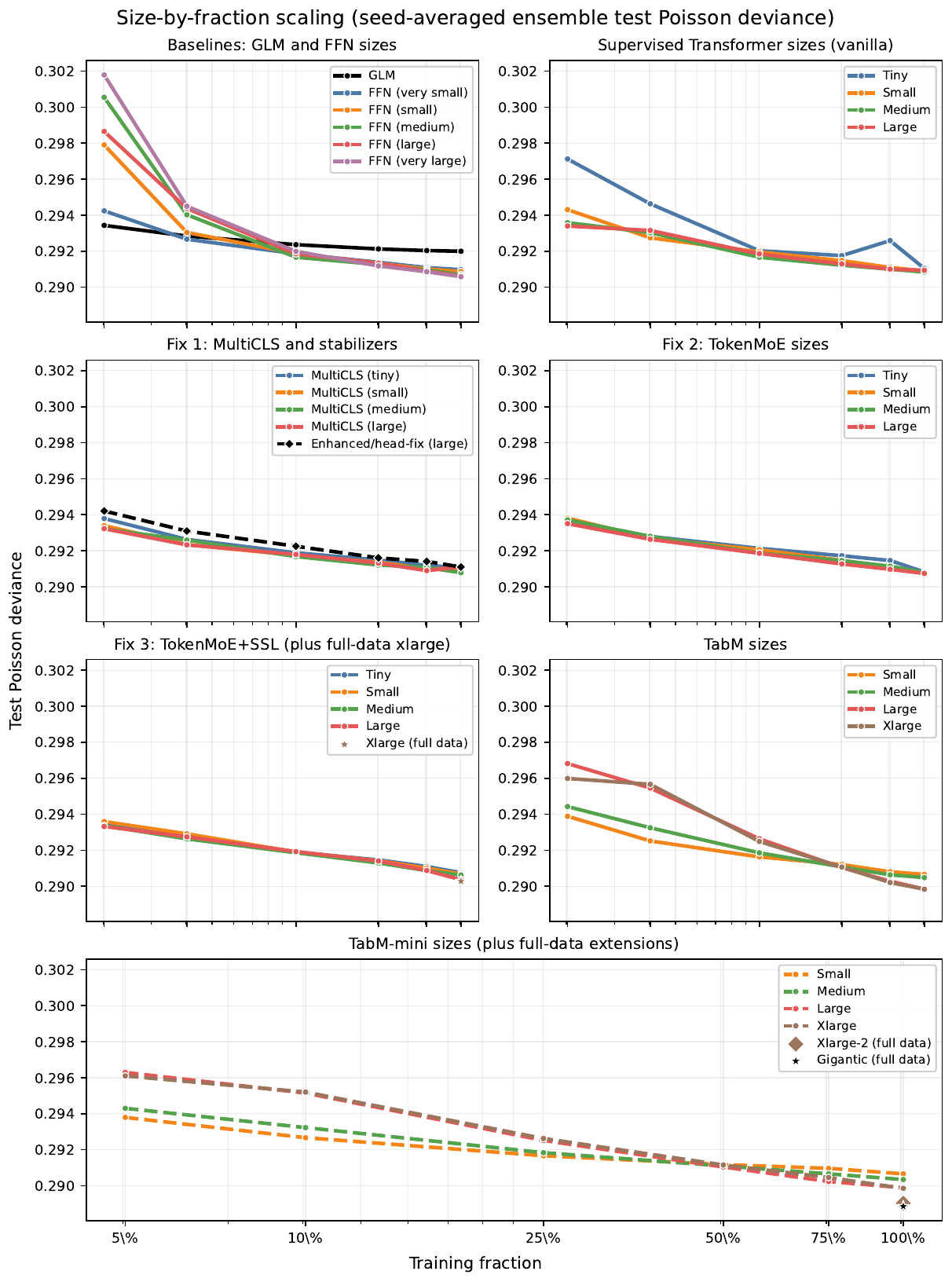}%
}{%
\fbox{\parbox{0.98\textwidth}{\centering Missing figure: \texttt{figures/size\_scaling\_figure.pdf}}}%
}%
\caption{Size-by-fraction scaling for the main model families (seed-averaged ensemble test Poisson deviance; lower is better).
This figure makes explicit when larger models help (or hurt) as training data increases, complementing the best-of-family envelope summary in Table~\ref{tab:main-scaling}.}
\label{fig:size-scaling}
\end{figure}

To make the role of \emph{model size} explicit (beyond best-of-family envelopes), Figure~\ref{fig:size-scaling} plots \bl{seed-averaged ensemble} test deviance across training fractions for the full size ladders within each major family, and Tables~\ref{tab:size_scaling_baselines}--\ref{tab:size_scaling_tabm} provide the corresponding numeric values.
This view highlights five qualitative facts that frame the remainder of the Results.

First, GLMs remain strong and stable throughout, but larger FFNs only become favorable as $N$ increases: at small fractions the smallest FFNs are best, while at larger fractions the optimum shifts to the largest FFN (Figure~\ref{fig:size-scaling}, top-left), consistent with a small-data trap.
Second, purely supervised Transformers exhibit weak parameter scaling: increasing size yields only marginal improvements, and the best size is not consistently the largest (Figure~\ref{fig:size-scaling}, top-middle).
Third, MultiCLS pooling and stabilization/value-side capacity mechanisms (e.g., head-fix with layer-wise v embeddings) provide modest improvements over the vanilla Transformer but do not fundamentally change the scaling picture (Figure~\ref{fig:size-scaling}, top-right).
Fourth, TokenMoE improves monotonicity with size but still yields small absolute gains under a purely supervised objective; adding swap-style self-supervision (TokenMoE+SSL) produces a clearer improvement at larger $N$ and yields the best Transformer-family full-data result (Figure~\ref{fig:size-scaling}, bottom-middle). \bl{The TokenMoE+SSL configurations differ from the TokenMoE ladder in width and depth schedule (Appendix~\ref{app:model_configs}), so we interpret the comparison as recipe-family evidence; a fixed-architecture auxiliary-loss ablation remains outside the present experiment.}
Fifth, TabM-style models scale most strongly with data and capacity: the compute-optimal configuration shifts to larger TabM/TabM-mini variants as $N$ grows, and the largest TabM-mini extensions achieve the best full-data performance overall (Figure~\ref{fig:size-scaling}, bottom-right; Table~\ref{tab:main-scaling}). Concretely, within the standard TabM size ladder the best configuration shifts from \texttt{tabm\_small} at 5\%--25\% to \texttt{tabm\_large} at 50\% and \texttt{tabm\_xlarge} at 75\%--100\% (Table~\ref{tab:size_scaling_tabm}).

For full results by model family, see \ref{sec:results-size-ladders}.

\subsection{Overall performance and scaling with data}
\label{sec:results-overall}

Table~\ref{tab:main-scaling} summarizes test Poisson deviance as a function of training fraction.
For model families, we report the \emph{best-of-size envelope} within the \emph{main sweep} of configurations trained at all fractions (i.e., excluding full-data-only variants).
Here, the \emph{size ladder} for a family is the fixed set of discrete capacity settings we sweep (e.g., tiny/small/medium/large), and the \emph{envelope} is the pointwise minimum test deviance across this ladder at each training fraction.
We additionally report a small set of \emph{full-data-only} runs that extend the largest configurations (notably, TokenMoE+SSL xlarge and TabM-mini xlarge variants).
This envelope view is the basis for the scaling-law fits in Section~\ref{sec:results-scaling-laws} and for the crossover and robustness diagnostics below.
\bl{While these absolute improvements are numerically modest (e.g., 0.29200 $\rightarrow$ 0.28885 is a 0.00315 deviance reduction, about 1.1\% relative), the full-data GLM-versus-TabM/TabM-mini improvements are consistent across the repeated seeds; smaller family crossover gaps require the seed-level uncertainty check reported below.}

\begin{table}[t]
\centering
\begingroup\color{blue}
\beautytable
\begin{adjustbox}{max width=\textwidth}
\begin{tabular}{@{}L{0.31\textwidth}rrrrrr@{}}
\toprule
\textbf{Family (best-of-size)} & \textbf{5\%} & \textbf{10\%} & \textbf{25\%} & \textbf{50\%} & \textbf{75\%} & \textbf{100\%} \\
\midrule
GLM & 0.29343 & 0.29284 & 0.29236 & 0.29212 & 0.29204 & 0.29200 \\
FFN (envelope) & 0.29424 & 0.29267 & 0.29166 & 0.29118 & 0.29086 & 0.29058 \\
	Vanilla transformer (envelope) & 0.29340 & 0.29274 & 0.29166 & 0.29122 & 0.29100 & 0.29085 \\
	MultiCLS transformer (envelope) & 0.29321 & 0.29232 & 0.29168 & 0.29120 & 0.29089 & 0.29078 \\
	Transformer+TabM (envelope) & \textbf{0.29282} & \textbf{0.29217} & \textbf{0.29147} & 0.29108 & 0.29069 & 0.29056 \\
	TokenMoE transformer (envelope) & 0.29350 & 0.29263 & 0.29186 & 0.29126 & 0.29097 & 0.29074 \\
	TokenMoE+SSL (envelope) & 0.29332 & 0.29263 & 0.29186 & 0.29129 & 0.29087 & 0.29036 \\
	TabM / TabM-mini (envelope) & 0.29380 & 0.29252 & 0.29163 & \textbf{0.29104} & \textbf{0.29021} & 0.28984 \\
	\midrule
	TokenMoE+SSL (xlarge, full-data only) & -- & -- & -- & -- & -- & 0.29027 \\
	TabM-mini (xlarge-2, full-data only) & -- & -- & -- & -- & -- & 0.28903 \\
	TabM-mini (gigantic, full-data only) & -- & -- & -- & -- & -- & \textbf{0.28885} \\
\bottomrule
\end{tabular}
\end{adjustbox}
	\caption{Test Poisson deviance (lower is better) as a function of training fraction.
	For families we report the best configuration within the main sweep (trained at all fractions) at each fraction.
	Bold indicates the best value at each training fraction (including full-data-only extensions where present).
	The final rows highlight selected full-data-only runs, including the best transformer-family TokenMoE+SSL xlarge variant and the largest TabM-mini variants.}
	\label{tab:main-scaling}
\endgroup
\end{table}

\subsubsection{Crossover behavior (small-$N$ vs large-$N$)}
A notable crossover occurs around the mid-data regime:
\begin{itemize}
  \item \bl{At \textbf{5\% to 25\%}, the lowest seed-averaged ensemble deviances in the main-sweep envelope come from \textbf{Transformer+TabM},
  indicating that TabM-style adaptation inside the transformer improves sample-efficiency in these runs.}
  \item \bl{At \textbf{50\%}, a \textbf{TabM} variant is numerically best on the seed-averaged ensemble envelope, but the gap is too small to treat as a stable crossover. At \textbf{75\% and 100\%}, TabM variants have stronger seed-level support and the gap widens with scale.}
\end{itemize}
This is consistent with the qualitative observation that a purely supervised transformer objective underutilizes extra capacity,
while architectures or objectives that enforce robustness to feature transformations unlock better scaling.

\begin{table}[t]
\centering
\begingroup\color{blue}
\beautytable
\begin{adjustbox}{max width=\textwidth}
\begin{tabular}{@{}L{0.18\textwidth}L{0.24\textwidth}L{0.22\textwidth}L{0.30\textwidth}@{}}
\toprule
\textbf{Data regime} & \textbf{Fractions ($N$ range)} & \textbf{Numerical leader} & \textbf{Interpretation} \\
\midrule
Small-$N$ & 5\%--25\% ($\sim$0.2M--1.0M) & Transformer+TabM & Better sample efficiency (vertical shift / constant-factor gain). \\
Mid-$N$ & 50\% ($\sim$2.0M) & TabM (marginal) & Crossover point; gaps are within seed variability. \\
Large-$N$ & 75\%--100\% ($\sim$3.0M--4.0M) & TabM & Stronger scaling regime (slope) and increasing gap. \\
\bottomrule
\end{tabular}
\end{adjustbox}
\caption{\bl{Numerical regime summary from the main-sweep envelopes (Table~\ref{tab:main-scaling}). Small-$N$ leaders primarily reflect sample-efficiency improvements, while large-$N$ leaders reflect stronger scaling with data and capacity.}}
\label{tab:regime-winners}
\endgroup
\end{table}

\subsubsection{Crossover magnitudes relative to seed variability}
Because deviance differences between strong models can be small, it is useful to compare crossover gaps to training variance.
Table~\ref{tab:crossover-gaps} reports the difference between the best and second-best \emph{main-sweep} family envelopes at each fraction (Table~\ref{tab:main-scaling}),
along with \bl{a seed-level bootstrap interval for the paired repetition-level gap. For each fraction, we take the two configurations defining the ensemble envelope gap, compute the five paired single-seed differences (runner-up minus winner, matched by repetition index), and bootstrap the mean of those five differences using 50{,}000 nonparametric bootstrap resamples. This interval quantifies training-run variability in the existing repeated experiments; it is not an observation-level bootstrap over individual policies.}
\bl{The crossover at 50\% should not be over-interpreted: the ensemble gap is only 0.00004, and the paired single-seed differences actually favor Transformer+TabM rather than TabM.}
\bl{At larger fractions (75\% and 100\%), the seed-level intervals are positive, supporting the qualitative conclusion that TabM is the better-supported large-data leader in this experiment.}
More broadly, the ensemble gains in Table~\ref{tab:full-data-stability} and Figure~\ref{fig:stability-scaling} are often of similar magnitude to small-$N$ crossover gaps, suggesting that under data scarcity variance reduction (small ensembles) can be a more effective use of compute than increasing model size.

\begin{table}[t]
\centering
\begingroup\color{blue}
\beautytable
\begin{adjustbox}{max width=\textwidth}
\begin{tabular}{@{}rL{0.20\textwidth}L{0.20\textwidth}rrrr@{}}
\toprule
\textbf{$f$} & \textbf{Best} & \textbf{Runner-up} & \textbf{Ens. gap} & \textbf{Mean seed gap} & \textbf{95\% seed-bootstrap CI} & \textbf{Seeds} \\
\midrule
5\% & Transformer+TabM & MultiCLS transformer & 0.00039 & +0.00258 & [+0.00204, +0.00340] & 5/5 \\
10\% & Transformer+TabM & MultiCLS transformer & 0.00016 & +0.00111 & [+0.00094, +0.00130] & 5/5 \\
25\% & Transformer+TabM & TabM & 0.00016 & +0.00034 & [+0.00018, +0.00047] & 5/5 \\
50\% & TabM & Transformer+TabM & 0.00004 & -0.00027 & [-0.00057, -0.00005] & 0/5 \\
75\% & TabM & Transformer+TabM & 0.00048 & +0.00029 & [+0.00012, +0.00046] & 5/5 \\
100\% & TabM & TokenMoE+SSL & 0.00053 & +0.00112 & [+0.00098, +0.00126] & 5/5 \\
\bottomrule
\end{tabular}
\end{adjustbox}
\caption{\bl{Crossover magnitudes with seed-level bootstrap uncertainty, computed from main-sweep envelopes (Table~\ref{tab:main-scaling}).
Ens. gap is the runner-up ensemble deviance minus the best ensemble deviance (positive favors the reported ensemble leader).
Mean seed gap and the 95\% bootstrap interval are computed from five paired single-seed differences between the same two underlying configurations, using 50{,}000 nonparametric bootstrap resamples.
The final column counts how many of the five paired seed differences favor the reported ensemble leader.}}
\label{tab:crossover-gaps}
\endgroup
\end{table}

\subsection{Compute and parameter efficiency at full data}
\label{sec:results-compute}

To contextualize scaling in terms of resources, Table~\ref{tab:compute-full} reports parameter count and approximate training FLOPs per epoch
for representative full-data runs.
\bl{FLOPs/epoch is reported per trained model (one seed); since reported performance throughout uses the five-seed seed-averaged ensemble, the total training compute behind each ensemble point is approximately $5\times$ the per-run FLOPs/epoch.}
TabM achieves the best deviance but requires higher compute.
TokenMoE+SSL improves transformer scaling and reaches the best transformer-family score,
but remains behind large TabM-mini runs.

\begin{table}[t]
\centering
\begingroup\color{blue}
\beautytable
\begin{adjustbox}{max width=\textwidth}
\begin{tabular}{@{}L{0.44\textwidth}rrr@{}}
\toprule
\textbf{Model (full data)} & \textbf{Params (M)} & \textbf{FLOPs/epoch ($C_1$; one seed)} & \textbf{PoisDev} \\
\midrule
GLM & 0.0006 & $3.0\times 10^9$ & 0.29200 \\
FFN (very large) & 0.648 & $1.5\times 10^{13}$ & 0.29058 \\
Vanilla transformer (large) & 0.507 & $7.6\times 10^{14}$ & 0.29094 \\
TokenMoE transformer (large) & 1.081 & $1.7\times 10^{14}$ & 0.29074 \\
TokenMoE+SSL transformer (xlarge) & 4.082 & $1.3\times 10^{15}$ & 0.29027 \\
Transformer+TabM (multicls+TabM-mini, medium) & 0.293 & $6.4\times 10^{15}$ & 0.29056 \\
TabM (xlarge) & 2.962 & $5.1\times 10^{15}$ & 0.28984 \\
TabM-mini (xlarge-2) & 8.246 & $5.9\times 10^{15}$ & 0.28903 \\
TabM-mini (xlarge-3, gigantic) & 31.784 & $2.4\times 10^{16}$ & \textbf{0.28885} \\
\bottomrule
\end{tabular}
\end{adjustbox}
\caption{Full-data resource summary for representative models.
Params and \bl{$C_1$ FLOPs/epoch} are approximate and computed from the training graph \emph{per trained model} (one seed).
PoisDev is the \bl{five-seed seed-averaged ensemble} test deviance.
For comparisons that account for the training cost of ensembling, \bl{use $C_{\mathrm{ens}}=5C_1$ for the ensemble points; this constant factor does not affect fitted exponents. FLOPs are shown in scientific notation to keep the small GLM and large neural configurations readable on the same scale.} The largest TabM-mini run is the strongest model overall, but also the most compute-intensive.}
\label{tab:compute-full}
\endgroup
\end{table}

\begin{figure}[!htbp]
\centering
\IfFileExists{figures/compute_performance_figure.pdf}{%
\includegraphics[width=\textwidth]{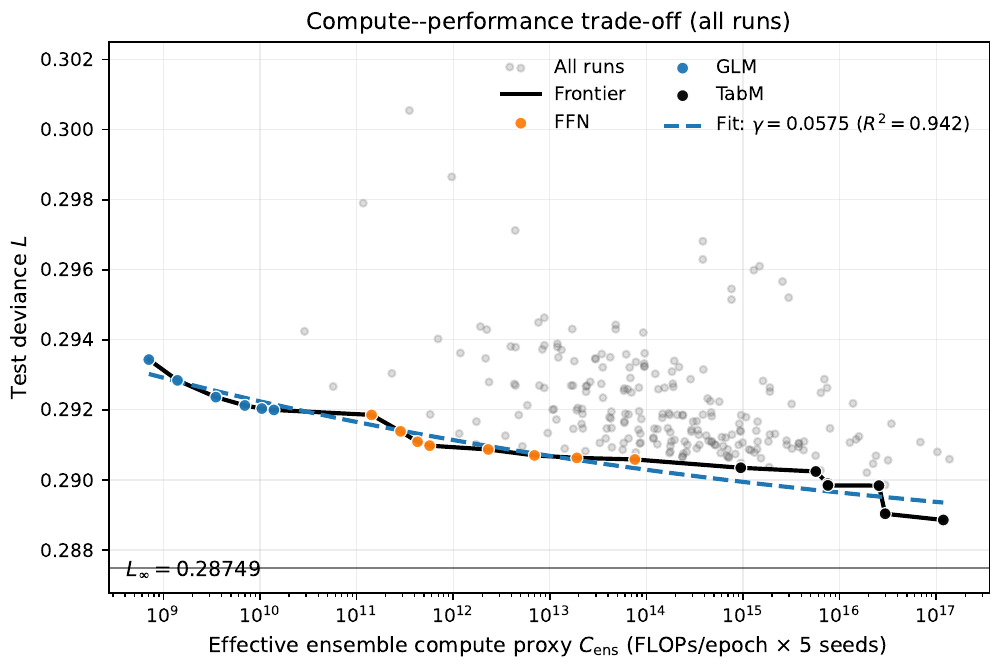}%
}{%
\fbox{\parbox{0.95\textwidth}{\centering Missing figure: \texttt{figures/compute\_performance\_figure.pdf}}}%
}%
\caption{Compute--performance trade-off across all runs. The x-axis is \bl{$C_{\mathrm{ens}}=5C_1$, the effective per-epoch compute proxy for the five independently trained seeds whose averaged predictions produce the reported ensemble deviance}; the y-axis is seed-averaged ensemble test deviance. Each point represents one model configuration. Grey points show all runs and the black curve traces the non-dominated (Pareto) compute frontier. The dashed line shows the fitted Kaplan-style power law \bl{$L(C)=L_\infty + A_C C^{-\gamma}$}.}
\label{fig:compute-performance}
\end{figure}

\begin{figure}[!htbp]
\centering
\IfFileExists{figures/param_performance_figure.pdf}{%
\includegraphics[width=\textwidth]{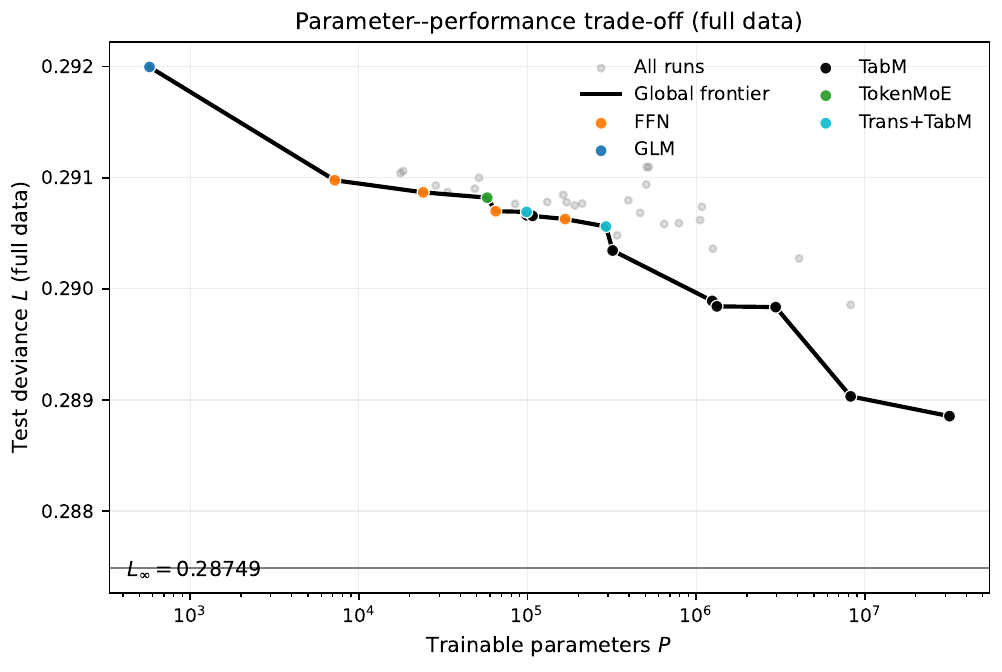}%
}{%
\fbox{\parbox{0.95\textwidth}{\centering Missing figure: \texttt{figures/param\_performance\_figure.pdf}}}%
}%
\caption{Parameter--performance trade-off at full data (seed-averaged ensemble test deviance). Grey points show all runs and the black curve traces the non-dominated (global Pareto) frontier in $(P,L)$. Frontier points are colored by family.}
\label{fig:param-performance}
\end{figure}

To make the ``optimal frontier'' view explicit in parameter space, Figure~\ref{fig:param-performance} plots the global full-data Pareto frontier of test deviance versus trainable parameters.
\bl{At small parameter counts the GLM is compute-efficient; in the mid-range the frontier includes FFNs, TokenMoE, and the Transformer+TabM hybrid; and at large parameter counts the observed frontier is occupied by TabM/TabM-mini configurations.}

\subsection{Kaplan-style and Chinchilla-style scaling fits}
\label{sec:results-scaling-laws}

\begin{figure}[p]
\centering
\IfFileExists{figures/kaplan_summary_figure.pdf}{%
\includegraphics[width=\textwidth,height=0.82\textheight,keepaspectratio]{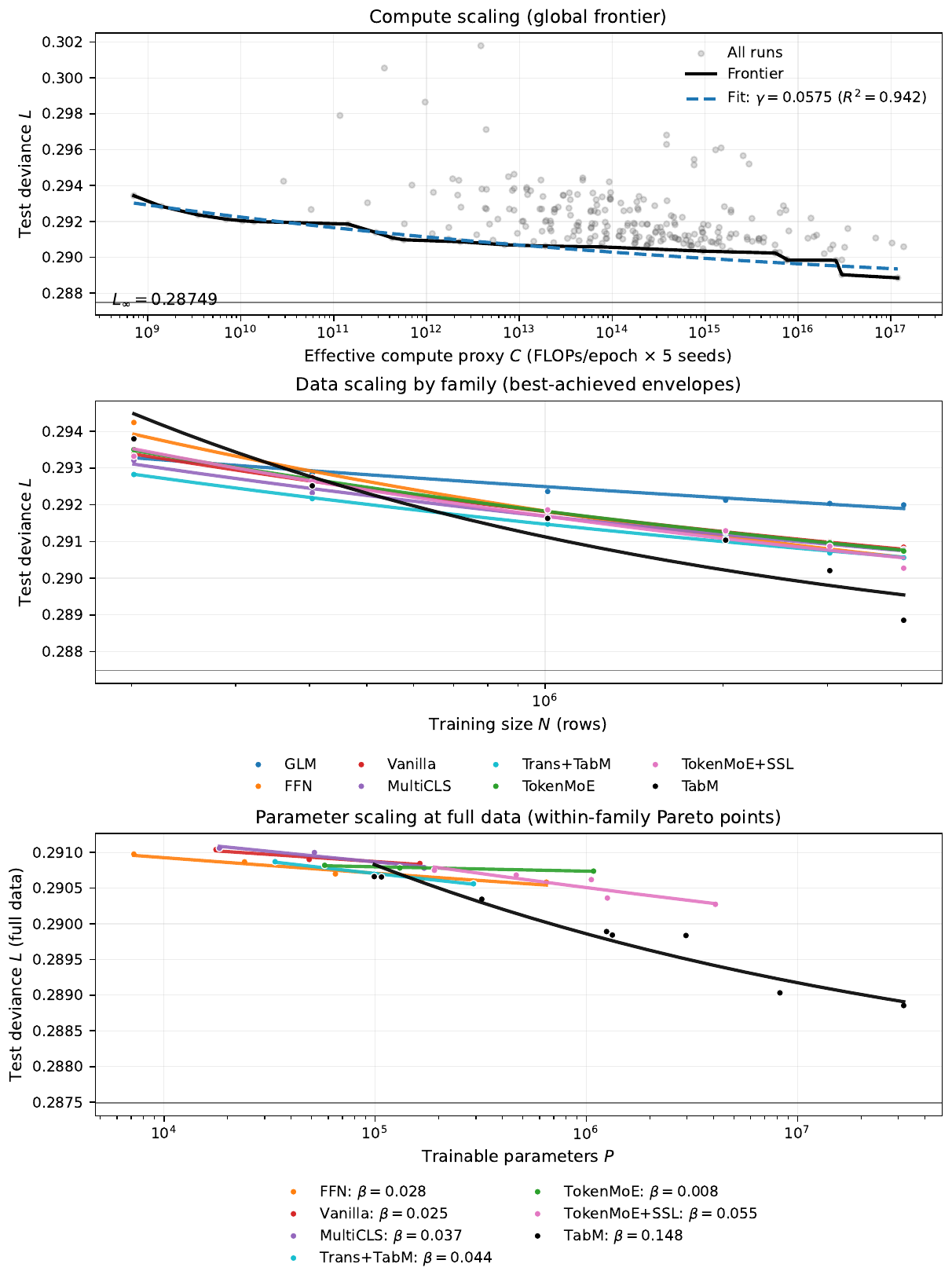}%
}{%
\fbox{\parbox{0.95\textwidth}{\centering Missing figure: \texttt{figures/kaplan\_summary\_figure.pdf}}}%
}%
\caption{Kaplan-style scaling fits.
Top: compute-efficient frontier fit \bl{$L(C)=L_{\infty}+A_C C^{-\gamma}$}, where \bl{$C=C_{\mathrm{ens}}=5C_1$ is the effective per-epoch compute proxy matching the five-seed ensemble deviance}.
Middle: best-of-family data scaling fits $L(N)-L_{\infty}\propto N^{-\alpha}$ (including full-data-only extensions when available).
Bottom: parameter scaling at full data using \emph{within-family} parameter--loss Pareto points, fit as $L(P)-L_{\infty}\propto P^{-\beta}$.
The strongest exponents are consistently observed for TabM.}
\label{fig:kaplan-summary}
\end{figure}

Figure~\ref{fig:kaplan-summary} summarizes our scaling-law fits in the style of Kaplan et al.\ and our compute-optimal discussion in the style of Chinchilla.
We fit standard ``power law plus floor'' relationships:
\begin{equation}
L(X) \approx L_{\infty} + A X^{-k},
\end{equation}
where $L_{\infty}$ is a fitted performance floor and $k$ is the \emph{scaling exponent} (a straight-line slope in log--log space after subtracting $L_{\infty}$).
Throughout, we distinguish three scaling questions and their corresponding exponents:
\begin{itemize}
  \item \textbf{Data scaling ($\alpha$):} \bl{$L(N)-L_{\infty}\approx A_N N^{-\alpha}$}, where $N$ is the number of training rows. Larger $\alpha$ means faster improvement with more data (doubling $N$ multiplies $L-L_{\infty}$ by $2^{-\alpha}$).
  \item \textbf{Parameter scaling ($\beta$):} \bl{$L(P)-L_{\infty}\approx A_P P^{-\beta}$} at fixed data, where $P$ is the number of trainable parameters. Larger $\beta$ means added capacity converts more reliably into generalization gains.
  \item \textbf{Compute scaling ($\gamma$):} $L(C)-L_{\infty}\approx A_C C^{-\gamma}$ along a compute-efficient frontier, where $C$ is an effective per-epoch compute proxy.
\end{itemize}
We fit a compute frontier over all runs and adopt the same fitted floor $L_{\infty}$ across subsequent fits.
Why these exponents matter: they separate ``more data helps'' ($\alpha$) from ``more parameters help'' ($\beta$), and they determine compute-optimal allocation rules when both can be increased under a compute budget (Section~\ref{sec:results-big-model-intuition}).

\subsection{Compute-optimality and the ``big model'' intuition}
\label{sec:results-big-model-intuition}

\bl{In compute-efficient scaling analyses of large Transformers, \citeA{Kaplan2020} emphasize the \emph{compute-optimal frontier}: under a fixed compute budget, it can be efficient to train larger models on relatively less data
and/or stop early, a point they summarize succinctly as ``big models may be more important than big data.'' Our experiment does not implement a full fixed-compute
training optimization (we primarily measure a per-epoch compute proxy and use early stopping rather than sweeping total training FLOPs), but our
results nonetheless allow us to evaluate a closely related question: \emph{when we spend more per-epoch compute at a fixed data fraction, does the compute-
efficient frontier shift toward larger models?}}

\bl{Empirically, the answer depends on the model family and data regime.} Along the global compute frontier (Figure~\ref{fig:kaplan-summary}, top), the compute-efficient set transitions from GLMs at extremely low compute to neural models (FFNs) and then to TabM/TabM-mini variants at the
highest compute levels, indicating that (in the multi-million-row regime) the best attainable deviance at higher compute is obtained by moving to higher-capacity
non-linear architectures rather than by further refining small models.

In the smallest-data regime, we observe a family-dependent version of the ``big model'' effect. For the Transformer+TabM hybrid, increasing model size
improves test deviance even at 5\%--10\% of the training rows (Table~\ref{tab:size_scaling_transformer_fixes}), and the hybrid occupies the compute
frontier at these fractions. In contrast, for FFNs and TabM/TabM-mini, increasing size at 5\% can \emph{hurt} (Table~\ref{tab:size_scaling_baselines};
Table~\ref{tab:size_scaling_tabm}), consistent with the negative/near-zero effective low-$N$ parameter sensitivity in Figure~\ref{fig:regime-diagnostics}
(right). \bl{In the data-scarce regime, bigger models are compute-efficient mainly when the architecture also improves sample efficiency and/or reduces optimization variance (here, the Transformer+TabM hybrid, and at larger $N$ TokenMoE+SSL).}

Finally, in the large-$N$ regime, TabM exhibits much stronger parameter scaling than supervised Transformers (Table~\ref{tab:beta_param_split}), and the
compute-optimal size within the TabM ladder shifts upward with $N$ (Table~\ref{tab:size_scaling_tabm}; Figure~\ref{fig:size-scaling}), aligning more
closely with the Kaplan-style intuition that the compute-efficient frontier increasingly favors larger-capacity models once sufficient data are available.

\subsubsection{Compute scaling (Kaplan-style)}
Along the compute-efficient frontier, test deviance follows a power law:
\begin{equation}
L(C) \approx L_{\infty} + A_C \, C^{-\gamma},
\end{equation}
	with $L_{\infty} \approx 0.28749$ and $\gamma \approx 0.0575$ (Figure~\ref{fig:kaplan-summary}, top; $R^2\approx 0.94$ on $(L-L_{\infty})$).
Because $\gamma$ is small, compute returns are slow: large multiplicative increases in training compute are required to produce visible absolute deviance gains.
In our experiment, the compute-efficient frontier is occupied by GLMs at low compute, transitions to FFNs, and then transitions to TabM/TabM-mini variants at the highest compute levels (Figure~\ref{fig:kaplan-summary}, top).

\subsubsection{Data scaling by family}
For each family, we fit
\begin{equation}
L(N) - L_{\infty} \approx A_N \, N^{-\alpha}.
\end{equation}
Using the main-sweep envelopes in Table~\ref{tab:main-scaling}, Table~\ref{tab:alpha-robustness} reports the fitted exponents and simple robustness diagnostics (piecewise fits and leave-one-fraction-out ranges).
This quantifies the core empirical finding: \textbf{TabM improves faster with data than supervised transformers.}
On the main sweep, supervised Transformers exhibit non-trivial data scaling ($\alpha\approx 0.18$--$0.19$), indicating clear improvement with more data, but TabM-style models improve faster and achieve lower asymptotic deviance at full data.
Moreover, the full-data-only TabM-mini extensions (Table~\ref{tab:main-scaling}) further lower deviance; if we include the best full-data TabM-mini point in the best-achieved envelope (as visualized in Figure~\ref{fig:kaplan-summary}, middle), the fitted exponent increases to $\alpha \approx 0.409$ and the log--log fit quality decreases (about $R^2\approx 0.80$), consistent with capacity crossovers and piecewise scaling rather than a single universal regime.

\begin{table}[t]
\centering
\begingroup\color{blue}
\beautytable
\begin{adjustbox}{max width=\textwidth}
\begin{tabular}{@{}L{0.30\textwidth}rrrr@{}}
\toprule
\textbf{Family} & $\boldsymbol{\alpha}$ & $\boldsymbol{\alpha_{\text{low}}}$ & $\boldsymbol{\alpha_{\text{high}}}$ & \textbf{LOO range} \\
\midrule
GLM & 0.091 & 0.122 & 0.041 & [0.075, 0.098] \\
FFN & 0.246 & 0.295 & 0.253 & [0.217, 0.256] \\
Vanilla transformer & 0.194 & 0.218 & 0.153 & [0.189, 0.199] \\
MultiCLS transformer & 0.182 & 0.192 & 0.178 & [0.171, 0.187] \\
Transformer+TabM & 0.183 & 0.182 & 0.229 & [0.180, 0.186] \\
TokenMoE & 0.202 & 0.198 & 0.215 & [0.200, 0.205] \\
TokenMoE+SSL & 0.220 & 0.179 & 0.398 & [0.197, 0.238] \\
TabM / TabM-mini & 0.309 & 0.259 & 0.601 & [0.283, 0.323] \\
\bottomrule
\end{tabular}
\end{adjustbox}
\caption{Robustness diagnostics for fitted data-scaling exponents using the main-sweep family envelopes in Table~\ref{tab:main-scaling} and a shared $L_{\infty}=0.28749$.
$\alpha_{\text{low}}$ fits only the first three fractions (5\%, 10\%, 25\%), while $\alpha_{\text{high}}$ fits only the last three fractions (50\%, 75\%, 100\%).
LOO is the leave-one-fraction-out range over six fits.}
\label{tab:alpha-robustness}
\endgroup
\end{table}

\subsubsection{Parameter scaling at fixed (full) data}
On the full-data Pareto set, we fit
\begin{equation}
L(P) - L_{\infty} \approx A_P \, P^{-\beta},
\end{equation}
and report the fitted exponents in Table~\ref{tab:beta_param_split}. The very small $\beta$ for vanilla Transformers and TokenMoE matches the observed ``weak scaling'' when increasing Transformer size without additional inductive bias or auxiliary training signal, while TokenMoE+SSL and TabM show markedly stronger parameter scaling.

\begin{table}[t]
\centering
\begingroup\color{blue}
\beautytable
\begin{adjustbox}{max width=\textwidth}
\begin{tabular}{@{}L{0.34\textwidth}rrr@{}}
\toprule
\textbf{Family} & $\boldsymbol{\beta}$ & $\boldsymbol{\beta_{\text{emb}}}$ & $\boldsymbol{\beta_{\text{non-emb}}}$ \\
\midrule
FFN & 0.028 & 0.063 & 0.026 \\
Vanilla transformer & 0.025 & 0.060 & 0.020 \\
MultiCLS transformer & 0.037 & 0.089 & 0.029 \\
Transformer+TabM & 0.044 & 0.102 & 0.040 \\
TokenMoE & 0.008 & 0.008 & 0.007 \\
TokenMoE+SSL & 0.055 & 0.053 & 0.058 \\
TabM / TabM-mini & 0.148 & 0.249 & 0.146 \\
\bottomrule
\end{tabular}
\end{adjustbox}
\caption{Full-data parameter scaling exponents on the full-data Pareto set, using a shared $L_{\infty}=0.28749$. We report scaling with total trainable parameters $P$ as well as the decomposition into trainable embedding parameters $P_{\text{emb}}$ and trainable non-embedding parameters $P_{\text{non-emb}}$. The decomposition should be interpreted descriptively, since $P_{\text{emb}}$ and $P_{\text{non-emb}}$ co-vary across the preset model sizes.}
\label{tab:beta_param_split}
\endgroup
\end{table}

\bl{This embedding/backbone decomposition is descriptive rather than causal. We include it because scaling-law studies commonly make parameter-accounting choices explicit, especially when some parameters belong to input embeddings and others belong to the model body \cite{Kaplan2020,Hoffmann2022}. In our experiment, $P_{\text{emb}}$ and $P_{\text{non-emb}}$ are not independently randomized; both are induced by the preset size ladders for each architecture family. The separate exponents should therefore be read as diagnostics of where parameters reside along those ladders, not as evidence that moving a parameter from embeddings to the backbone would by itself change generalization.}

The embedding/non-embedding split is especially relevant for tabular models because the definition of ``model size'' depends on feature tokenization and categorical cardinalities. On the full-data Pareto points used for Table~\ref{tab:beta_param_split}, TabM is strongly \emph{backbone-dominated}: only about 0.8\%--8.3\% of trainable parameters are embeddings (median 2.6\%). In contrast, TokenMoE and TokenMoE+SSL are \emph{embedding-dominated}: roughly 66\%--91\% of trainable parameters are embeddings (medians $\sim$74\% and $\sim$86\%, respectively), reflecting that in our Transformer implementations most parameters arise from the per-feature embedding tables rather than the attention blocks. FFNs and the baseline Transformer families lie between these extremes (roughly 3\%--52\% embedding share across their Pareto points).

This decomposition helps interpret why ``Transformer scaling'' is weak in our purely supervised setting. TokenMoE without SSL exhibits essentially flat parameter scaling ($\beta\approx 0.008$) even though its parameter count is largely embedding-based: simply increasing embedding capacity does not translate into lower deviance unless the architecture and training objective can reliably convert those representations into interaction signal. By contrast, TokenMoE+SSL shows materially larger exponents for both embedding and non-embedding subsets ($\beta_{\text{emb}}\approx 0.053$, $\beta_{\text{non-emb}}\approx 0.058$), consistent with the view that the auxiliary swap task improves optimization and representation learning throughout the model. TabM shows the strongest parameter scaling overall ($\beta\approx 0.148$), and because embeddings are a small share of its parameters, $\beta$ is close to $\beta_{\text{non-emb}}$, suggesting that its gains are primarily driven by scaling the non-embedding backbone and its lightweight ensemble adapters rather than by enlarging embedding tables.

\subsubsection{TabM exhibits a different scaling law than independent MLP ensembles}
\label{sec:results-tabm-vs-ensembles}

A natural hypothesis is that TabM's advantage is primarily an ensembling effect: if TabM behaved like ``$K$ good FFNs averaged'', improvements over a single FFN should be dominated by variance reduction, and the data-scaling slope should broadly match that of independent MLP ensembles.
Our results do not support this simple explanation.
TabM has the steepest data scaling exponent ($\alpha\approx 0.309$ on the main sweep and $\alpha\approx 0.409$ when including the largest full-data TabM-mini extension; Table~\ref{tab:alpha-robustness}), and it converts additional parameters into generalization gains far more effectively than FFNs and supervised Transformers ($\beta\approx 0.148$ versus $\beta\approx 0.028$ and $\beta\approx 0.025$; Table~\ref{tab:beta_param_split}).
Moreover, the gains from seed-averaging are comparatively small for TabM-style models (average gain 0.000360) relative to supervised attention-based models (0.001003 for vanilla Transformers; Figure~\ref{fig:stability-scaling}), and at full data the mean \emph{single-seed} TabM scores remain substantially better than the mean single-seed FFN scores (Table~\ref{tab:full-data-stability}).
Taken together, these facts indicate that TabM occupies a different scaling regime in this actuarial tabular setting: its improvements with $N$ and $P$ are not well explained by classic deep-ensemble variance averaging alone.

Mechanistically, this is consistent with TabM's design as a jointly trained, weight-sharing ensemble with lightweight adapters \cite{TabM}. In particular, several structural features plausibly contribute to its stronger scaling:
\begin{itemize}
  \item \textbf{Joint training of implicit submodels}: all members update a shared backbone in a single optimization run, so the learned representation is shaped by multiple slightly different ``views'' of the data rather than by one model trajectory.
  \item \textbf{Weight sharing with lightweight adapters}: shared kernels act as regularization while per-member adapters induce structured diversity that the mean prediction can exploit.
  \item \textbf{First-adapter tilting (TabM-mini)}: the initial non-shared adapter maps inputs into $K$ different representation spaces before feature mixing, improving sample efficiency in large-$N$ regimes.
  \item \textbf{Improved width utilization and smoother optimization}: gradients to shared weights aggregate across members, which can behave like a larger effective batch for the backbone and reduce wasted capacity.
\end{itemize}

A useful direction for future work is to isolate these effects via matched-$K$ comparisons to independent deep ensembles and TabM ablations (e.g., TabM vs TabM-mini and adapter-initialization variants), and by tracking per-member versus mean-prediction losses across $N$.

\subsubsection{Chinchilla-style compute-optimal implication}
Assuming compute scales approximately as $C \propto N P$, and the loss decomposes into independent power laws in $N$ and $P$,
the compute-optimal trade-off obeys $N_{\star} \propto P^{\beta/\alpha}$ and equivalently
$P_{\star} \propto C^{\alpha/(\alpha+\beta)}$ and $N_{\star} \propto C^{\beta/(\alpha+\beta)}$.
\bl{Appendix~\ref{app:chinchilla_explainer} gives the short derivation and notation translation from the original paper to our notation for this allocation rule.}
\bl{This is a per-epoch approximation: with early stopping, realized total training compute is better written as $C_{\mathrm{tot}}\propto N P\,E_{\mathrm{stop}}(P,N)$, so an exact stochastic-stopping optimum would require modelling the stopping epoch as part of the scaling law.}
Intuitively, $\alpha$ and $\beta$ quantify diminishing returns: $L-L_{\infty}$ falls as $N^{-\alpha}$ when increasing data at fixed capacity, and as $P^{-\beta}$ when increasing capacity at fixed data.
The ratio $\beta/\alpha$ therefore summarizes the compute-optimal \emph{data-to-parameter growth rule}: when $\beta/\alpha<1$, the compute-optimal trajectory grows parameters faster than data (since $N_{\star}$ increases sublinearly in $P_{\star}$), while larger $\beta/\alpha$ implies that data must grow more aggressively with parameters to remain compute-optimal.
Using $\beta$ together with the best-achieved TabM data exponent (including the full-data TabM-mini extension, $\alpha \approx 0.409$) gives $\beta/\alpha \approx 0.36$ for TabM (so $P_{\star} \propto C^{0.73}$ and $N_{\star} \propto C^{0.27}$), while transformers and FFNs yield ratios near $\approx 0.12$,
suggesting that (within this experimental regime) \textbf{TabM benefits from joint increases in both data and parameters more than transformers do}.
Because TabM is backbone-dominated, this conclusion is essentially unchanged if we replace $\beta$ by $\beta_{\text{non-emb}}$ (Table~\ref{tab:beta_param_split}); for embedding-dominated Transformer variants, $\beta_{\text{non-emb}}$ remains small without SSL, so larger attention backbones alone do not materially change the compute-optimal trade-off in our setting.

\begin{figure}[p]
\centering
\IfFileExists{figures/chinchilla_fit_tabm.pdf}{%
\includegraphics[width=\textwidth,height=0.82\textheight,keepaspectratio]{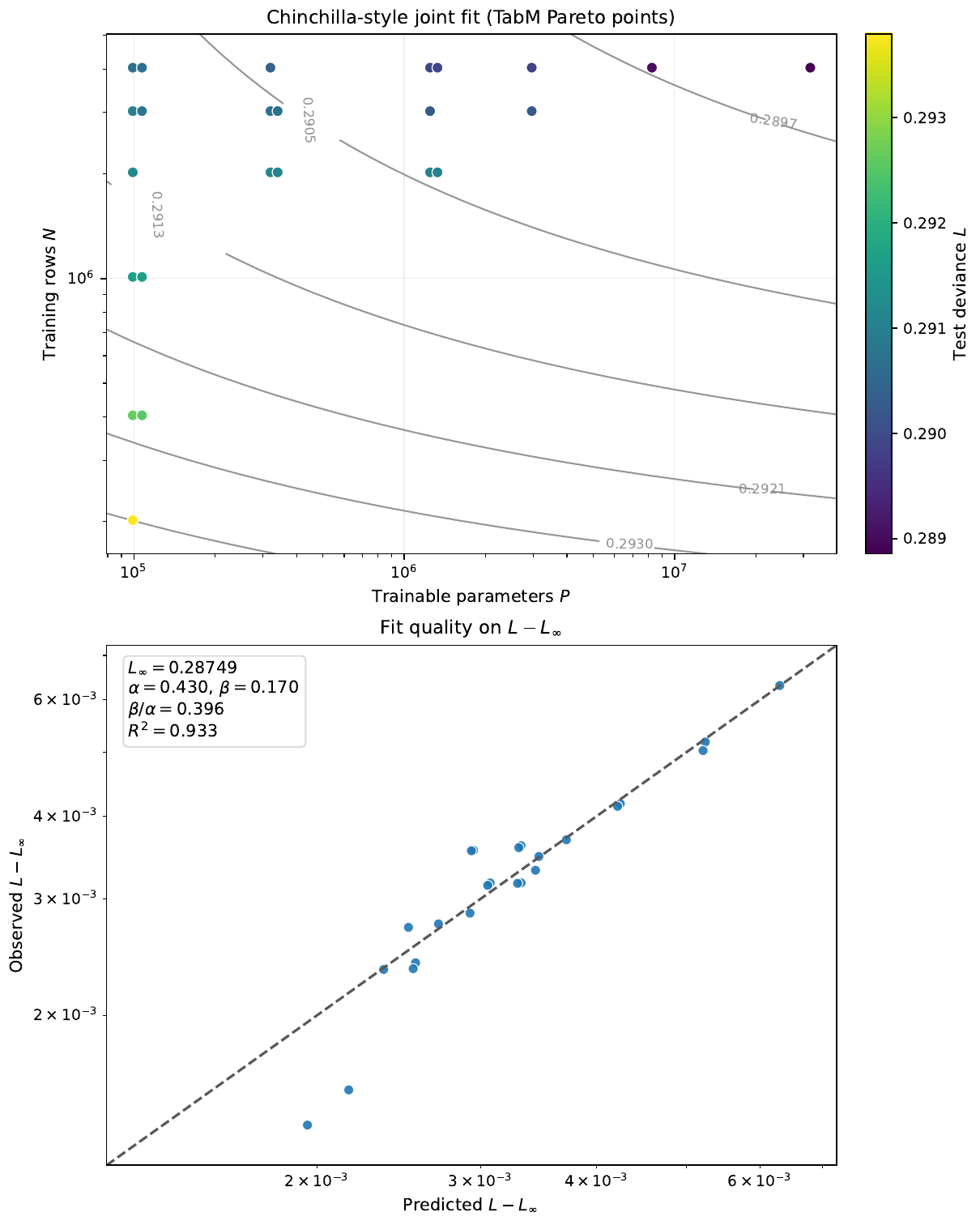}%
}{%
\fbox{\parbox{0.95\textwidth}{\centering Missing figure: \texttt{figures/chinchilla\_fit\_tabm.pdf}}}%
}%
\caption{Chinchilla-style bivariate fit for TabM/TabM-mini.
We fit \bl{$L(P,N)=L_{\infty} + A_N N^{-\alpha} + A_P P^{-\beta}$} using the per-threshold TabM/TabM-mini parameter--loss Pareto points (so each point reflects a configuration where increased parameters improves deviance at that data fraction).
The fitted exponents imply a compute-optimal allocation that grows parameters faster than data under $C\propto NP$. \bl{Because training uses early stopping, this allocation is a first-order per-epoch compute heuristic rather than an exact total-training-FLOPs optimum.}}
\label{fig:chinchilla-fit-tabm}
\end{figure}

As a single bivariate check in the style of \citeA{Hoffmann2022}, we also fit the additive model
\bl{$L(P,N)\approx L_{\infty} + A_N N^{-\alpha} + A_P P^{-\beta}$} directly on the TabM/TabM-mini per-threshold parameter--loss Pareto points.
This yields $\alpha \approx 0.430$ and $\beta \approx 0.170$ ($R^2\approx 0.933$ on $L-L_{\infty}$), implying $\beta/\alpha \approx 0.396$ and a compute-optimal split $P_{\star} \propto C^{0.716}$ and $N_{\star} \propto C^{0.284}$ under $C\propto NP$ (Figure~\ref{fig:chinchilla-fit-tabm}).
\bl{We therefore interpret this allocation as a first-order per-epoch compute heuristic rather than an exact total-training-FLOPs optimum.}
This supports the ``big model'' intuition of compute-efficient scaling analyses in the large-$N$ regime: as compute increases, the compute-optimal allocation grows model size faster than data volume, while still increasing both.

\begin{table}[t]
\centering
\begingroup\color{blue}
\beautytable
\begin{adjustbox}{max width=\textwidth}
\begin{tabular}{@{}L{0.21\textwidth}L{0.20\textwidth}L{0.14\textwidth}ccccL{0.16\textwidth}@{}}
\toprule
\textbf{Study} & \textbf{Domain/task} & \textbf{Metric} & $\boldsymbol{\alpha}$ \textbf{(data)} & $\boldsymbol{\beta}$ \textbf{(params)} & $\boldsymbol{\gamma}$ \textbf{(compute)} & $\boldsymbol{\beta/\alpha}$ & \textbf{Notes} \\
\midrule
Kaplan et al.\ (2020) & Language modelling (LLM scaling) & Cross-entropy & 0.095 & 0.076 & 0.050 & 0.80 & Token ratio decreases with $P$ under compute-optimal allocation. \\
Hoffmann et al.\ (2022) & Compute-optimal LLMs (Chinchilla) & Cross-entropy & -- & -- & -- & $\approx 1$ & Compute-optimal training uses much more data; empirically $D_{\star}\approx kP$ with $k\approx 20$ tokens/parameter. \\
This paper (vanilla transformer) & Motor frequency (tabular) & Poisson deviance & 0.194 & 0.025 & 0.0575 & 0.13 & Best-of-size envelope in Table~\ref{tab:main-scaling}. \\
This paper (TokenMoE+SSL) & Motor frequency (tabular) & Poisson deviance & 0.220 & 0.055 & 0.0575 & 0.25 & Best Transformer-family envelope (Table~\ref{tab:main-scaling}). \\
This paper (TabM/TabM-mini) & Motor frequency (tabular) & Poisson deviance & 0.409 & 0.148 & 0.0575 & 0.36 & $\alpha$ uses the best-achieved TabM point including full-data TabM-mini extensions. \\
\bottomrule
\end{tabular}
\end{adjustbox}
\caption{Scaling-law comparison across domains. Kaplan et al.\ (2020) and Hoffmann et al.\ (2022) report LLM cross-entropy scaling in token space; this paper reports Poisson deviance for tabular frequency prediction. Exponents are not directly comparable across metrics; the table contextualizes relative scaling strength and the implied compute-optimal data--parameter balance. The ratio $\beta/\alpha$ summarizes the compute-optimal trade-off under an additive law \bl{$L(P,N)\approx L_\infty + A_P P^{-\beta} + A_N N^{-\alpha}$} with compute \bl{$C\propto P N$}.}
\label{tab:scaling_compare_llm}
\endgroup
\end{table}

\subsubsection{A pooled multi-family scaling fit (global exponents, family terms)}
\begin{figure}[p]
\centering
\IfFileExists{figures/global_family_scaling_fit.pdf}{%
\includegraphics[width=\textwidth,height=0.82\textheight,keepaspectratio]{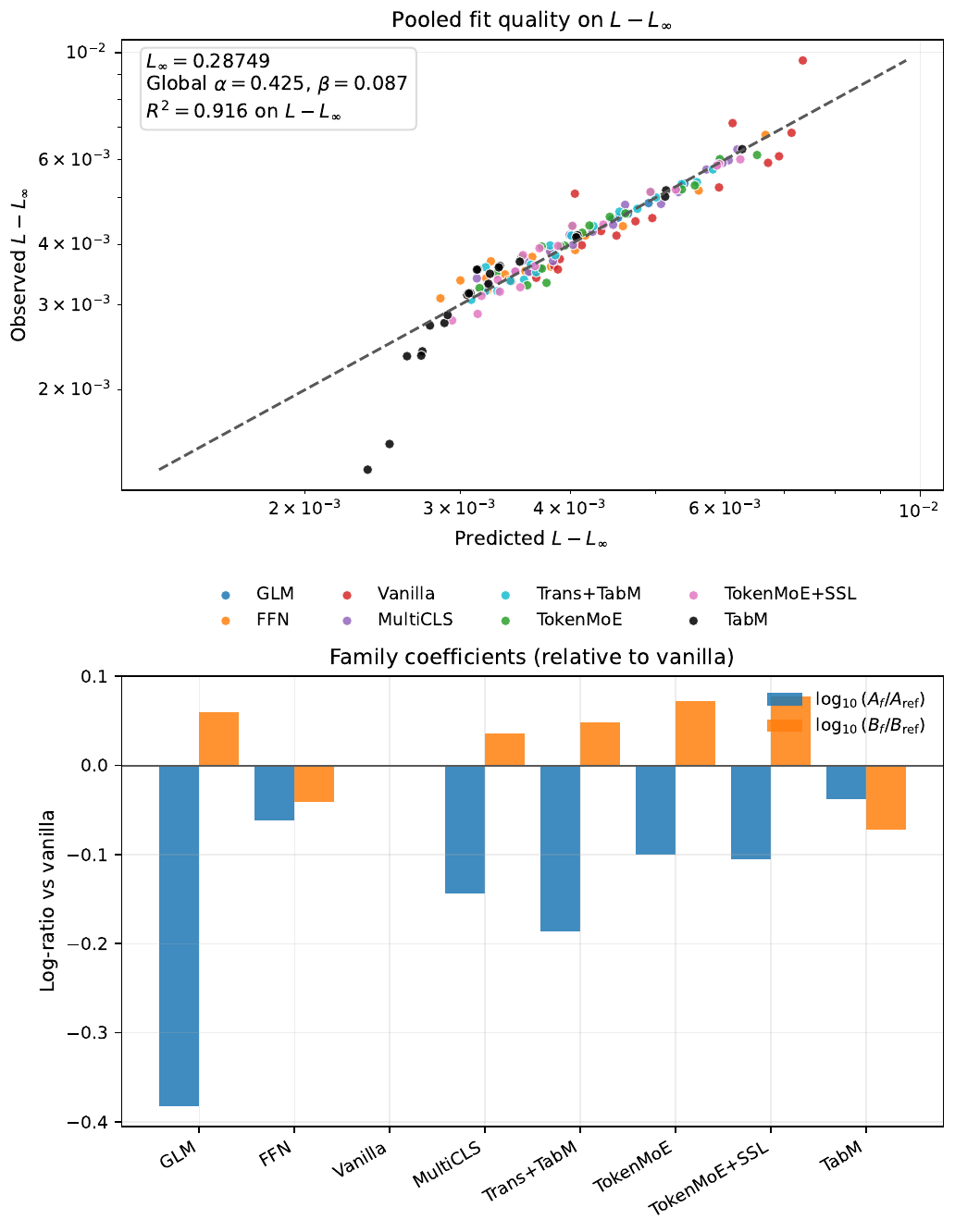}%
}{%
\fbox{\parbox{0.98\textwidth}{\centering Missing figure: \texttt{figures/global\_family\_scaling\_fit.pdf}}}%
}%
\caption{Global pooled scaling fit with family terms.
\bl{We fit $L(P,N,f) \approx L_{\infty} + A_{N,f} N^{-\alpha} + A_{P,f} P^{-\beta}$} on the ensemble deviance (5-seed prediction averaging), using per-threshold parameter--loss Pareto points within each family.
Left: observed versus predicted $(L-L_{\infty})$.
Right: fitted family coefficients relative to the vanilla transformer baseline.}
\label{fig:global-family-scaling-fit}
\end{figure}

As a complementary ``one-model'' summary across all architectures, we fit a single pooled scaling law with \emph{global} exponents but \emph{family-specific} coefficients,
\begin{equation}
L(P,N,f) \approx L_{\infty} + A_{N,f} N^{-\alpha} + A_{P,f} P^{-\beta},
\end{equation}
where $N$ is the number of training rows, $P$ is the number of trainable parameters, $f$ indexes the model family, and \bl{$(A_{N,f},A_{P,f})$} are family-specific coefficients (vertical shifts) that summarize relative sample efficiency \bl{($A_{N,f}$)} and parameter efficiency \bl{($A_{P,f}$)} under shared global exponents $(\alpha,\beta)$.
using the per-threshold parameter--loss Pareto points within each family (to avoid mixing in configurations where increasing parameters worsens loss at that fraction).
This pooled fit yields $\alpha \approx 0.425$ and $\beta \approx 0.087$ with $R^2\approx 0.916$ on $L-L_{\infty}$ (Figure~\ref{fig:global-family-scaling-fit}).
Interpreting this additive form under $C\propto NP$ gives a compute-optimal split $P_{\star} \propto C^{0.830}$ and $N_{\star} \propto C^{0.170}$, reflecting a strongly parameter-heavy optimum in this pooled summary.
\bl{As above, this allocation uses the per-epoch $NP$ proxy and is approximate under early stopping.}
We emphasize that this is a descriptive fit across heterogeneous architectures: it does not replace the family-wise exponents reported above, and it will not capture regime changes such as small-data traps or objective-induced shifts (e.g., SSL).

\subsection{Regime diagnostics: when do compute and capacity matter?}
\label{sec:results-regime}
\begin{figure}[p]
\centering
\IfFileExists{figures/regime_diagnostics_figure.pdf}{%
\includegraphics[width=\textwidth,height=0.82\textheight,keepaspectratio]{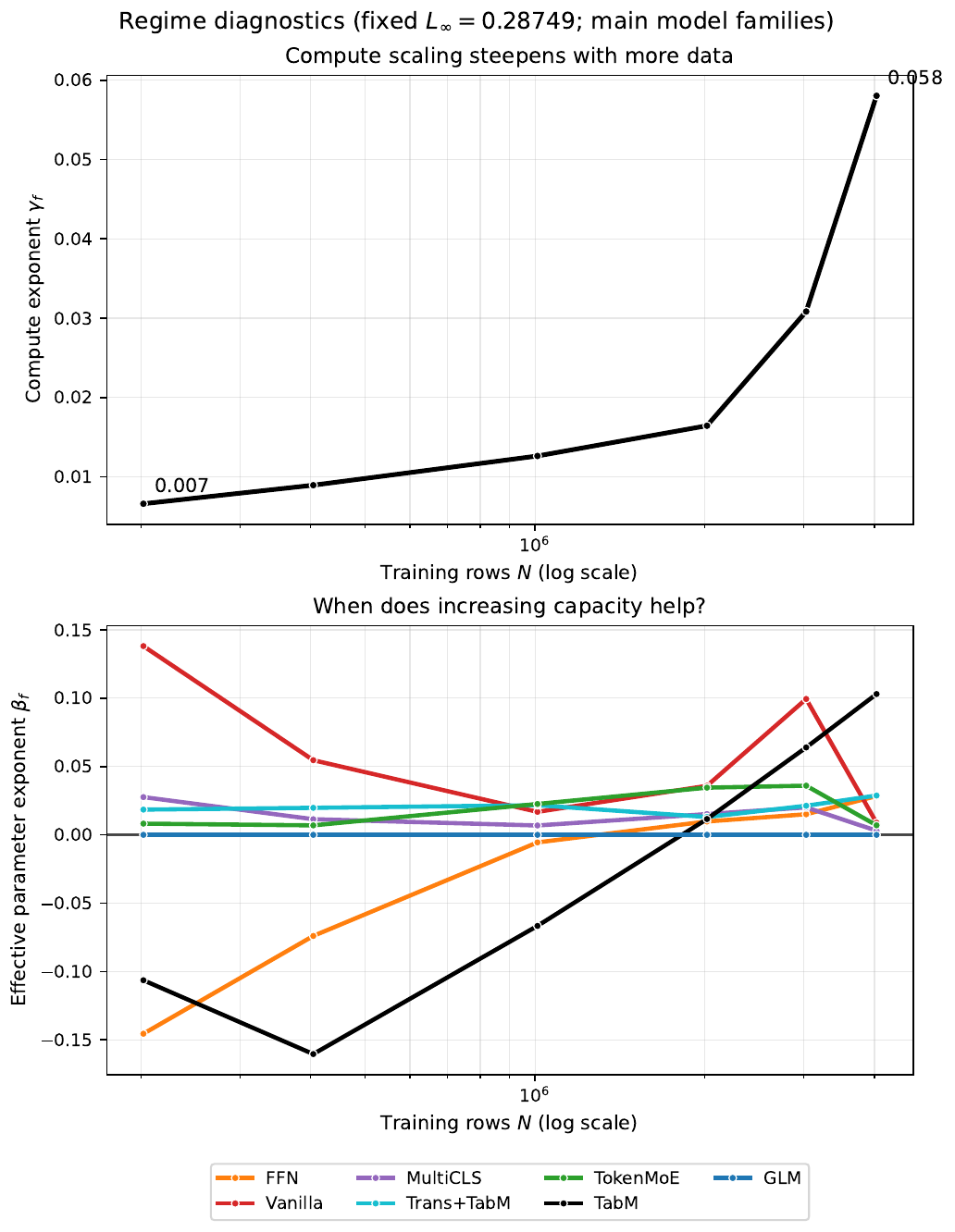}%
}{%
\fbox{\parbox{\textwidth}{\centering Missing figure: \texttt{figures/regime\_diagnostics\_figure.pdf}}}%
}%
\caption{Regime diagnostics using $L_{\infty}=0.28749$.
\bl{Top:} fraction-specific compute exponent $\gamma_f$ fitted on the compute-efficient frontier at each training fraction.
\bl{Bottom:} fraction-specific effective parameter exponent $\beta_f$ within each family, estimated across the standard size ladder; $\beta_f<0$ indicates that increasing capacity hurts in that regime, while $\beta_f>0$ indicates beneficial parameter scaling.}
\label{fig:regime-diagnostics}
\end{figure}

The fitted exponents in Section~\ref{sec:results-scaling-laws} summarize scaling behavior over the entire range of $N$.
However, actuarial applications often operate in mixed regimes (e.g., $10^5$--$10^6$ versus multi-million rows), and in such regimes it is informative to examine how \emph{compute sensitivity} and \emph{capacity sensitivity} change with data size.
Figure~\ref{fig:regime-diagnostics} reports two simple diagnostics derived directly from our experiment logs.

First, we fit the compute frontier separately at each training fraction and estimate a fraction-specific compute exponent $\gamma_f$.
Compute scaling is nearly flat in the smallest-data regime ($\gamma_{5\%}\approx 0.0066$), but steepens substantially at full data ($\gamma_{100\%}\approx 0.0580$),
consistent with a data-limited regime at small $N$ and a compute-limited regime only emerging once sufficient data are available.

Second, we estimate a fraction-specific \emph{effective parameter exponent} $\beta_f$ within each family by regressing $\log(L-L_{\infty})$ against $\log(P)$ across the standard size ladder.
For TabM and FFNs, $\beta_f$ is negative at small fractions (larger models can hurt when data are scarce) and becomes positive at larger fractions, consistent with the ``small-data trap'' intuition.
In contrast, supervised Transformer variants exhibit $\beta_f$ near zero across fractions, reflecting weaker parameter scaling unless augmented with additional inductive bias.

\subsection{Why transformers improve when we add TabM ideas or SSL}
\label{sec:results-why}

A purely supervised objective on tabular Transformers appears to produce limited returns from added capacity.
Two observations sharpen this finding.

First, \textbf{Transformers show meaningful data scaling but weak parameter scaling under purely supervised training.} Vanilla and MultiCLS Transformers exhibit non-trivial data scaling exponents ($\alpha \approx 0.18$--$0.20$; Table~\ref{tab:alpha-robustness}), indicating that additional training rows do improve generalization. However, their parameter scaling exponents are tiny ($\beta \approx 0.025$--$0.037$; Table~\ref{tab:beta_param_split}), meaning that increasing model size---more layers, wider embeddings, additional heads---yields negligible gains. \bl{The limiting factor is the model's ability to convert capacity into useful representations.}

Second, \bl{\textbf{the largest Transformer improvement comes from additional learning signal.}} TokenMoE without SSL has $\beta \approx 0.008$---essentially flat parameter scaling despite its sophisticated routing mechanism. Adding the swap-style SSL objective jumps parameter scaling to $\beta \approx 0.055$, a nearly 7$\times$ improvement (Table~\ref{tab:beta_param_split}). \bl{This is our cleanest evidence that loss function objective changes are central to unlocking Transformer scaling in the tabular regime.}

Two modifications change this baseline behavior:

\subsubsection{TabM-style adaptation inside the transformer}
Adding TabM mechanisms (adaptive transformations over token embeddings under input perturbations) yields the best models in the small-$N$ regime
(Table~\ref{tab:main-scaling}).
Empirically, this behaves like injecting a structured form of feature adaptability that improves sample efficiency.

\subsubsection{Self-supervised learning (SSL) via feature corruption prediction}
Adding an auxiliary objective that predicts whether a given feature column has been corrupted by a row-swap transformation
\bl{adds a column-level signal about ``what changed'' alongside the supervised target.}
In practice, TokenMoE+SSL delivers the best Transformer-family performance at full data (0.29027),
and shows a more consistent improvement with increasing model size than supervised Transformers.

Taken together, these results support the view that \textbf{scaling tabular Transformers requires additional inductive bias or training signal}
beyond the supervised likelihood, especially to make effective use of increased parameter counts.

\subsection{Variance and the benefit of seed-averaged ensembles}
\label{sec:results-variance}

Training variance is non-trivial even at full data.
For example, for a large TabM-mini configuration at full data, 5 independent seeds yield a mean test deviance of 0.28939 with
standard deviation 0.00021, while the seed-averaged ensemble improves to 0.28903 (an absolute gain of 0.00035).
We therefore report seed-averaged ensemble scores throughout as a low-cost, high-impact variance reduction technique.
For readers interested in non-ensemble performance, Table~\ref{tab:full-data-stability} also reports the mean single-seed test deviance; this is approximately the ensemble test deviance plus the reported ensemble gain.
Because the Poisson negative log-likelihood (and hence Poisson deviance) is convex in the predicted mean, averaging predictions across seeds can systematically reduce loss when predictions vary, and can therefore affect which configuration appears best within a family when models are high-variance.

\begin{table}[t]
\centering
\begingroup\color{blue}
\beautytable
\begin{adjustbox}{max width=\textwidth}
\begin{tabular}{@{}L{0.34\textwidth}rrrrrr@{}}
\toprule
\textbf{Model (full data)} & \textbf{Train} & \textbf{Test} & \textbf{Mean seed test} & \textbf{Gap} & $\boldsymbol{\sigma_{\text{seed}}}$ & \textbf{Ens.\ gain} \\
\midrule
GLM & 0.29018 & 0.29200 & 0.29201 & 0.00182 & 0.00002 & 0.00002 \\
FFN (very large) & 0.28661 & 0.29058 & 0.29118 & 0.00398 & 0.00009 & 0.00060 \\
Vanilla transformer (medium) & 0.28792 & 0.29085 & 0.29138 & 0.00293 & 0.00004 & 0.00054 \\
MultiCLS transformer (medium) & 0.28781 & 0.29078 & 0.29136 & 0.00297 & 0.00004 & 0.00058 \\
Transformer+TabM (medium) & 0.28726 & 0.29056 & 0.29065 & 0.00330 & 0.00006 & 0.00009 \\
TokenMoE transformer (large) & 0.28702 & 0.29074 & 0.29144 & 0.00372 & 0.00014 & 0.00071 \\
TokenMoE+SSL (large) & 0.28548 & 0.29036 & 0.29114 & 0.00488 & 0.00012 & 0.00078 \\
TokenMoE+SSL (xlarge) & 0.28479 & 0.29027 & 0.29111 & 0.00548 & 0.00008 & 0.00084 \\
TabM (xlarge) & 0.28000 & 0.28984 & 0.29003 & 0.00984 & 0.00013 & 0.00019 \\
TabM-mini (xlarge-2) & 0.26840 & 0.28903 & 0.28939 & 0.02063 & 0.00020 & 0.00035 \\
TabM-mini (gigantic) & 0.26182 & 0.28885 & 0.28936 & 0.02703 & 0.00009 & 0.00051 \\
\bottomrule
\end{tabular}
\end{adjustbox}
\caption{Full-data stability diagnostics.
Train/test/gap report seed-averaged ensemble Poisson deviance.
Mean seed test and $\sigma_{\text{seed}}$ summarize the distribution of \emph{single-seed} test deviances across 5 independent training runs.
Ens.\ gain is mean(seed test) $-$ ensemble test (positive indicates that averaging predictions improves performance).}
\label{tab:full-data-stability}
\endgroup
\end{table}

Table~\ref{tab:full-data-stability} highlights two complementary stability facts. First, supervised transformer variants exhibit relatively larger variance across random seeds and substantially larger ensemble gains than GLMs and TabM-style models, suggesting that their performance is more sensitive to optimization noise and that prediction averaging is especially valuable for attention-based architectures. Second, TabM and TabM-mini achieve the strongest test scores but also exhibit much larger train--test gaps at high capacity, indicating aggressive fit on the training distribution; nevertheless, the multi-million-row regime provides enough signal for these larger models to generalize, and the additional gains from ensembling are comparatively smaller.

\begin{figure}[p]
\centering
\IfFileExists{figures/stability_scaling_figure.pdf}{%
\includegraphics[width=\textwidth,height=0.82\textheight,keepaspectratio]{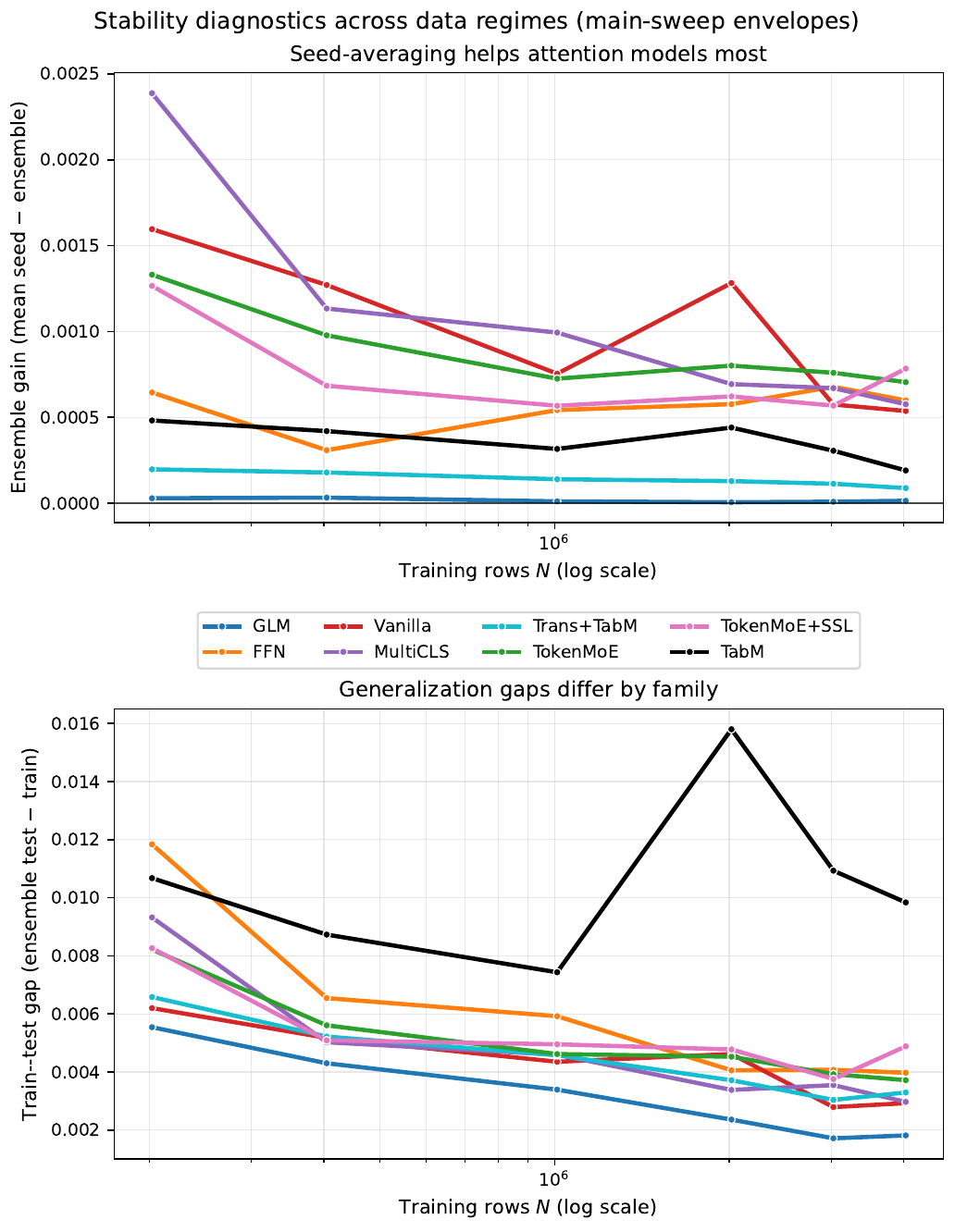}%
}{%
\fbox{\parbox{\textwidth}{\centering Missing figure: \texttt{figures/stability\_scaling\_figure.pdf}}}%
}%
\caption{Stability diagnostics across data regimes (main-sweep envelopes).
\bl{Top:} ensemble gain, defined as mean(single-seed test deviance) $-$ seed-averaged ensemble test deviance (positive indicates improvement from averaging predictions across seeds).
\bl{Bottom:} train--test gap for the seed-averaged ensemble.}
\label{fig:stability-scaling}
\end{figure}

\noindent
Figure~\ref{fig:stability-scaling} extends these stability diagnostics across all training fractions using the main-sweep family envelopes.
Ensemble gains are largest for supervised attention-based models (average gain 0.001003 for vanilla Transformers versus 0.000018 for the GLM), while TabM exhibits substantially larger train--test gaps (0.00984 at full data) yet achieves the best test deviance.
\bl{Notably, at full data the vanilla Transformer has a much smaller gap than TabM (0.00293 vs 0.00984) despite worse test deviance, consistent with representation/optimization limits as the main explanation.}

\section{\texorpdfstring{\bl{Conclusion}}{Conclusion}}
\label{sec:discussion}

\subsection{\texorpdfstring{\bl{Discussion}}{Discussion}}
\label{sec:discussion-sub}

The central finding of this paper is that \emph{neural scaling laws exist in actuarial tabular data}. Across all model families, test Poisson deviance decreases smoothly and predictably as training data increases, following power-law relationships that parallel those observed in large-language-model research - but here applied to a multi-million-row motor insurance dataset. This is, to our knowledge, the first systematic demonstration of Kaplan-style and Chinchilla-style scaling behavior in a production-scale actuarial setting.

Three main conclusions emerge from our experiments. First, scaling behavior differs materially by architecture family. TabM-style models, especially TabM-mini, exhibit the steepest improvement with data ($\alpha\approx 0.31$--$0.41$) and achieve the lowest full-data deviance, while purely supervised Transformers show weaker returns unless augmented with additional inductive bias. Second, the ``best'' model family depends on the data regime: \bl{a Transformer+TabM hybrid has the lowest main-sweep envelope at 5\%--25\% of the data, while TabM is only marginally ahead on the 50\% ensemble envelope and becomes the better-supported leader at 75\%--100\% once seed-level uncertainty is considered} (Table~\ref{tab:main-scaling}). Strikingly, larger-capacity models can improve performance even at small training fractions when paired with the right inductive bias, a counterintuitive result given the conventional wisdom that small data favors small models.

Third, to achieve these results we introduced several novel Transformer modifications for tabular data - including MultiCLS pooling, layer-wise value embeddings (the ``v stream''), TokenMoE routing, and swap-style self-supervision - and studied a Transformer+TabM hybrid that combines attention-based representation learning with the ensemble-style structure of TabM (\bl{Section~\ref{sec:model_descriptions}}). While these enhancements improve Transformer scaling relative to vanilla baselines, none match the strongest TabM-mini runs at full data, indicating that inductive bias choice remains the dominant factor in this regime.

Finally, variability across random seeds remains non-trivial even at full data, and inference-time seed averaging provides consistent improvements - especially for attention-based models. The average ensemble gain is roughly 50$\times$ larger for supervised Transformers (0.001003) than for the GLM (0.000018), while TabM achieves the best test deviance with comparatively smaller ensemble gains (0.000360) but larger train--test gaps (Figure~\ref{fig:stability-scaling}; Table~\ref{tab:full-data-stability}). \bl{This pattern - smaller generalization gaps yet worse test performance for Transformers - points to representation or optimization capacity as the limiting factor.} Understanding and correcting this limitation is an avenue for future work.

For practitioners, the central implication of our findings is a \emph{regime shift}: as training data grows from $\sim 10^5$ rows into the multi-million range, the compute-optimal choice changes. In smaller data regimes, GLMs and other strongly regularized baselines remain difficult to beat - they are statistically efficient, stable under resampling, and cheap to train. In the multi-million regime studied here (up to $\sim 4\times 10^6$ training rows), non-linear tabular architectures continue to improve meaningfully with additional data, compute-optimal capacity increases, and the case for higher-capacity tabular deep models strengthens. Our fitted scaling exponents quantify this: TabM-style models convert additional data and parameters into generalization gains far more effectively than purely supervised Transformers, and the compute-optimal allocation for TabM grows model size faster than data volume under $C\propto NP$ (Section~\ref{sec:results-big-model-intuition}).

\bl{The allocation result is more immediately actionable for model capacity than for labelled data volume. An insurer generally cannot create additional labelled policy-period rows on demand: frequency outcomes require written policies, exposure, and claims emergence. Increasing capacity or training compute is operationally more available, although governance, monitoring, inference cost, and validation burden still constrain deployment. A practical way to loosen the data bottleneck is to use policy or quote records without emerged claim outcomes for self-supervised representation learning, as in the swap-style auxiliary objective studied here. Such data can be available at larger scale than fully labelled claim outcomes and can add structure before supervised frequency fitting.}

\begin{figure}[!htbp]
\centering
\IfFileExists{figures/practical_significance_figure.pdf}{%
\includegraphics[width=\textwidth]{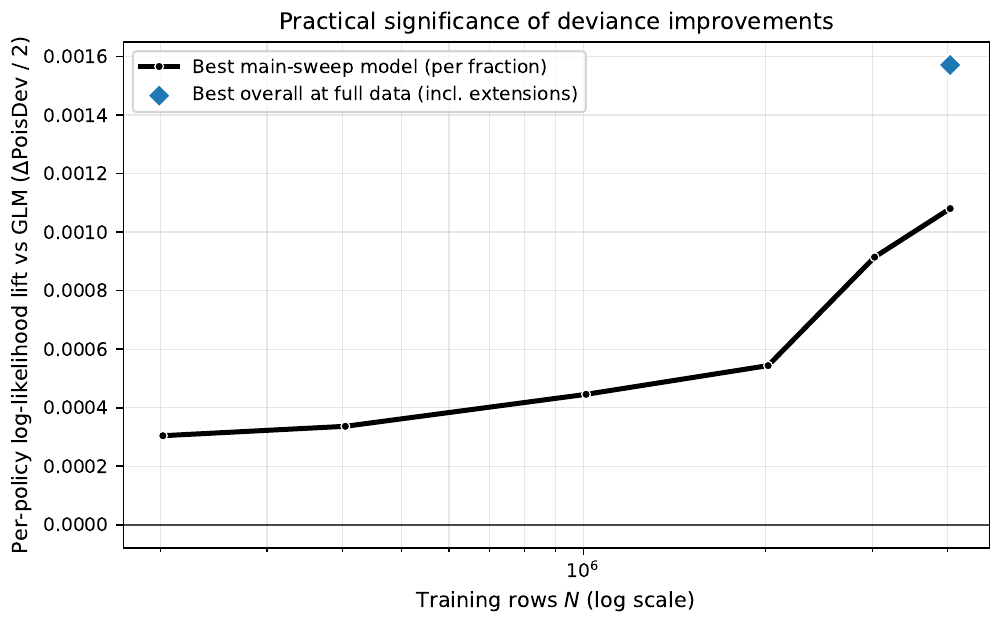}%
}{%
\fbox{\parbox{0.9\textwidth}{\centering Missing figure: \texttt{figures/practical\_significance\_figure.pdf}}}%
}%
\caption{Per-policy log-likelihood lift versus GLM implied by deviance improvements ($\Delta$PoisDev$/2$).
The line uses the best main-sweep model at each fraction; the full-data marker includes full-data-only extensions.}
\label{fig:practical-significance}
\end{figure}

\bl{Poisson deviance differences correspond directly to log-likelihood lift. For the mean deviance reported in this paper, an improvement of $\Delta$ deviance is an average per-policy held-out log-likelihood lift of $\Delta/2$ nats, meaning $\Delta/2$ units on the natural-logarithm scale (Section~\ref{sec:rationale_explainer}). At full data, TabM (xlarge) improves over the GLM by 0.00216 deviance (0.00108 per-policy log-likelihood), and the best TabM-mini full-data extension improves by 0.00315 deviance (0.00158 per-policy log-likelihood). Using the five repeated single-seed runs as a seed-level uncertainty check, the mean single-seed full-data TabM improvement over the GLM is 0.00198 deviance with bootstrap 95\% CI [0.00184, 0.00208], and the mean single-seed full-data TabM-mini gigantic improvement over the GLM is 0.00265 with CI [0.00258, 0.00273]. These correspond to per-policy log-likelihood lift intervals of approximately [0.00092, 0.00104] and [0.00129, 0.00137], respectively. These seed-level intervals are computed on paired single-seed differences and therefore differ slightly from the seed-averaged ensemble gaps reported in Table~\ref{tab:main-scaling}, because prediction averaging changes the loss. They should be interpreted as training-run stability diagnostics, with observation-level policy uncertainty left for future work. For scale, the 0.00315 mean-deviance improvement corresponds to a total deviance reduction of about 63 on a 20,000-policy held-out portfolio, or an aggregate held-out log-likelihood lift of 31.5 nats on the natural-logarithm scale. Equivalently, this is a 31.5-unit improvement in the held-out log-likelihood assigned to the observed claim-count vector on that scale. We avoid interpreting this likelihood ratio as a direct business value multiplier: the business interpretation depends on the pricing layer, severity or pure-premium modelling, calibration, margins, expenses, retention, competitive effects, and governance constraints. Recent pricing work accordingly evaluates deviance alongside calibration, risk differentiation, competitiveness, loss ratios, and fairness, while separate business-impact frameworks link model validation metrics to loss-ratio assumptions \cite{Israni2026DualEvaluation,Hedges2025LossRatio}.} Figure~\ref{fig:practical-significance} visualizes the implied per-policy log-likelihood lift across training fractions.

\bl{The supplementary public French MTPL (FMTPL) workflow provides a reproducible illustration of this interpretation on non-confidential data. This dataset, technically the \texttt{freMTPL2} frequency file within \texttt{CASdatasets}, has been extensively studied in recent actuarial pricing and actuarial machine-learning literature. In that smaller reproduction, the GLM and TabM-mini differ only modestly at the aggregate portfolio level, but the improved deviance corresponds to visible changes in cell-level indicated frequencies, including driver-age bands and age-by-bonus-malus interactions. We view the supplement as showing how likelihood gains can translate into actuarial model-output diagnostics, rather than as a second production-scale validation of the proprietary-data results.}

\subsection{\texorpdfstring{\bl{Limitations}}{Limitations}}
\label{sec:limitations}

Our empirical scaling claims are necessarily conditional on the experimental regime. Results are based on a single large motor insurance dataset with a fixed feature set, so fitted exponents and ranking crossovers may differ for other lines of business, jurisdictions, feature engineering choices, or targets (frequency versus severity versus pure premium). Our primary train/test split is an IID/random split; time-based splits or deployment-style temporal evaluation could change both absolute performance and scaling trends in the presence of non-stationarity. While we include strong GLM, FFN, and multiple Transformer/TabM families, we do not present a full gradient-boosted decision tree (GBDT) baseline sweep with comparable hyperparameter budgets, so our conclusions should be read as comparisons among the model families studied rather than a definitive leaderboard across all tabular methods. We do not conduct systematic hyperparameter optimization within each family; a more exhaustive tuning effort could improve absolute performance and shift specific crossovers, but our main goal is to compare scaling trends under a consistent recipe rather than to claim the best possible leaderboard scores for each method. \bl{The seed-level bootstrap intervals added in Table~\ref{tab:crossover-gaps} quantify variability across repeated training runs, not sampling uncertainty over individual held-out policies. A paired per-policy bootstrap, computed from policy-level loss differences between competing models on the same test rows, would be a useful next validation step for the most important crossover and practical-significance claims.} \bl{Likewise, the embedding/backbone parameter decomposition in Table~\ref{tab:beta_param_split} is not a matched ablation: a fixed-total-parameter experiment that varies the allocation between embedding tables and backbone weights would be a useful next step for testing whether the parameter-scaling exponent changes.} Finally, our scaling fits extrapolate within the observed range (up to $\sim 4\times 10^6$ rows and the capacities trained here); extending claims to substantially larger datasets (e.g., $\gtrsim 10^7$ rows) or to materially different compute budgets requires additional experiments.

\subsection{\texorpdfstring{\bl{What did we learn?}}{What did we learn?}}
\label{sec:what-did-we-learn}

\begingroup\color{blue}
For a pricing actuary, the practical message is that scaling laws are a way to put a measured slope on familiar questions about credibility, model complexity, and diminishing returns. The paper asks how predictably each model family improves when we add more policy-period rows, more parameters, or more training compute.

The first lesson is that actuarial frequency models can sit in a genuine scaling regime. In this motor portfolio, held-out Poisson deviance falls smoothly as the labelled training set grows. This is potentially surprising. Ratemaking data are tabular, claims are sparse, signal-to-noise ratios are low, and the response is shaped by underwriting, portfolio mix, and operational data processes. Smooth power-law behavior was therefore not guaranteed simply because similar laws appear in language modelling. Observing such behavior on one large real insurance portfolio is best read as an existence result and a calibration point: scaling laws can be a useful empirical lens for ratemaking, while the exact exponents should still be expected to vary across lines of business, jurisdictions, feature sets, temporal splits, and targets. This means that smaller experiments can give useful evidence about how much additional data or capacity may be worth, although the estimates remain local to the dataset, target, features, and validation design.

The second lesson is that the model family matters as much as the resource being scaled. The GLM remains a strong, stable benchmark and is especially hard to beat when data are limited. TabM and TabM-mini make better use of additional rows and parameters in the multi-million-row regime. Additional data helps supervised Transformers; additional parameters alone do not automatically translate into better test deviance. The stronger Transformer results come when we add tabular inductive bias or auxiliary self-supervised signal.

The third lesson is that small deviance changes should be interpreted on the likelihood scale. A difference such as 0.002--0.003 in mean Poisson deviance can look small on the page, but over a portfolio it represents a systematic improvement in the likelihood assigned to emerging claim counts. The supplementary material makes this tangible on the French MTPL (FMTPL) data: the lower-deviance TabM model produces a materially different two-way interaction surface between two covariates, illustrating that a small average deviance gain can correspond to a substantially different model that is more closely calibrated to the observed experience. This is a statistical model-fit statement. Turning it into loss ratio, profit, or premium adequacy still requires severity modelling, calibration, expenses, margins, retention effects, and governance constraints.

The final lesson is operational. When labelled claim outcomes are scarce, the most defensible path is usually a stable benchmark, careful validation, and modest model complexity. When labelled data reach the multi-million-row scale, capacity and variance reduction become more attractive. Since insurers cannot instantly create emerged claim outcomes, quote and policy records without outcomes may be valuable for self-supervised representation learning before the supervised frequency model is fitted.
\endgroup

\subsection{Conclusions}
\label{sec:conclusions-sub}

We have demonstrated that neural scaling laws - the smooth power-law relationships between data, parameters, compute, and generalization loss that underpin modern large-language-model research - also hold in actuarial tabular data. Using a multi-million-row motor insurance dataset and a consistent protocol across nested training fractions and five random seeds per configuration, we compared GLMs, FFNs, tabular Transformers, and TabM/TabM-mini architectures, including several novel Transformer enhancements developed for this study (MultiCLS pooling, TokenMoE routing, layer-wise value embeddings, TabM-style adaptation, and swap-style self-supervision).

Three implications follow for actuarial practice. First, as datasets grow beyond $\sim 10^5$ rows, it becomes rational to invest in higher-capacity tabular deep models and variance-reduction techniques such as small ensembles - precisely because these models continue to improve with additional data while GLMs saturate. \bl{Second, Transformer scaling in this regime depends on depth, width, auxiliary inductive bias, and training signal, so ``Transformer scaling'' should be treated as an architectural and loss function objective design problem.} Third, the fitted compute frontier and empirical floor indicate diminishing marginal returns to simply increasing compute without concurrent increases in data or improvements in inductive bias - a cautionary note for compute-heavy deployments.

Extending these experiments to additional lines of business, time-based splits, and broader baseline families (including GBDTs) is a natural next step. Studying severity and pure premium scaling in a comparable setup would clarify whether the exponents and rankings we observe for claim frequency generalize across actuarial prediction tasks.

\section*{Acknowledgements}
We thank the Casualty Actuarial Society (CAS) for funding this research. We also thank the contributing insurer for providing the anonymized motor portfolio extract used in this study. We thank Morgan Bugbee and the CAS project group for their guidance and insights, and we thank the anonymous reviewers for comments that substantially improved the manuscript.

\bibliography{yieldcurve}

\clearpage
\pagebreak

 \appendix

\section{\texorpdfstring{\bl{Notation Reference}}{Notation Reference}}
\label{app:math_primer}

\bl{This appendix collects the notation used throughout the paper. The prose explanation of the frequency task, Poisson deviance, and scaling-law quantities now appears earlier in Section~\ref{sec:rationale_explainer}; dense component-level model formulae are kept in Appendix~\ref{app:component_formulae}.}

\subsection{Quick reference: symbols and dimensions}
\label{app:math_quickref}

\begin{table}[htbp]
\centering
\begingroup\color{blue}
\beautytable
\caption{\bl{Core notation for the data, loss, and scaling-law quantities.}}
\begin{adjustbox}{max width=\textwidth}
\begin{tabular}{@{}L{0.18\textwidth}L{0.76\textwidth}@{}}
\toprule
\textbf{Symbol} & \textbf{Meaning (units / typical shape)} \\
\midrule
$i$ & Observation (policy-period row) index. \\
$t$ & Feature (rating-factor column / token) index; $t\in\{1,\ldots,T\}$. \\
$N$ & Training set size in rows (examples). \\
\bl{$N_f$} & \bl{Number of training rows at training fraction $f$; $f$ indexes the nested subset and $N_f$ is the row count used in scaling fits.} \\
$E_i$ & Exposure for row $i$ (exposure-years); used as an offset. \\
$y_i$ & Claim count for row $i$ (non-negative integer; in our extract $y_i\in\{0,1\}$). \\
$r_i$ & Claim \emph{rate} for row $i$ (claims per exposure-year). \\
$\lambda_i$ & Mean claim \emph{count} for row $i$; $\lambda_i = E_i r_i$. \\
$\eta_i$ & Log-rate (linear predictor); $\eta_i=\log r_i$. \\
$f_\theta(\cdot)$ & Model mapping covariates to $\eta$; parameters $\theta$ include all trainable weights. \\
$P$ & Number of trainable parameters; we also use $P_{\text{emb}}$ and $P_{\text{non-emb}}$ for embedding vs non-embedding subsets. \\
$C_1$, $C_{\mathrm{ens}}$ & \bl{$C_1$ is the training FLOPs/epoch proxy for one seed run (per model). $C_{\mathrm{ens}}=5C_1$ is the effective per-epoch compute proxy used when plotting the five-seed ensemble deviance. Total training compute is approximately $C_1\times$ (epochs trained) per seed, or $C_{\mathrm{ens}}\times$ (epochs trained) for the five-seed ensemble if all seeds train for the same number of epochs.} \\
$L$ & Test loss (Poisson deviance) reported in the paper. \\
$L_\infty$ & Empirical performance floor used in scaling fits (not a literal Bayes risk). \\
$\alpha,\beta,\gamma$ & Data-, parameter-, and compute-scaling exponents respectively. \\
\bl{$A_N,A_P,A_C$} & \bl{Fitted amplitude constants in the data-, parameter-, and compute-scaling laws. We use indexed amplitudes in the revised paper to avoid unindexed $B$ or $D$ being confused with batch size, data size, or parameter count.} \\
\bottomrule
\end{tabular}
\end{adjustbox}
\endgroup
\end{table}

\begin{table}[htbp]
\centering
\begingroup\color{blue}
\beautytable
\begin{adjustbox}{max width=\textwidth}
\begin{tabular}{@{}L{0.18\textwidth}L{0.76\textwidth}@{}}
\toprule
\textbf{Symbol} & \textbf{\bl{Architecture-specific meaning}} \\
\midrule
\bl{$T$} & \bl{Number of feature tokens/covariates after preprocessing.} \\
\bl{$b,d$} & \bl{Embedding dimension and post-positional-fusion token dimension.} \\
\bl{$M$} & \bl{Number of attention heads when discussing multi-head attention. The introduction no longer uses $M$ for Monte Carlo simulation count.} \\
\bl{$K$} & \bl{Number of implicit TabM/TabM-mini ensemble members.} \\
\bl{$E$} & \bl{Number of TokenMoE experts; context distinguishes this from exposure $E_i$.} \\
\bl{$n_{\mathrm{cls}},n_{\mathrm{swap}}$} & \bl{Number of main CLS tokens and swap-auxiliary CLS tokens.} \\
\bl{$\mathbf{X}_i^\circ,\mathbf{X}_i^+$} & \bl{Raw feature-token matrix for row $i$ and the corresponding token matrix after positional fusion and appending CLS tokens.} \\
\bl{$\mathbf{H}_{\mathrm{cls},i}$} & \bl{Matrix of final CLS states used for multi-CLS readout.} \\
\bl{$\mathbf{Q},\mathbf{K},\mathbf{V}$} & \bl{Query, key, and value projections inside attention.} \\
\bl{$\mathbf{V}^{\mathrm{mix}}_\ell$} & \bl{Layer-$\ell$ mixed value tensor in head-fix variants; defined only in the component formulae in Appendix~\ref{app:component_formulae}.} \\
\bl{$B_{\mathrm{batch}}$} & \bl{Minibatch size in Appendix~\ref{app:component_formulae}; we avoid using bare $B$ as a scaling-law amplitude in the revised empirical notation.} \\
\bottomrule
\end{tabular}
\end{adjustbox}
\caption{\bl{Additional notation for the neural architecture descriptions.}}
\label{tab:architecture_notation}
\endgroup
\end{table}

\FloatBarrier

\clearpage
\section{Technical Appendix: Model Configurations}
\label{app:model_configs}

This appendix records (i) the component-level mathematical definitions used in our implementations and (ii) the main architectural hyperparameters for all model variants. Unless noted, positional embedding dimension equals embedding dimension; baseline/multi-CLS variants concatenate positional embeddings, while head-fix/TokenMoE variants add them (with an optional projection). All models use the exposure-scaled Poisson objective.

\subsection{Experiment Model Identifiers}
\label{app:model_ids}

Table~\ref{tab:model_identifier_map} lists the models we utilized in this work grouped by family.

\begin{table}[htbp]
\centering
\begingroup\color{blue}
\caption{Experiment model identifiers grouped by family.}
\label{tab:model_identifier_map}
\beautytablecompact
\begin{adjustbox}{max width=\textwidth}
\begin{tabular}{@{}L{0.22\textwidth}L{0.74\textwidth}@{}}
\toprule
\textbf{Family} & \textbf{Identifiers} \\
\midrule
GLM & \texttt{glm} \\
FFN & \texttt{ffn\_very\_small}, \allowbreak \texttt{ffn\_small}, \allowbreak \texttt{ffn\_medium}, \allowbreak \texttt{ffn\_large}, \allowbreak \texttt{ffn\_very\_large} \\
TabM & \texttt{tabm\_small}, \allowbreak \texttt{tabm\_medium}, \allowbreak \texttt{tabm\_large}, \allowbreak \texttt{tabm\_xlarge} \\
TabM-mini & \texttt{tabm\_mini\_small}, \allowbreak \texttt{tabm\_mini\_medium}, \allowbreak \texttt{tabm\_mini\_large}, \allowbreak \texttt{tabm\_mini\_xlarge}, \allowbreak \texttt{tabm\_mini\_xlarge\_2}, \allowbreak \texttt{tabm\_mini\_xlarge3} \bl{\textnormal{(reported as gigantic)}} \\
Transformer (vanilla) & \texttt{transformer\_tiny}, \allowbreak \texttt{transformer\_small}, \allowbreak \texttt{transformer\_medium}, \allowbreak \texttt{transformer\_large} \\
Transformer (multi-CLS) & \texttt{transformer\_multicls\_tiny}, \allowbreak \texttt{transformer\_multicls\_small}, \allowbreak \texttt{transformer\_multicls\_medium}, \allowbreak \texttt{transformer\_multicls\_large}, \allowbreak \texttt{transformer\_multicls\_large\_4}, \allowbreak \texttt{transformer\_multicls\_large\_16}, \allowbreak \texttt{transformer\_multicls\_large\_largeffn\_noreg} \\
Transformer (enhanced) & \texttt{transformer\_enhanced\_tiny}, \allowbreak \texttt{transformer\_enhanced\_small}, \allowbreak \texttt{transformer\_enhanced\_medium}, \allowbreak \texttt{transformer\_enhanced\_large} \\
Transformer (TokenMoE) & \texttt{transformer\_tokenmoe\_tiny}, \allowbreak \texttt{transformer\_tokenmoe\_small}, \allowbreak \texttt{transformer\_tokenmoe\_medium}, \allowbreak \texttt{transformer\_tokenmoe\_large} \\
Transformer (TokenMoE+SSL) & \texttt{transformer\_tokenmoe\_ssl\_tiny}, \allowbreak \texttt{transformer\_tokenmoe\_ssl\_small}, \allowbreak \texttt{transformer\_tokenmoe\_ssl\_medium}, \allowbreak \texttt{transformer\_tokenmoe\_ssl\_large}, \allowbreak \texttt{transformer\_tokenmoe\_ssl\_xlarge} \\
Transformer+TabM-mini & \texttt{transformer\_multicls\_tabm\_mini\_tiny}, \allowbreak \texttt{transformer\_multicls\_tabm\_mini\_small}, \allowbreak \texttt{transformer\_multicls\_tabm\_mini\_medium}, \allowbreak \texttt{transformer\_multicls\_tabm\_mini\_large} \\
\bottomrule
\end{tabular}
\end{adjustbox}
\endgroup
\end{table}

\subsection{Component Formulae}
\label{app:component_formulae}
We summarize the reusable building blocks used across our model families. We use $T$ for the number of feature tokens (covariates), $n_{\text{cls}}$ for the number of CLS tokens, $M$ for the number of attention heads, and $d$ for the token dimension (after positional fusion). Unless otherwise noted, operations are applied row-wise over tokens and batch-wise over observations.

\subsubsection{Exposure-scaled Poisson mean and loss}
All models predict a positive mean claim count $\hat{\lambda}_i$ via
\begin{equation}
\hat{r}_i = \exp(\eta_i), \qquad \hat{\lambda}_i = E_i \hat{r}_i, \qquad \eta_i = f_\theta(x_i),
\end{equation}
where $E_i$ is earned exposure, $\hat{r}_i$ is the predicted claim rate (per exposure-year), and $\eta_i$ is the predicted log-rate. With claim count $y_i\in\mathbb{N}$, the (per-observation) Poisson negative log-likelihood is
\begin{equation}
\mathcal{L}_{\text{Pois}}(y_i,\hat{\lambda}_i) = \hat{\lambda}_i - y_i\log(\hat{\lambda}_i + \varepsilon),
\end{equation}
with small $\varepsilon>0$ for numerical safety.

\subsubsection{GLM baseline}
After preprocessing/encoding features into a vector $z_i\in\mathbb{R}^T$, the GLM uses a linear predictor
\begin{equation}
\eta_i = \beta_0 + \sum_{t=1}^T \beta_t z_{i,t},
\end{equation}
with the exposure-scaled Poisson mean $\hat{\lambda}_i=E_i\exp(\eta_i)$.

\subsubsection{Feed-forward network (FFN)}
Let $\mathbf{v}_i\in\mathbb{R}^F$ denote the concatenation of all learned embeddings for policy $i$ (equivalently, flattening $\mathbf{X}^\circ$ across tokens). An $L$-layer FFN computes
\begin{align}
\mathbf{h}_i^{(0)} &= \mathbf{v}_i,\\
\mathbf{h}_i^{(\ell)} &= \sigma(\mathbf{W}_\ell \mathbf{h}_i^{(\ell-1)}+\mathbf{b}_\ell),\qquad \ell=1,\ldots,L,\\
\eta_i &= \mathbf{w}^\top \mathbf{h}_i^{(L)} + b,
\end{align}
where $\sigma$ is a pointwise nonlinearity (ReLU in our FFN baselines) and dropout may be applied between layers.

\subsubsection{Quantile binning and entity embeddings}
For each covariate $x_t$ we apply a preprocessing map $q_t(\cdot)$ that yields an integer code in $\{0,\ldots,n_t-1\}$. For continuous variables, $q_t$ is a quantile-binning map; for categorical variables, $q_t$ is an integer encoding of levels. Each covariate is embedded via an embedding matrix $\mathbf{E}_t\in\mathbb{R}^{n_t\times b}$:
\begin{equation}
\mathbf{e}_t(x_t) = \mathbf{E}_t[q_t(x_t)] \in \mathbb{R}^b,
\end{equation}
and the raw token matrix is
\begin{equation}
\mathbf{X}^\circ = [\mathbf{e}_1(x_1)^\top;\ldots;\mathbf{e}_T(x_T)^\top]\in\mathbb{R}^{T\times b}.
\end{equation}
The tokenization parameters sum to $\varrho^{\rm input}=b\sum_{t=1}^T n_t$.

\subsubsection{Extended embeddings for v streams and MoE banks}
Several Transformer variants allocate a larger per-token embedding vector than the base dimension $b$ in order to provide (i) layer-wise value embeddings (the v stream) and (ii) an expert bank for TokenMoE.
Let $L$ denote the number of Transformer layers.

\emph{Layer-wise v embeddings.} In enhanced/head-fix variants, each covariate embedding table outputs an extended vector in $\mathbb{R}^{(L+1)b}$,
\begin{equation}
\mathbf{e}^{\rm all}_t(x_t)=\mathbf{E}^{\rm all}_t[q_t(x_t)]\in\mathbb{R}^{(L+1)b},
\end{equation}
which we split as
\begin{equation}
\mathbf{e}^{\rm all}_t(x_t)
=
\left[\mathbf{e}^{\rm base}_t(x_t);\ \mathbf{e}^{\rm v,1}_t(x_t);\ \cdots;\ \mathbf{e}^{\rm v,L}_t(x_t)\right],
\end{equation}
with $\mathbf{e}^{\rm base}_t(x_t)\in\mathbb{R}^{b}$ and $\mathbf{e}^{\rm v,\ell}_t(x_t)\in\mathbb{R}^{b}$.
Collecting across tokens yields a base matrix $\mathbf{X}^{\rm base}\in\mathbb{R}^{T\times b}$ and a concatenated v matrix $\mathbf{X}^{\rm v}\in\mathbb{R}^{T\times (Lb)}$.
Layer $\ell$ receives $\mathbf{X}^{\rm v,\ell}\in\mathbb{R}^{T\times b}$ obtained by slicing $\mathbf{X}^{\rm v}$ along the feature dimension.
When CLS tokens are appended, we append matching v embeddings set to zero (so the v stream is anchored in observed covariates).

\emph{TokenMoE bank.} In TokenMoE variants with $E$ experts, each embedding additionally stores an expert bank of $E$ slices, each of dimension $(L+1)b$.
Equivalently, the embedding output dimension becomes $(E+1)(L+1)b$ and can be written as
\begin{equation}
\mathbf{e}^{\rm all}_t(x_t)
=
\left[\mathbf{e}^{\rm base}_t(x_t);\ \mathbf{e}^{\rm v,1}_t(x_t);\ \cdots;\ \mathbf{e}^{\rm v,L}_t(x_t);\ \mathbf{u}_{t,1}(x_t);\ \cdots;\ \mathbf{u}_{t,E}(x_t)\right],
\end{equation}
where each expert slice $\mathbf{u}_{t,e}(x_t)\in\mathbb{R}^{(L+1)b}$ can itself be split into a base and v component.
The TokenMoE router selects (top-1) and mixes expert slices to produce the final base and v streams; see the routing definition below.

\subsubsection{Positional embeddings (concatenation vs addition)}
We learn a positional embedding $\mathbf{p}(t)\in\mathbb{R}^p$ for each token index $t\in\{1,\ldots,T\}$. In the concatenation mode,
\begin{equation}
\mathbf{x}_t = [\mathbf{e}_t(x_t);\mathbf{p}(t)] \in \mathbb{R}^{b+p}, \qquad \mathbf{X}=[\mathbf{x}_1^\top;\ldots;\mathbf{x}_T^\top]\in\mathbb{R}^{T\times d}, \ \ d=b+p.
\end{equation}
In the additive mode we optionally project to $\mathbb{R}^b$ and add,
\begin{equation}
\mathbf{x}_t = \mathbf{e}_t(x_t) + \mathbf{P}\mathbf{p}(t) \in \mathbb{R}^{b}, \qquad \mathbf{P}\in\mathbb{R}^{b\times p}.
\end{equation}
The positional embedding contributes $\varrho^{\rm position}=Tp$ parameters.

\subsubsection{CLS tokens and multi-CLS pooling}
We append learned CLS tokens $\mathbf{c}_1,\ldots,\mathbf{c}_{n_{\text{cls}}}\in\mathbb{R}^d$:
\begin{equation}
\mathbf{X}^+ = [\mathbf{X}^\top,\mathbf{c}_1^\top,\ldots,\mathbf{c}_{n_{\text{cls}}}^\top]^\top\in\mathbb{R}^{(T+n_{\text{cls}})\times d}.
\end{equation}
After the Transformer stack, we extract the final CLS states and form the matrix
\begin{equation}
\mathbf{H}_{{\rm cls},i} = [\mathbf{h}_{{\rm cls},1,i}^\top;\ldots;\mathbf{h}_{{\rm cls},n_{\text{cls}},i}^\top]\in\mathbb{R}^{n_{\text{cls}}\times d}.
\end{equation}
We use two readout patterns. In the \emph{concatenated} readout, we vectorize the CLS matrix (optionally after layer normalization) and map it to a log-rate:
\begin{equation}
\mathbf{h}^{\rm flat}_i=\text{vec}(\text{LN}(\mathbf{H}_{{\rm cls},i}))\in\mathbb{R}^{n_{\text{cls}}d},\qquad
\hat{\lambda}_i = E_i \exp(g(\mathbf{h}^{\rm flat}_i)).
\end{equation}
In the \emph{ensemble} readout, we apply one head per CLS token and average the resulting rates:
\begin{equation}
\hat{\lambda}_{i,c} = E_i \exp(g_c(\mathbf{h}_{{\rm cls},c,i})),\qquad
\hat{\lambda}_i = \frac{1}{n_{\text{cls}}}\sum_{c=1}^{n_{\text{cls}}}\hat{\lambda}_{i,c}.
\end{equation}
For $n_{\text{cls}}=1$ the two readouts coincide.

\subsubsection{Column weighting}
We learn a per-token weight $w_t\in[0,1]$ by storing an unconstrained parameter $a_t\in\mathbb{R}$ and clipping (hard sigmoid):
\begin{equation}
w_t = \min\{1,\max\{0,a_t\}\}, \qquad \tilde{\mathbf{x}}_t = w_t\,\mathbf{x}_t.
\end{equation}
Optionally, we add a sparsity penalty $\mathcal{L}_{\text{cw}}=\lambda_{\text{cw}}\sum_{t=1}^T w_t$.

\subsubsection{Self-attention with learned bias and temperature}
Let $\mathbf{Z}\in\mathbb{R}^{S\times d}$ denote a token matrix ($S=T+n_{\text{cls}}$). For head $m\in\{1,\ldots,M\}$ we compute (using layer normalization and a smooth activation $\phi$, GELU in our implementation)
\begin{equation}
\mathbf{Q}_m=\phi(\text{LN}(\mathbf{Z})\mathbf{W}^Q_m),\quad
\mathbf{K}_m=\phi(\text{LN}(\mathbf{Z})\mathbf{W}^K_m),\quad
\mathbf{V}_m=\phi(\text{LN}(\mathbf{Z})\mathbf{W}^V_m),
\end{equation}
and attention weights
\begin{equation}
\mathbf{A}_m=\text{softmax}\!\left(\frac{\mathbf{Q}_m\mathbf{K}_m^\top+\mathbf{B}_m}{\tau_m}\right),\qquad
\mathbf{H}_m=\mathbf{A}_m\mathbf{V}_m,
\end{equation}
where $\mathbf{B}_m$ is a learned additive attention bias and $\tau_m>0$ is a learned per-head temperature.
The multi-head output is
\begin{equation}
\text{MHA}(\mathbf{Z})=\text{Concat}(\mathbf{H}_1,\ldots,\mathbf{H}_M)\mathbf{W}^O.
\end{equation}
Our Transformer blocks apply dropout and layer normalization on the branch output before the residual addition:
\begin{align}
\mathbf{Z}' &= \mathbf{Z} + \text{LN}\!\left(\text{Dropout}(\text{MHA}(\text{LN}(\mathbf{Z})))\right),\\
\mathbf{Z}''&= \mathbf{Z}' + \text{LN}\!\left(\text{Dropout}(\text{FFN}(\text{LN}(\mathbf{Z}')))\right).
\end{align}

\subsubsection{Gated FFN (SwiGLU)}
The FFN uses a gated activation. For input $\mathbf{u}\in\mathbb{R}^d$,
\begin{equation}
\text{SwiGLU}(\mathbf{u}) = (\mathbf{u}\mathbf{W}_1+\mathbf{b}_1)\odot \text{silu}(\mathbf{u}\mathbf{W}_2+\mathbf{b}_2),
\end{equation}
followed by a linear projection $\mathbf{W}_3$ to return to dimension $d$.

\subsubsection{Head-fix value mixing, sink tokens, and stabilizers}
Head-fix variants enrich the value pathway and stabilize deep training.

Let $\mathbf{V}^{\rm base}=\phi(\text{LN}(\mathbf{Z})\mathbf{W}^V)$ be the current-layer values, let $\mathbf{V}^{\rm prev}$ denote a cached value tensor from the previous layer, and let $\mathbf{V}^{\rm emb}=\phi(\mathbf{Z}^{\rm emb}\mathbf{W}^{V,\rm emb})$ be a layer-specific value embedding (when enabled). We mix these via learned logits $\boldsymbol{\alpha}\in\mathbb{R}^3$:
\begin{equation}
(w_0,w_1,w_2)=\text{softmax}(\boldsymbol{\alpha}),\qquad
\mathbf{V}^{\rm mix}=w_0\mathbf{V}^{\rm base}+w_1\mathbf{V}^{\rm prev}+w_2\mathbf{V}^{\rm emb}.
\end{equation}
Head-fix layers additionally apply RMS normalization to per-head queries/keys, optionally append \emph{sink tokens} (which appear only as keys/values), apply per-head scalar gating of attended values, and use residual-branch stabilizers (LayerScale, an FFN gate, and drop-path).

\emph{RMS normalization.} For a vector $\mathbf{u}\in\mathbb{R}^{d_h}$ define
\begin{equation}
\text{RMSNorm}_\epsilon(\mathbf{u})=\frac{\mathbf{u}}{\sqrt{\frac{1}{d_h}\|\mathbf{u}\|_2^2+\epsilon}}.
\end{equation}
For each head $m$ we apply $\mathbf{Q}_m\leftarrow\text{RMSNorm}_\epsilon(\mathbf{Q}_m)$ and $\mathbf{K}_m\leftarrow\text{RMSNorm}_\epsilon(\mathbf{K}_m)$ before computing attention scores.

\emph{Sink tokens.} Let $\mathbf{S}\in\mathbb{R}^{n_{\text{sinks}}\times d}$ be learned sink embeddings and let $s_{\text{sink}}>0$ be a fixed scale factor. For each head $m$ we extend keys/values as
\begin{equation}
\mathbf{K}_m^{\rm full} = [\mathbf{K}_m;\ \phi(\text{LN}(\mathbf{S})\mathbf{W}_m^K)],\qquad
\mathbf{V}_m^{\rm full} = [\mathbf{V}_m^{\rm mix};\ s_{\text{sink}}\phi(\text{LN}(\mathbf{S})\mathbf{W}_m^V)].
\end{equation}
Let $S=T+n_{\text{cls}}$ be the main sequence length and let $I_{\rm cls}=\{S-n_{\text{cls}}+1,\ldots,S\}$ denote CLS indices. We add a sink mask $\mathbf{M}^{\rm sink}\in\mathbb{R}^{S\times(S+n_{\text{sinks}})}$ to the attention logits, with entries
\begin{equation}
\mathbf{M}^{\rm sink}_{qk}=
\begin{cases}
0, & k\le S \ \text{or}\ (k>S\ \text{and}\ q\in I_{\rm cls}),\\
-\infty, & \text{otherwise},
\end{cases}
\end{equation}
so that only CLS queries can attend to sinks.

\emph{Per-head value gating.} After computing head output $\mathbf{H}_m=\mathbf{A}_m\mathbf{V}_m^{\rm full}$ we apply a learned scalar gate $s_m=\sigma(g_m)$ and set $\mathbf{H}_m\leftarrow \tilde{s}_m\,\mathbf{H}_m$, where $\tilde{s}_m$ applies dropout to $s_m$ during training.

\emph{LayerScale, FFN gate, and drop-path.} Let $\gamma^{\rm attn},\gamma^{\rm ffn}$ be trainable LayerScale scalars and let $g_{\rm ffn}$ be a trainable FFN gate. With drop probability $p$, define
\begin{equation}
\text{DropPath}_{p}(\mathbf{u})=\frac{B}{1-p}\,\mathbf{u},\qquad B\sim{\rm Bernoulli}(1-p),
\end{equation}
broadcast across non-batch dimensions. The stabilized residual updates can be written as
\begin{align}
\mathbf{Z}' &= \mathbf{Z} + \text{DropPath}_{p^{\rm attn}}\!\left(\gamma^{\rm attn}\,\text{Dropout}(\text{MHA}(\mathbf{Z}))\right),\\
\mathbf{Z}''&= \mathbf{Z}' + \text{DropPath}_{p^{\rm ffn}}\!\left(\gamma^{\rm ffn}\,\sigma(g_{\rm ffn})\,\text{Dropout}(\text{FFN}(\text{LN}(\mathbf{Z}')))\right).
\end{align}
In our implementation the output projections in attention and FFN blocks are initialized at (near) zero, so that residual branches start close to the identity map, improving early optimization stability in deep stacks.

\subsubsection{TokenMoE routing and load balancing}
TokenMoE implements token-wise top-1 routing over an expert bank of embedding slices. Let $\mathbf{b}_{i,t}\in\mathbb{R}^{b}$ denote the \emph{base} embedding stream for token $t$ of observation $i$, and let $\mathbf{v}_{i,t}\in\mathbb{R}^{Lb}$ collect the $L$ layer-wise value streams. TokenMoE appends an expert bank $\mathbf{U}_{i,t}\in\mathbb{R}^{E\times(b+Lb)}$ (reshaped from a larger embedding tensor) and uses a router to compute logits
\begin{equation}
\mathbf{r}_{i,t} = \mathbf{W}_r\,\text{LN}(\mathbf{b}_{i,t}) + \boldsymbol{\epsilon}_{i,t},\qquad \boldsymbol{\epsilon}_{i,t}\sim\mathcal{N}(\mathbf{0},\sigma^2 \mathbf{I}),
\end{equation}
optionally preceded by a small router MLP. The routing probabilities are $\mathbf{p}_{i,t}=\text{softmax}(\mathbf{r}_{i,t})$. We select a top-1 expert $e_{i,t}=\arg\max_e p_{i,t,e}$ and form a straight-through gate
\begin{equation}
\mathbf{g}_{i,t} = \text{stopgrad}(\mathbf{1}_{e_{i,t}}-\mathbf{p}_{i,t})+\mathbf{p}_{i,t},
\end{equation}
so that forward routing uses the one-hot assignment while gradients flow through $\mathbf{p}_{i,t}$. The selected expert slice is
\begin{equation}
\mathbf{u}_{i,t} = \sum_{e=1}^{E} g_{i,t,e}\,\mathbf{U}_{i,t,e},
\end{equation}
which we split into base and value components $(\mathbf{u}^{\rm base}_{i,t},\mathbf{u}^{\rm v}_{i,t})$. With trainable mixing coefficients $\alpha_{\rm base},\alpha_{\rm v}\in(0,1)$ (implemented as sigmoids of unconstrained logits),
\begin{equation}
\tilde{\mathbf{b}}_{i,t}=\mathbf{b}_{i,t}+\alpha_{\rm base}\,\mathbf{u}^{\rm base}_{i,t},\qquad
\tilde{\mathbf{v}}_{i,t}=\mathbf{v}_{i,t}+\alpha_{\rm v}\,\mathbf{u}^{\rm v}_{i,t},
\end{equation}
or, in the hard-select mode, we replace $(\mathbf{b}_{i,t},\mathbf{v}_{i,t})$ by the selected slice.

To encourage utilization of all experts, we add a Switch-style load-balancing term based on routing \emph{importance} and empirical \emph{load} on a minibatch:
\begin{equation}
\text{importance}_e = \mathbb{E}_{i,t}[p_{i,t,e}],\qquad
\text{load}_e = \mathbb{E}_{i,t}[\mathbf{1}\{e_{i,t}=e\}],\qquad
\mathcal{L}_{\text{lb}} = E\sum_{e=1}^{E}\text{importance}_e\,\text{load}_e,
\end{equation}
and add $\lambda_{\text{lb}}\mathcal{L}_{\text{lb}}$ to the supervised loss.

\subsubsection{Swap-style self-supervision}
Swap-style self-supervision perturbs minibatches by swapping token embeddings across rows. \bl{For a minibatch of size $B_{\mathrm{batch}}$}, let $S_{i,t}\sim{\rm Bernoulli}(\alpha_{\text{swap}})$ indicate whether token $t$ of row $i$ is swapped, and let \bl{$\pi(i,t)\sim{\rm Uniform}\{1,\ldots,B_{\mathrm{batch}}\}$} denote a random donor row. The perturbed token embedding is
\begin{equation}
\mathbf{x}_{i,t}^{\rm swap} = (1-S_{i,t})\mathbf{x}_{i,t} + S_{i,t}\mathbf{x}_{\pi(i,t),t}.
\end{equation}
We append $n_{\text{swap}}=T$ additional ``swap CLS'' tokens (one per feature position). After the Transformer stack, let $\mathbf{h}^{\rm swap}_{i,t}\in\mathbb{R}^{d}$ denote the final state of the swap token for position $t$. A small sigmoid head predicts swap probabilities
\begin{equation}
\hat{s}_{i,t} = \sigma(h_{\psi}(\mathbf{h}^{\rm swap}_{i,t})),
\end{equation}
and we add a binary cross-entropy term
\begin{equation}
\mathcal{L}_{\text{swap}}=\frac{1}{B_{\mathrm{batch}}T}\sum_{i=1}^{B_{\mathrm{batch}}}\sum_{t=1}^{T}\Big(-S_{i,t}\log(\hat{s}_{i,t}+\varepsilon)-(1-S_{i,t})\log(1-\hat{s}_{i,t}+\varepsilon)\Big).
\end{equation}
The SSL models optimize $\mathcal{L}_{\text{Pois}}+\lambda_{\text{swap}}\mathcal{L}_{\text{swap}}$ (and, when TokenMoE is enabled, also include the load-balancing term $\lambda_{\text{lb}}\mathcal{L}_{\text{lb}}$).

\subsubsection{TabM-mini and TabM adapters}
TabM-mini applies a single multiplicative adapter that broadcasts a base feature vector $\mathbf{z}_i\in\mathbb{R}^F$ to $K$ ensemble members:
\begin{equation}
\mathbf{z}_{i,k} = \mathbf{z}_i \odot \mathbf{r}_k,\qquad \mathbf{r}_k\in\mathbb{R}^F.
\end{equation}
TabM uses BatchEnsemble-style rank-1 adapters throughout dense blocks. In one layer with shared kernel $\mathbf{W}$ and member-specific adapters $(\mathbf{r}_k,\mathbf{s}_k,\mathbf{b}_k)$,
\begin{equation}
\mathbf{h}^{\rm out}_{i,k} = \left((\mathbf{h}^{\rm in}_{i,k}\odot \mathbf{r}_k)\mathbf{W}\right)\odot \mathbf{s}_k + \mathbf{b}_k.
\end{equation}
Member log-rates $\eta_{i,k}$ are mapped to exposure-scaled rates $\hat{\lambda}_{i,k}=E_i\exp(\eta_{i,k})$ and averaged:
\begin{equation}
\hat{\lambda}_i=\frac{1}{K}\sum_{k=1}^{K}\hat{\lambda}_{i,k}.
\end{equation}

\subsection{Model Variant Feature Matrix}
\label{app:model_variant_matrix}
Table~\ref{tab:transformer_variant_matrix} provides a compact view of how the major Transformer-derived experiment families differ. The mathematical definitions of each component are given in Appendix~\ref{app:component_formulae}.

\begin{table}[htbp]
\centering
\begingroup\color{blue}
\caption{Transformer-family variant matrix (checkmark indicates that the component is central to the variant).}
\label{tab:transformer_variant_matrix}
\beautytablecompact
\begin{adjustbox}{max width=\textwidth}
\begin{tabular}{@{}L{0.36\textwidth}ccccc@{}}
\toprule
\textbf{Variant family} & \textbf{Multi-CLS} & \textbf{Head-fix} & \textbf{MoE} & \textbf{Swap SSL} & \textbf{TabM-mini head} \\
\midrule
\texttt{transformer\_\allowbreak *} &  &  &  &  &  \\
\texttt{transformer\_\allowbreak multicls\_\allowbreak *} & $\checkmark$ &  &  &  &  \\
\texttt{transformer\_\allowbreak enhanced\_\allowbreak *} & $\checkmark$ & $\checkmark$ &  &  &  \\
\texttt{transformer\_\allowbreak tokenmoe\_\allowbreak *} & $\checkmark$ &  & $\checkmark$ &  &  \\
\texttt{transformer\_\allowbreak tokenmoe\_\allowbreak ssl\_\allowbreak *} & $\checkmark$ &  & $\checkmark$ & $\checkmark$ &  \\
\texttt{transformer\_\allowbreak multicls\_\allowbreak tabm\_\allowbreak mini\_\allowbreak *} & $\checkmark$ &  &  &  & $\checkmark$ \\
\bottomrule
\end{tabular}
\end{adjustbox}
\endgroup
\end{table}

\subsection{Models Configurations}
\begin{table}[htbp]
\centering
\begingroup\color{blue}
\caption{GLM and FFN configurations.}
\label{tab:appendix_glm_ffn}
\beautytable
\begin{adjustbox}{max width=\textwidth}
\begin{tabular}{@{}L{0.30\textwidth}C{0.14\textwidth}L{0.24\textwidth}C{0.12\textwidth}C{0.14\textwidth}@{}}
\toprule
\textbf{Model} & \textbf{Embedding dim} & \textbf{Hidden layers} & \textbf{Dropout} & \textbf{Learning rate} \\
\midrule
\texttt{glm} & -- & -- & -- & 0.001 \\
\texttt{ffn\_very\_small} & 5 & \texttt{[16]} & 0.01 & 0.001 \\
\texttt{ffn\_small} & 10 & \texttt{[32, 16]} & 0.02 & 0.001 \\
\texttt{ffn\_medium} & 15 & \texttt{[64, 32]} & 0.03 & 0.001 \\
\texttt{ffn\_large} & 20 & \texttt{[128, 64, 32]} & 0.04 & 0.001 \\
\texttt{ffn\_very\_large} & 40 & \texttt{[256, 128, 64]} & 0.05 & 0.001 \\
\bottomrule
\end{tabular}
\end{adjustbox}
\endgroup
\end{table}

\begin{table}[htbp]
\centering
\begingroup\color{blue}
\caption{TabM and TabM-Mini configurations.}
\label{tab:appendix_tabm}
\beautytable
\begin{adjustbox}{max width=\textwidth}
\begin{tabular}{@{}L{0.26\textwidth}L{0.12\textwidth}rC{0.13\textwidth}L{0.24\textwidth}cc@{}}
\toprule
\textbf{Model} & \textbf{Type} & $\boldsymbol{K}$ & \textbf{Embedding dim} & \textbf{Dense layers} & \textbf{Dropout} & \textbf{Weight decay} \\
\midrule
\texttt{tabm\_small} & tabm & 32 & 16 & \texttt{[64, 32]} & 0.1 & 0.001 \\
\texttt{tabm\_medium} & tabm & 32 & 32 & \texttt{[128, 64, 32]} & 0.15 & 0.0015 \\
\texttt{tabm\_large} & tabm & 64 & 64 & \texttt{[256, 128, 64]} & 0.2 & 0.002 \\
\texttt{tabm\_xlarge} & tabm & 96 & 96 & \texttt{[384, 192, 96]} & 0.25 & 0.0025 \\
\texttt{tabm\_mini\_small} & tabm\_mini & 32 & 16 & \texttt{[64, 32]} & 0.1 & 0.001 \\
\texttt{tabm\_mini\_medium} & tabm\_mini & 32 & 32 & \texttt{[128, 64, 32]} & 0.15 & 0.0015 \\
\texttt{tabm\_mini\_large} & tabm\_mini & 64 & 64 & \texttt{[256, 128, 64]} & 0.2 & 0.002 \\
\bottomrule
\end{tabular}
\end{adjustbox}
\endgroup
\end{table}

\begin{table}[htbp]
\centering
\begingroup\color{blue}
\caption{TabM-Mini xlarge variants with expanded embeddings.}
\label{tab:appendix_tabm_xlarge_variants}
\beautytable
\begin{adjustbox}{max width=\textwidth}
\begin{tabular}{@{}L{0.28\textwidth}L{0.12\textwidth}rC{0.13\textwidth}L{0.28\textwidth}cc@{}}
\toprule
\textbf{Model} & \textbf{Type} & $\boldsymbol{K}$ & \textbf{Embedding dim} & \textbf{Dense layers} & \textbf{Dropout} & \textbf{Weight decay} \\
\midrule
\texttt{tabm\_mini\_xlarge} & tabm\_mini & 32 & 256 & \texttt{[512,\allowbreak 256,\allowbreak 128,\allowbreak 64,\allowbreak 32]} & 0.1 & 0.002 \\
\texttt{tabm\_mini\_xlarge\_2} & tabm\_mini & 32 & 256 & \texttt{[512,\allowbreak 256,\allowbreak 128,\allowbreak 64,\allowbreak 32]} & 0.1 & 0.002 \\
\texttt{tabm\_mini\_xlarge3} & tabm\_mini & 32 & 512 & \texttt{[1024,\allowbreak 512,\allowbreak 256,\allowbreak 128,\allowbreak 64,\allowbreak 32]} & 0.1 & 0.002 \\
\bottomrule
\end{tabular}
\end{adjustbox}
\endgroup
\end{table}

The identifier \texttt{tabm\_mini\_xlarge\_2} corresponds to the same architecture as \texttt{tabm\_mini\_xlarge} but evaluated via a separate workflow.

\begin{table}[htbp]
\centering
\begingroup\color{blue}
\caption{Vanilla Transformer configurations.}
\label{tab:appendix_transformer_base}
\beautytable
\begin{adjustbox}{max width=\textwidth}
\begin{tabular}{@{}L{0.31\textwidth}cccccc@{}}
\toprule
\textbf{Model} & $\boldsymbol{d}$ & \textbf{Heads} & \textbf{Layers} & \textbf{FFN dim} & $\boldsymbol{n_{\text{cls}}}$ & \textbf{Dropout (attn/ffn)} \\
\midrule
\texttt{transformer\_tiny} & 16 & 1 & 1 & 32 & 1 & 0.025/0.01 \\
\texttt{transformer\_small} & 24 & 2 & 2 & 48 & 1 & 0.025/0.015 \\
\texttt{transformer\_medium} & 40 & 2 & 3 & 80 & 1 & 0.035/0.025 \\
\texttt{transformer\_large} & 64 & 4 & 4 & 128 & 1 & 0.04/0.05 \\
\bottomrule
\end{tabular}
\end{adjustbox}
\endgroup
\end{table}

\begin{table}[htbp]
\centering
\begingroup\color{blue}
\caption{Multi-CLS Transformer configurations.}
\label{tab:appendix_transformer_multicls}
\beautytable
\begin{adjustbox}{max width=\textwidth}
\begin{tabular}{@{}L{0.34\textwidth}cccccc@{}}
\toprule
\textbf{Model} & $\boldsymbol{d}$ & \textbf{Heads} & \textbf{Layers} & \textbf{FFN dim} & $\boldsymbol{n_{\text{cls}}}$ & \textbf{Dropout (attn/ffn)} \\
\midrule
\texttt{transformer\_multicls\_tiny} & 16 & 1 & 1 & 32 & 2 & 0.025/0.01 \\
\texttt{transformer\_multicls\_small} & 24 & 2 & 2 & 48 & 4 & 0.025/0.015 \\
\texttt{transformer\_multicls\_medium} & 40 & 2 & 3 & 80 & 6 & 0.035/0.025 \\
\texttt{transformer\_multicls\_large} & 64 & 4 & 4 & 128 & 8 & 0.04/0.05 \\
\texttt{transformer\_multicls\_large\_4} & 64 & 4 & 4 & 128 & 4 & 0.04/0.05 \\
\texttt{transformer\_multicls\_large\_16} & 64 & 4 & 4 & 128 & 16 & 0.04/0.05 \\
\bottomrule
\end{tabular}
\end{adjustbox}
\endgroup
\end{table}

The enhanced variants (\texttt{transformer\_enhanced\_*}) share the same hyperparameters as the corresponding \texttt{transformer\_multicls\_*} sizes but use the head-fix attention block described in Section~\ref{sec:model_descriptions}. \bl{The variant \texttt{transformer\_multicls\_large\_largeffn\_noreg} widens the Transformer FFN and disables dropout and weight decay.}

\begin{table}[htbp]
\centering
\begingroup\color{blue}
\caption{Transformer plus TabM-Mini configurations.}
\label{tab:appendix_transformer_tabm_mini}
\beautytable
\begin{adjustbox}{max width=\textwidth}
\begin{tabular}{@{}L{0.36\textwidth}ccccccc@{}}
\toprule
\textbf{Model} & $\boldsymbol{d}$ & \textbf{Heads} & \textbf{Layers} & \textbf{FFN dim} & $\boldsymbol{n_{\text{cls}}}$ & $\boldsymbol{K}$ & \textbf{Dropout (attn/ffn)} \\
\midrule
\texttt{transformer\_\allowbreak multicls\_\allowbreak tabm\_\allowbreak mini\_\allowbreak tiny} & 16 & 1 & 1 & 32 & 2 & 8 & 0.025/0.01 \\
\texttt{transformer\_\allowbreak multicls\_\allowbreak tabm\_\allowbreak mini\_\allowbreak small} & 24 & 2 & 2 & 48 & 4 & 16 & 0.025/0.015 \\
\texttt{transformer\_\allowbreak multicls\_\allowbreak tabm\_\allowbreak mini\_\allowbreak medium} & 40 & 2 & 3 & 80 & 6 & 24 & 0.035/0.025 \\
\texttt{transformer\_\allowbreak multicls\_\allowbreak tabm\_\allowbreak mini\_\allowbreak large} & 64 & 4 & 4 & 128 & 8 & 32 & 0.04/0.05 \\
\bottomrule
\end{tabular}
\end{adjustbox}
\endgroup
\end{table}

All TokenMoE runs use \texttt{moe\_hard\_select = False} and \texttt{moe\_router\_hidden = 0}.

\begin{table}[htbp]
\centering
\begingroup\color{blue}
\caption{TokenMoE baseline configurations.}
\label{tab:appendix_tokenmoe_base}
\beautytablecompact
\begin{adjustbox}{max width=\textwidth}
\begin{tabular}{@{}L{0.29\textwidth}cccccccccc@{}}
\toprule
\textbf{Model} & $\boldsymbol{d}$ & \textbf{Heads} & \textbf{Layers} & \textbf{FFN dim} & $\boldsymbol{n_{\text{cls}}}$ & \textbf{Experts} & \textbf{Router noise} & \textbf{Load balance} & \textbf{Alpha (base/v)} & \textbf{Dropout (attn/ffn)} \\
\midrule
\texttt{transformer\_tokenmoe\_tiny} & 10 & 1 & 1 & 20 & 2 & 4 & 0.1 & 0.01 & 0.1/0.05 & 0.025/0.01 \\
\texttt{transformer\_tokenmoe\_small} & 14 & 2 & 2 & 28 & 4 & 4 & 0.1 & 0.01 & 0.1/0.05 & 0.025/0.015 \\
\texttt{transformer\_tokenmoe\_medium} & 24 & 2 & 3 & 48 & 6 & 6 & 0.1 & 0.01 & 0.1/0.05 & 0.035/0.025 \\
\texttt{transformer\_tokenmoe\_large} & 40 & 4 & 4 & 80 & 8 & 8 & 0.1 & 0.01 & 0.1/0.05 & 0.04/0.05 \\
\bottomrule
\end{tabular}
\end{adjustbox}
\endgroup
\end{table}

\begin{table}[htbp]
\centering
\begingroup\color{blue}
\caption{TokenMoE+SSL configurations (swap-style self-supervision). \bl{In the experiment logs these configurations were named \texttt{transformer\_tokenmoe\_*\_selfsup}; in the manuscript we refer to them as TokenMoE+SSL.}}
\label{tab:appendix_tokenmoe_ssl}
\beautytablecompact
\begin{adjustbox}{max width=\textwidth}
\begin{tabular}{@{}L{0.28\textwidth}ccccccccccccc@{}}
\toprule
\textbf{Model} & $\boldsymbol{d}$ & \textbf{Heads} & \textbf{Layers} & \textbf{FFN dim} & $\boldsymbol{n_{\text{cls}}}$ & \textbf{Experts} & \textbf{Router noise} & \textbf{Load balance} & \textbf{Alpha (base/v)} & \textbf{Swap alpha} & \textbf{Swap loss} & \textbf{Swap FFN} & \textbf{Dropout (attn/ffn)} \\
\midrule
\texttt{transformer\_\allowbreak tokenmoe\_\allowbreak ssl\_\allowbreak tiny} & 16 & 2 & 2 & 32 & 4 & 4 & 0.1 & 0.01 & 0.1/0.05 & 0.1 & 0.1 & 32 & 0.03/0.03 \\
\texttt{transformer\_\allowbreak tokenmoe\_\allowbreak ssl\_\allowbreak small} & 24 & 2 & 3 & 64 & 6 & 6 & 0.1 & 0.01 & 0.1/0.05 & 0.1 & 0.1 & 32 & 0.035/0.04 \\
\texttt{transformer\_\allowbreak tokenmoe\_\allowbreak ssl\_\allowbreak medium} & 32 & 4 & 4 & 96 & 8 & 8 & 0.1 & 0.01 & 0.1/0.05 & 0.1 & 0.1 & 32 & 0.04/0.045 \\
\texttt{transformer\_\allowbreak tokenmoe\_\allowbreak ssl\_\allowbreak large} & 40 & 4 & 4 & 80 & 8 & 8 & 0.1 & 0.01 & 0.1/0.05 & 0.1 & 0.1 & 32 & 0.04/0.05 \\
\texttt{transformer\_\allowbreak tokenmoe\_\allowbreak ssl\_\allowbreak xlarge} & 64 & 8 & 6 & 128 & 12 & 12 & 0.1 & 0.01 & 0.1/0.05 & 0.1 & 0.1 & 32 & 0.05/0.05 \\
\bottomrule
\end{tabular}
\end{adjustbox}
\endgroup
\end{table}

\bl{The identifier \texttt{transformer\_\allowbreak tokenmoessl\_\allowbreak xlarge} is an internal alias for \texttt{transformer\_\allowbreak tokenmoe\_\allowbreak ssl\_\allowbreak xlarge}; the paper uses the latter naming convention throughout.}

\subsection{Scaling by model size: full data split results}
\label{sec:results-size-ladders}
This subsection reports the seed-averaged ensemble test Poisson deviance for the main \emph{size ladders} within each model family, across the same nested training fractions used throughout the paper.
These tables complement the envelope summary in Table~\ref{tab:main-scaling} by making explicit how the best-performing size shifts with data, and by documenting the extent to which increasing Transformer size yields diminishing returns under purely supervised training.

\begin{table}[htbp]
\centering
\begingroup\color{blue}
\beautytablecompact
\begin{adjustbox}{max width=\textwidth}
\begin{tabular}{@{}L{0.34\textwidth}rrrrrr@{}}
\toprule
\textbf{Model} & \textbf{5\%} & \textbf{10\%} & \textbf{25\%} & \textbf{50\%} & \textbf{75\%} & \textbf{100\%} \\
\midrule
GLM & 0.29343 & 0.29284 & 0.29236 & 0.29212 & 0.29204 & 0.29200 \\
FFN (very small) & 0.29424 & 0.29267 & 0.29185 & 0.29139 & 0.29109 & 0.29098 \\
FFN (small) & 0.29790 & 0.29304 & 0.29187 & 0.29132 & 0.29101 & 0.29087 \\
FFN (medium) & 0.30055 & 0.29403 & 0.29166 & 0.29126 & 0.29096 & 0.29070 \\
FFN (large) & 0.29865 & 0.29438 & 0.29185 & 0.29134 & 0.29089 & 0.29063 \\
FFN (very large) & 0.30179 & 0.29450 & 0.29199 & 0.29118 & 0.29086 & 0.29058 \\
Transformer (tiny) & 0.29712 & 0.29463 & 0.29203 & 0.29175 & 0.29259 & 0.29104 \\
Transformer (small) & 0.29431 & 0.29274 & 0.29196 & 0.29148 & 0.29110 & 0.29090 \\
Transformer (medium) & 0.29358 & 0.29303 & 0.29166 & 0.29122 & 0.29100 & 0.29085 \\
Transformer (large) & 0.29340 & 0.29315 & 0.29185 & 0.29130 & 0.29100 & 0.29094 \\
\bottomrule
\end{tabular}
\end{adjustbox}
\caption{Baseline size-by-fraction results: seed-averaged ensemble test Poisson deviance for GLM, FFN, and vanilla Transformer size ladders.}
\label{tab:size_scaling_baselines}
\endgroup
\end{table}

\begin{table}[htbp]
\centering
\begingroup\color{blue}
\beautytablecompact
\begin{adjustbox}{max width=\textwidth}
\begin{tabular}{@{}L{0.38\textwidth}rrrrrr@{}}
\toprule
\textbf{Model} & \textbf{5\%} & \textbf{10\%} & \textbf{25\%} & \textbf{50\%} & \textbf{75\%} & \textbf{100\%} \\
\midrule
MultiCLS (tiny) & 0.29379 & 0.29263 & 0.29188 & 0.29148 & 0.29118 & 0.29106 \\
MultiCLS (small) & 0.29339 & 0.29234 & 0.29174 & 0.29136 & 0.29100 & 0.29100 \\
MultiCLS (medium) & 0.29321 & 0.29259 & 0.29168 & 0.29120 & 0.29106 & 0.29078 \\
MultiCLS (large) & 0.29322 & 0.29232 & 0.29179 & 0.29131 & 0.29089 & 0.29110 \\
Enhanced/head-fix (large) & 0.29421 & 0.29309 & 0.29225 & 0.29160 & 0.29140 & 0.29110 \\
TokenMoE (tiny) & 0.29362 & 0.29279 & 0.29213 & 0.29172 & 0.29146 & 0.29082 \\
TokenMoE (small) & 0.29381 & 0.29269 & 0.29205 & 0.29147 & 0.29105 & 0.29078 \\
TokenMoE (medium) & 0.29371 & 0.29280 & 0.29187 & 0.29145 & 0.29114 & 0.29080 \\
TokenMoE (large) & 0.29350 & 0.29263 & 0.29186 & 0.29126 & 0.29097 & 0.29074 \\
TokenMoE+SSL (tiny) & 0.29350 & 0.29269 & 0.29189 & 0.29146 & 0.29110 & 0.29075 \\
TokenMoE+SSL (small) & 0.29359 & 0.29291 & 0.29190 & 0.29142 & 0.29101 & 0.29068 \\
TokenMoE+SSL (medium) & 0.29337 & 0.29263 & 0.29186 & 0.29129 & 0.29087 & 0.29062 \\
TokenMoE+SSL (large) & 0.29332 & 0.29275 & 0.29192 & 0.29140 & 0.29087 & 0.29036 \\
TokenMoE+SSL (xlarge, full-data only) & -- & -- & -- & -- & -- & 0.29027 \\
Transformer+TabM (tiny) & 0.29321 & 0.29250 & 0.29186 & 0.29129 & 0.29099 & 0.29087 \\
Transformer+TabM (small) & 0.29288 & 0.29223 & 0.29166 & 0.29112 & 0.29085 & 0.29069 \\
Transformer+TabM (medium) & 0.29282 & 0.29217 & 0.29147 & 0.29124 & 0.29069 & 0.29056 \\
Transformer+TabM (large) & 0.29288 & 0.29219 & 0.29160 & 0.29108 & 0.29080 & 0.29059 \\
\bottomrule
\end{tabular}
\end{adjustbox}
\caption{Size-by-fraction results for progressive Transformer fixes. The TokenMoE+SSL xlarge point is a full-data-only run.}
\label{tab:size_scaling_transformer_fixes}
\endgroup
\end{table}

\begin{table}[htbp]
\centering
\begingroup\color{blue}
\beautytablecompact
\begin{adjustbox}{max width=\textwidth}
\begin{tabular}{@{}L{0.38\textwidth}rrrrrr@{}}
\toprule
\textbf{Model} & \textbf{5\%} & \textbf{10\%} & \textbf{25\%} & \textbf{50\%} & \textbf{75\%} & \textbf{100\%} \\
\midrule
TabM (small) & 0.29389 & 0.29252 & 0.29163 & 0.29121 & 0.29080 & 0.29066 \\
TabM (medium) & 0.29443 & 0.29325 & 0.29186 & 0.29108 & 0.29063 & 0.29048 \\
TabM (large) & 0.29681 & 0.29545 & 0.29263 & 0.29104 & 0.29028 & 0.28984 \\
TabM (xlarge) & 0.29599 & 0.29566 & 0.29249 & 0.29107 & 0.29021 & 0.28984 \\
TabM-mini (small) & 0.29380 & 0.29267 & 0.29167 & 0.29118 & 0.29097 & 0.29066 \\
TabM-mini (medium) & 0.29430 & 0.29323 & 0.29184 & 0.29110 & 0.29066 & 0.29034 \\
TabM-mini (large) & 0.29629 & 0.29515 & 0.29252 & 0.29104 & 0.29024 & 0.28989 \\
TabM-mini (xlarge) & 0.29610 & 0.29520 & 0.29263 & 0.29116 & 0.29046 & 0.28986 \\
TabM-mini (xlarge-2, full-data only) & -- & -- & -- & -- & -- & 0.28903 \\
TabM-mini (gigantic, full-data only) & -- & -- & -- & -- & -- & 0.28885 \\
\bottomrule
\end{tabular}
\end{adjustbox}
\caption{Size-by-fraction results for TabM and TabM-mini. The final two rows are full-data-only extensions.}
\label{tab:size_scaling_tabm}
\endgroup
\end{table}

\clearpage
\section{\texorpdfstring{\bl{Chinchilla-Style Scaling: Technical Explainer}}{Chinchilla-Style Scaling: Technical Explainer}}
\label{app:chinchilla_explainer}
\begingroup\color{blue}

This appendix records the derivation behind the Chinchilla-style allocation statement used in Section~\ref{sec:results-big-model-intuition}.

\subsection{Setup and notation}

We use \(N\) for labelled training rows and \(P\) for trainable parameters. Our data exponent is \(\alpha\) and our parameter exponent is \(\beta\). The additive bivariate scaling law is
\begin{equation}
\label{eq:appendix_chinchilla_law}
L(P,N)-L_{\infty}\approx aN^{-\alpha}+bP^{-\beta}.
\end{equation}
The first term is the finite-data contribution to reducible loss; the second is the finite-parameter contribution. Thus, at fixed \(P\), doubling \(N\) multiplies the data term by \(2^{-\alpha}\). At fixed \(N\), doubling \(P\) multiplies the parameter term by \(2^{-\beta}\).

Those isolated comparisons are not yet a compute-optimal statement. Chinchilla-style arguments ask a different question: under a compute constraint, how should \(N\) and \(P\) co-grow? We use the first-order approximation
\begin{equation}
\label{eq:appendix_compute_constraint}
C\propto NP.
\end{equation}
At fixed \(C\), increasing \(P\) forces \(N\) down, and increasing \(N\) forces \(P\) down. The problem is therefore a constrained allocation problem, not a direct comparison of \(\alpha\) and \(\beta\).

\subsection{Derivation}

Ignoring constants in the compute proxy, minimize
\[
U(P,N)=aN^{-\alpha}+bP^{-\beta}
\]
subject to \(NP=C\). The Lagrangian is
\[
\mathcal{J}(P,N,\lambda)=aN^{-\alpha}+bP^{-\beta}+\lambda(NP-C).
\]
The first-order conditions imply
\[
-\alpha aN^{-\alpha-1}+\lambda P=0,\qquad
-\beta bP^{-\beta-1}+\lambda N=0.
\]
Multiplying the first equation by \(N\) and the second by \(P\) gives the marginal balance condition
\begin{equation}
\label{eq:appendix_balance}
\alpha aN^{-\alpha}=\beta bP^{-\beta}.
\end{equation}
Solving Equation~\eqref{eq:appendix_balance} together with \(NP=C\) yields
\begin{align}
P_{\star}&\propto C^{\alpha/(\alpha+\beta)},\\
N_{\star}&\propto C^{\beta/(\alpha+\beta)},\\
N_{\star}&\propto P_{\star}^{\beta/\alpha}.
\end{align}
This is the allocation rule reported in the main text.

\subsection{Resolving the apparent paradox}

The apparent paradox is that the resource with the larger isolated exponent receives the smaller growth share under the compute-optimal allocation. The reason is that a larger exponent means that resource's own error term is easier to reduce. The optimizer spends enough on that resource to balance marginal returns, then the slower-shrinking term becomes the bottleneck.

Equation~\eqref{eq:appendix_balance} makes this precise:
\[
\frac{aN_{\star}^{-\alpha}}{bP_{\star}^{-\beta}}=\frac{\beta}{\alpha}.
\]
If \(\beta/\alpha<1\), the finite-parameter term is larger at the optimum and the compute-efficient path grows parameters faster than data. If the notation is reversed, as in some language-model scaling papers, the same algebra can imply faster token growth. This is why Chinchilla-style conclusions must always be read with the resource attached to each exponent stated explicitly.

Because all observed \(\beta/\alpha\) ratios are below one, the formal Chinchilla-style implication in our notation is parameter-heavy co-scaling: along the fitted local frontier, \(P\) grows faster than \(N\). The size of the ratio is important. Ratios close to zero, such as TokenMoE without SSL, mainly indicate weak parameter scaling: added capacity is not reliably converted into lower deviance under the supervised recipe. Ratios closer to one, especially for TabM/TabM-mini, describe a more balanced joint-scaling regime in which both additional labelled rows and additional parameters contribute measurably to lower held-out deviance.

\begin{center}
\begingroup\color{blue}
\beautytable
\begin{tabular}{@{}L{0.34\textwidth}cc@{}}
\toprule
\textbf{Family} & \(\boldsymbol{\beta/\alpha}\) & \(\boldsymbol{N_{\star}}\) growth if \(\boldsymbol{P_{\star}}\) doubles \\
\midrule
FFN & \(0.028/0.246\approx0.11\) & \(2^{0.11}\approx1.08\times\) \\
Vanilla transformer & \(0.025/0.194\approx0.13\) & \(2^{0.13}\approx1.09\times\) \\
MultiCLS transformer & \(0.037/0.182\approx0.20\) & \(2^{0.20}\approx1.15\times\) \\
Transformer+TabM & \(0.044/0.183\approx0.24\) & \(2^{0.24}\approx1.18\times\) \\
TokenMoE & \(0.008/0.202\approx0.04\) & \(2^{0.04}\approx1.03\times\) \\
TokenMoE+SSL & \(0.055/0.220\approx0.25\) & \(2^{0.25}\approx1.19\times\) \\
TabM/TabM-mini & \(0.148/0.409\approx0.36\) & \(2^{0.36}\approx1.28\times\) \\
\bottomrule
\end{tabular}
\endgroup
\end{center}

The table uses the TabM/TabM-mini summary exponent \(\alpha\approx0.409\). If instead we use the main-sweep TabM data exponent \(\alpha\approx0.309\), then \(\beta/\alpha\approx0.48\), or \(N_{\star}\) growth of about \(1.39\times\) per doubling of \(P_{\star}\). This alternative is less parameter-heavy than the \(0.36\) ratio in the table, but the ratio remains below one, so the allocation rule still has \(P_{\star}\) growing faster than \(N_{\star}\) while requiring meaningful data growth alongside capacity growth.

Thus, interpreted under the fitted additive law and \(C\propto NP\), labelled data remain relevant while the parameter-side term is the slower-shrinking bottleneck in this experimental range. For the weakest Transformer ratios, the practical conclusion is to improve the architecture or training signal before mechanically increasing size. For TabM/TabM-mini, the ratio is still below one but much larger, so parameter growth remains the faster axis while data growth remains a meaningful co-scaling requirement.

For our TabM/TabM-mini summary, \(\alpha\approx0.409\) and \(\beta\approx0.148\), so \(\beta/\alpha\approx0.36\). The heuristic allocation is therefore
\[
P_{\star}\propto C^{0.73},\qquad
N_{\star}\propto C^{0.27},\qquad
N_{\star}\propto P_{\star}^{0.36}.
\]
Both data and parameters increase with compute, but parameters increase faster along this fitted local frontier. This result is also only a per-epoch compute heuristic: with early stopping, realized training compute is closer to \(C_{\mathrm{tot}}\propto NP\,E_{\mathrm{stop}}(P,N)\), so an exact total-compute optimum would require modelling stopping time as part of the scaling law.

\endgroup

\end{document}